%% file: main.tex
\documentclass[a4paper,11pt]{article}
\usepackage[utf8]{inputenc}
\usepackage[english]{babel}
\usepackage[hidelinks]{hyperref}
\usepackage{graphicx}
\usepackage{diagbox} 
\usepackage[nottoc]{tocbibind} 
\usepackage{multirow}
\usepackage{tabularx}
\usepackage{footmisc} 

\usepackage{amsmath}
\usepackage{amsfonts}

\newcommand{\dataset}{\textsc{dast}} 
\newcommand{\q}[1]{``#1"}
\newcommand{\sklearn}{\texttt{scikit learn}}

\newcommand{\pheme}{\textsc{PHEME}}
\newcommand{\fone}{$F_1$}
\newcommand{\sota}{state-of-the-art}
\newcommand{\hmm}{\lambda}
\newcommand{\mshmm}{\omega}
\newcommand{\fake}{Fake News}

\newcommand{\R}{\mathbb{R}}

\begin{document}
\pagenumbering{gobble} 
\begin{titlepage}
	
	\newcommand{\HRule}{\rule{\linewidth}{0.5mm}} 
	
	\center 
	
	\textsc{\LARGE IT University of Copenhagen}\\[1.5cm] 
	\textsc{\Large MSc in Software Development}\\[0.5cm] 
	\textsc{\large Thesis project\\KISPECI1SE}\\[0.5cm] 
	
	\HRule \\[0.4cm]
	{ \huge \bfseries Danish Stance Classification and Rumour Resolution}\\[0.4cm] 
	\HRule \\[1.5cm]

	\begin{minipage}{0.5\textwidth}
		\begin{flushleft} \large
			\emph{Authors:}\\
			Anders E. \textsc{Lillie} \textit{aedl@itu.dk}\\
			Emil R. \textsc{Middelboe} \textit{erem@itu.dk}
		\end{flushleft}
	\end{minipage}
	~
	\begin{minipage}{0.4\textwidth}
		\begin{flushright} \large
			\emph{Supervisor:} \\
			Leon \textsc{Derczynski}\\\textit{leod@itu.dk} 
		\end{flushright}
	\end{minipage}\\[4cm]
	
	{\large July 2, 2019}\\[3cm] 
	
	\vfill 
	
\end{titlepage}
\clearpage

\input{abstract.tex}
\clearpage

\tableofcontents
\clearpage
\pagenumbering{arabic} 

\input{introduction.tex}
\input{background.tex}
\input{problem_analysis.tex}
\input{technologies.tex}
\input{data.tex}
\input{methods.tex}
\input{experiments_stance.tex}
\input{experiments_veracity.tex}
\input{discussion.tex}
\input{conclusion.tex}
\clearpage

\bibliography{references,thesis_prep} \clearpage
\listoffigures \clearpage
\listoftables \clearpage

\appendix
\input{dataset.tex} \clearpage
\input{experiment_data.tex} \clearpage
\input{app_experiments_veracity.tex} 

\end{document}

%% file: abstract.tex
\begin{abstract}
    The Internet is rife with flourishing rumours that spread through microblogs and social media. Recent work has shown that analysing the stance of the crowd towards a rumour is a good indicator for its veracity. One \sota{} system uses an LSTM neural network to automatically classify stance for posts on Twitter by considering the context of a whole branch, while another, more simple Decision Tree classifier, performs at least as well by performing careful feature engineering. One approach to predict the veracity of a rumour is to use stance as the only feature for a Hidden Markov Model (HMM). This thesis generates a stance-annotated Reddit dataset for the Danish language, and implements various models for stance classification. Out of these, a Linear Support Vector Machine provides the best results with an accuracy of 0.76 and macro $F_1$ score of 0.42. Furthermore, experiments show that stance labels can be used across languages and platforms with a HMM to predict the veracity of rumours, achieving an accuracy of 0.82 and $F_1$ score of 0.67. Even higher scores are achieved by relying only on the Danish dataset. In this case veracity prediction scores an accuracy of 0.83 and an \fone{} of 0.68. Finally, when using automatic stance labels for the HMM, only a small drop in performance is observed, showing that the implemented system can have practical applications. 
\end{abstract}

%% file: introduction.tex
\section{Introduction}
\label{introduction}
Social media has come to play a big role in our everyday lives as we use it to connect with our social network, but also to connect with the world. It is common to catch up on news through Facebook, or to be alerted with emerging events through Twitter. However these phenomena create a platform for the spread of rumours, that is, stories with unverified claims, which may or may not be true \cite{huang15}. This has lead to the concept of fake news, where the spreading of a misleading rumour is intentional \cite{shu2017fake}. Can we somehow automatically predict the veracity of rumours? Within recent years research has tried to tackle this problem \cite{qazvinian11}, but automated rumour veracity prediction is still in its infancy \cite{semeval_2019}.

This paper reports a thesis project carried out in the Spring semester, 2019, on the MSc programme in Software Development at the IT University of Copenhagen. The thesis investigates stance classification as a step for automatically determining the veracity of a rumour. Previous research has shown that the stance of a crowd is a strong indicator for veracity \cite{dungs18}, but that it is a difficult task to build a reliable classifier \cite{derczynski17}. Moreover a study has shown that careful feature engineering can have substantial influence on the accuracy of a classifier \cite{aker17}. A system able to verify or refute rumours is typically made up of four components: rumour detection, rumour tracking, stance classification, and veracity classification \cite{zubiaga18}. This project will mainly be concerned with stance classification and rumour veracity classification. A research paper on the subject was written in the Autumn semester, 2018, in the Thesis Preparation course, contributing as background research for this project \cite{thesis-prep}. Source code for the implementation of the system developed in this thesis project are publicly available on GitHub\footnote{\url{https://github.com/danish-stance-detectors}}.\\

Current research is mostly concerned with the English language (see section \ref{background}) \cite{thesis-prep}, and in particular data from Twitter is used as data source because of its availability and relevant news content. To our knowledge no research within this area has been carried out in a Danish context. 

\subsection{Research question}
The thesis project will attempt to answer the following questions: how do we build an automatic stance classification system for Danish? Further, how do we apply this system to verify or refute rumours and possibly detect fake news?

\subsection{Overview}
Background material and current research will be introduced in section \ref{background} with an overview of the research area, common approaches, and \sota{} systems. Section \ref{problem_analysis} will provide an analysis of the task at hand, considering different approaches, trade-offs and possible obstacles. Before going into a system description, technologies and frameworks utilised throughout the project is presented section \ref{technologies}. Section \ref{data} will introduce how the data has been gathered, and how it has been annotated using a custom built annotation tool. This is followed by section \ref{system_description}, which describes the models used for stance classification as well as feature vectors, and the approach taken for rumour veracity classification. Experiments are carried out and reported in section \ref{experiments} and \ref{sec:rumour_experiments}. Finally our findings are discussed in section \ref{discussion}, while the project is concluded in section \ref{conclusion}. 

%% file: background.tex
\section{Background}
\label{background}

Social Media and Big Data have been buzzwords for about a decade now, as the availability and accessibility of the Internet continually grows. While more people join social media, more and more data comes in circulation. Moreover, platforms such as Twitter, Facebook, and Reddit have become the primary sources of news for some people, because of the ability to follow real-time developments of events from first-hand sources \cite{huang15}. 

However, as you would possibly trust news that you hear in the radio or see in the TV, information spread on the Internet can be difficult to trust and verify. This gives rise to the spreading of \textit{rumours}, which can be defined as circulating information that is yet to be verified as true or false \cite[5.1]{shu2017fake}. Research has investigated why rumours are spread, and \cite{huang15} shows that, in particular in relation to crisis, physical and emotional proximity influence online information seeking and sharing behaviours. Furthermore some people exploit the phenomena of spreading rumours on social media for beneficial reasons such as finance or politics, which has come to be known as \textit{Fake News} \cite{shu2017fake}.

Because of circulating rumours and \fake{}, researchers have studied if and how we can use IT and computer science to detect and possibly debunk false statements \cite{derczynski17,shu2017fake}. One of the first studies in this area created a dataset with more than 10,000 tweets from Twitter over five different topics and built a system for detecting rumours as well as classifying tweets as being either supporting or denying \cite{qazvinian11}. Another study frames this task of identifying stance as a credibility assessment, and builds a system to automatically detect credibility from topics in collected newsworthy events from Twitter \cite{castillo11}.

This section will outline the research area and explore both previous and current relevant studies. First, a common architecture for rumour veracity classification is explored. Then related work, including stance classification, veracity classification, and \fake{} detection systems will be introduced. Finally \sota{} systems will be investigated.

\subsection{System architecture for rumour resolution}
\label{background:system_architecture}
\begin{figure}[h]
    \centering
    \includegraphics[width=\textwidth]{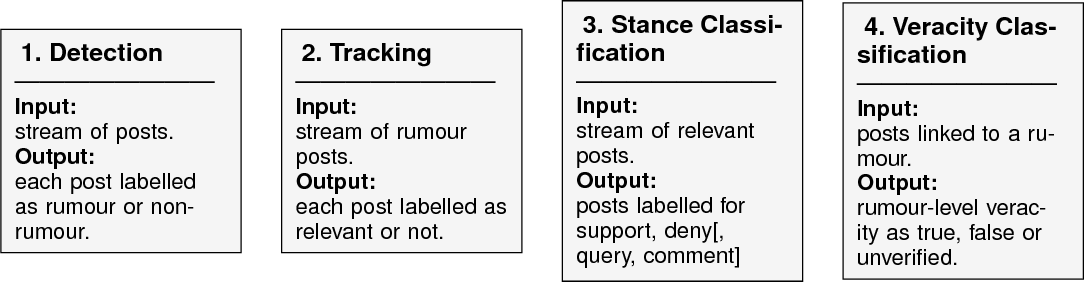}
    \caption{Rumour veracity classification system architecture. Source: \cite{zubiaga18} (From web version)}
    \label{fig:rumour_classification_system_architecture}
\end{figure}

One approach for determining veracity of rumours could be to divide the task into four sub-components as depicted in Figure \ref{fig:rumour_classification_system_architecture}.

\noindent That is, one must first do rumour detection, then track these rumours and feed them into a stance classifier, to ultimately perform veracity classification to determine whether the rumours are true or false. Expanding the components, one can define each task as the following:
\begin{enumerate}
    \itemsep0em
    \item Rumour detection
    \begin{enumerate}
        \itemsep0em
        \item Identify whether a piece of information constitutes a rumour
        \item Approach: binary classifier
        \item Input: a set of posts
        \item Output: a set posts, where each one is labelled as rumour or non-rumour
        \item Useful for \textit{emerging} rumours, but not necessary with \textit{a priori} rumours\footnote{\url{https://www.merriam-webster.com/dictionary/a\%20priori} 24-05-2019}
    \end{enumerate}
    \item Rumour tracking
    \begin{enumerate}
        \itemsep0em
        \item Once a rumour is identified, whether it being a priori or emerging, the \textit{tracking} component collects and filters posts discussing the rumour
        \item Approach: monitor social media to find posts discussing a rumour, while eliminating irrelevant posts
        \item Input: a rumour
        \item Output: a set of posts discussing the rumour
    \end{enumerate}
    \item Rumour stance classification
    \begin{enumerate}
        \itemsep0em
        \item Determine how each post is oriented towards a rumour's veracity
        \item Approach: multi-class classification
        \item Input: a set of posts associated with the same rumour
        \item Output: label of each post, where the labels are typically predefined as a set of types of stances, such as supporting, denying, querying, and commenting (SDQC).
        \item Can be useful for rumour veracity classification (next component), but can be omitted where stance is not considered useful 
    \end{enumerate}
    \item Rumour veracity classification
    \begin{enumerate}
        \itemsep0em
        \item Attempt to determine the actual truth value of a rumour
        \item Approach: binary classifier
        \item Input: a set of posts (could be collected by the rumour tracking component), and optionally stance labels
        \item Output: predicted truth value
        \item The input and output can optionally include relevant information from external sources, such as news media
    \end{enumerate}
\end{enumerate}

Together, these components allow for a system to automatically verify or refute rumours in (near) real-time, as you would have to wait for comments. However, as noted in the rumour detection component, a system can also be based on \textit{a priori} rumours, that is, historical events, which have concluded. This way the detection and tracking of rumours comprise of the task of finding rumours and gathering all relevant data tied to it. In this case the data can subsequently be used in stance classification and veracity classification. The result of this would be a model that is \textit{trained} on historical data, but is able to analyse and work as a rumour resolution system to unseen and possibly new/emerging rumours. 

Note that \textit{classification} will sometimes be used interchangeably with \textit{detection} and \textit{prediction} throughout the paper, as these more accurately apply for some contexts.

\subsection{Related work}
\label{ralated_work}
One of the major benefits of doing rumour veracity classification is its usefulness in debunking \fake{}. One major study within this area investigates \fake{} detection on social media and comes up with a \fake{} characterisation as well as a novel approach to building a detection system \cite{shu2017fake}. This research defines the following: while a rumour is a piece of circulating information whose veracity status is yet to be verified, \fake{} is articles that are intentionally and verifiable false. They dive into this subject based on the fact that social media has become the primary source for news. This leads to more noisy and lower quality news than found in traditional news. 

The introduction of \fake{} leads to a break with the authenticity of the whole news ecosystem as producers intentionally persuade consumers to accept biased or false beliefs \cite[1]{shu2017fake}. This changes the way people interpret and respond to \textit{real} news. In particular the key feature of \fake{} is its authenticity: \fake{} includes verifiably false information; and its intent: \fake{} is created with dishonest intention to mislead readers \cite[2.1]{shu2017fake}. What is interesting, is the psychological and social foundation of \fake{}. Due to naïve realism and confirmation bias, consumers tend to believe that their own perception of reality is the only accurate frame, while preferring to receive news that confirms their own bias \cite[2.2]{shu2017fake}. This is related to the ``echo chamber" effect, where users tend to form groups of like-minded people with polarised opinions, which facilitates the process of believing \fake{} \cite{echochamber}. \\


As an approach to tackle the problem of debunking \fake{}, The Fake News Challenge \cite{fakenewschallenge17} frames the problem as classifying stance for news articles as being either agreeing, disagreeing, or discussing (as well as unrelated) to a headline. This was an open research problem made available as a task/challenge for teams to participate in. The best scoring teams use both ensemble approaches of Decision Trees and CNNs as well as simple Multi-Layered Perceptrons \cite{hanselowski18}. It is an ongoing project, where stance detection is the first stage, just as introduced in section \ref{background:system_architecture}, serving as a ``useful building block in an AI-assisted fact-checking pipeline"\footnote{\label{fn:fake_news_challenge}\url{http://www.fakenewschallenge.org/} 24-04-2019}. 

Related to the task of verifying rumours and debunking \fake{} is the PHEME project, which deals with four kinds of false claims: rumours, disinformation, misinformation, and speculation \cite{pheme14}. As an extension to volume, velocity, and variety as being well known challenges working with Big Data in social media, PHEME introduces \textit{veracity} as being a fourth ``crucial, but hitherto largely unstudied, challenge"\footnote{\url{https://www.pheme.eu/} 24-04-2019}. In particular the project takes its name from the term \textit{meme}, which is ``an idea, behaviour, or style that spreads from person to person within a culture[..]"\footnote{\url{https://en.wikipedia.org/wiki/Meme} 24-04-2019}, containing information about veracity, but also the Greek goddess of fame and rumours. The project started in 2014 and ran for three years, yielding several studies and research projects within its area. 

Other projects and initiatives deal with \fake{} detection as the task of fact-checking, such as FEVER\footnote{\url{http://fever.ai/}} and Full Fact\footnote{\url{https://fullfact.org/}}. FEVER presents a shared task in \cite{Thorne18Fact} of classifying whether human-written factoid claims can be verified as supported or refuted using evidence retrieved from Wikipedia. Similarly Full Fact is a UK fact-checking charity, which performs automated end to end fact-checking by monitoring platforms such as Twitter and Facebook \cite{babakar17fullfact}. One related, but manual approach, exist in Denmark by Mandag Morgen\footnote{\url{https://www.mm.dk/}} who provides a fact-checking website, called ``TjekDet"\footnote{\url{https://www.mm.dk/tjekdet/}} (``check it"), which investigates misinformation in social media and online debate. Additionally TjekDet is a member of The International Fact-Checking Network (IFCN)\footnote{\url{https://www.poynter.org/ifcn/}}, which is a unit started in 2015 with the goal of uniting fact-checkers around the world. \\

``SemEval" (\textbf{Sem}antic \textbf{Eval}uation) is an ongoing series of evaluations of computational semantic analysis systems\footnote{\url{https://en.wikipedia.org/wiki/SemEval} 27-05-2019}. Each SemEval contains several tasks related to Natural Language Processing (NLP), Semantics and Computational Linguistics, for which teams are invited to submit solution systems. Within the last couple of years, SemEval has had tasks concerned specifically with rumour stance and rumour veracity classification, denoted as ``RumourEval" tasks. In particular task 8 in SemEval 2017 \cite{derczynski17} and task 7 in SemEval 2019 \cite{semeval_2019} concern themselves with these two subtasks. Resources including a stance labelled dataset is provided, which research teams should use to develop solution systems to tackle the task of determining rumour veracity and support for rumours. We have studied several of the relevant publications for SemEval, as they provide \sota{} research within our field of study and build upon each others' work \cite{thesis-prep}.

Additionally, task 6 in SemEval 2016 \cite{mohammad16} engages in detecting stance from tweets given a target entity, such as a person and organisation. The difficult part about this is the fact that the target may or may not be included in the tweet data, just as it may or may not be included in the target of opinion.

\subsection{State of the art}
\label{state_of_the_art}
This section presents \sota{} systems introduced within the last three years.

\subsubsection{Rumour stance classification}
\label{background:rumour_stance_classification}
The attitude that people express towards some statement can be used to predict veracity of rumours, and these attitudes can be modelled by stance classifiers. This section will present \sota{} systems for automatic stance classification. In particular Long-Short Term Memory (LSTM) neural network models are popular, as they have proven to be efficient for working with data within NLP (further described in section \ref{stance_lstm_classifier}). 

\cite{kochkina17} developed a stance classifier based on a ``Branch-LSTM" architecture: instead of considering a single tweet in isolation, whole branches are used as input to the classifier, capturing structural information of the conversation. The model is configured with several dense ReLU layers, a 50\% dropout layer, and a softmax output layer, scoring a 0.78 in accuracy and 0.43 macro \fone{} score. They are however unable to predict the under-represented ``denying" class.  

Another LSTM approach deals with the problem introduced above for the SemEval 2016 task 6 \cite{mohammad16}. The LSTM implements a bi-directional conditional structure, which classifies stance towards a target with the labels ``positive", ``negative", and ``neutral" \cite{augenstein16}. The approach is unsupervised, i.e. data is not labelled for the test targets in the training set. In this case the system achieves \sota{} performance with a macro $F_1$ score of 0.49, and further 0.58 when applying weak supervision.

A different approach is based on having well-engineered features for stance classification experiments using non-neural networks classifiers instead of Deep Learning (DL) methods \cite{aker17}. Common features such as CBOW and POS tagging are implemented, but are extended with problem-specific features, which are designed to capture how users react to tweets and express confidence in them. The best performing classifier is a Random Forest classifier, scoring an accuracy of 0.79\footnote{Unfortunately no $F_1$ score is reported, rendering us unable to compare the performance on that metric to the other state of the art results}. \\

RumourEval 2019 has been running in parallel with the writing of this thesis \cite{semeval_2019} and a first look at the scoreboard indicates very promising results\footnote{\label{fn:semeval2019}\url{https://competitions.codalab.org/competitions/19938} 26-05-2019}. With the Branch-LSTM approach as a baseline on the RumourEval 2019 dataset, scoring 0.4930 macro \fone{}, the ``BERT" system scores a 0.6167 macro $F_1$ \cite{butfit:semeval2019}. The implementation employs transfer learning on large English corpora, then an encoding scheme concatenates the embeddings of the source, previous and target post. Finally the output is fed through two dense layers to provide class probabilities. These BERT models are used in several different ensemble methods where the average class distribution is used as the final prediction. 

\subsubsection{Rumour veracity classification}
\label{background:rumour_veracity_classification}
Rumour veracity classification is considered a challenging task as one must typically predict a truth value from a single text, being the one that initiates the rumour. The best performing team for that task in RumourEval 2017 \cite{derczynski17} implements a Linear Support Vector Machine (SVM) with only few (useful) features \cite{enayet17}. They experiment with several common features such as hashtag existence, URL existence, and sentiment, but also incorporates an interesting feature of capturing whether a text is a question or not. Furthermore the percentage of replying tweets classified as supporting, denying, or querying from stance classification is applied. It is concluded that content and Twitter features were the most useful for the veracity classification task and score an accuracy of 0.53.

For the similar task, but different dataset, the second best scoring team, ``CLEARumor", in RumourEval 2019 \cite{semeval_2019} achieved an $F_1$ score of 0.286 (submitted) and since, 0.301 \cite{clearrumor:semeval2019}. According to the scoreboard\footref{fn:semeval2019} the best scoring team achieved an impressive 0.5765 $F_1$, but it seems that they have not published their work at the time of writing. CLEARumor implements a CNN-based deep-learning architecture for SDQC stance classification and use these estimates for predicting veracity through a Multi-Layered Perceptron (MLP) neural network. ELMo embeddings are used\footnote{\url{https://allennlp.org/elmo} 27-05-2019}, as a new word embeddings approach over for instance the widely used word2vec algorithm \cite{mikolov13word2vec}. Further, four auxiliary features are employed, including platform specific encodings for respectively Twitter and Reddit. The system in \cite{enayet17} introduced above is used as baseline, and as reference scores a macro \fone{} of 0.18 on the same test set.

While the above two systems engage in the task of resolving veracity given a single rumour text, another interesting approach is based on the use of crowd/collective stance, which is the set of stances over a conversation \cite{dungs18}. This system predicts the veracity of a rumour, based solely on crowd stance as well as tweet times. A Hidden Markov Model (HMM) is implemented, which is utilised such that individual stances over a rumour's lifetime is regarded as an ordered sequence of observations. This is then used to compare sequence occurrence probabilities for true and false rumours respectively. The best scoring model, which include both stance labels and tweet times, scores an $F_1$ of 0.804, while the HMM with only stance labels scores 0.756 $F_1$. The use of automatic stance labels from \cite{aker17} is also applied, which does not change performance much, proving the method to have practical applications. It is also shown that using the model for rumour veracity prediction is still useful when limiting the number of tweets to e.g. 5 and 10 tweets respectively. \\

This section has presented background theory and related work, including state of the art. This, together with a deeper investigation in \cite{thesis-prep}, allows us to analyse the problem at hand and choose an approach for tackling it, which will be discussed next, in section \ref{problem_analysis}.

%% file: problem_analysis.tex
\section{Problem Analysis}
\label{problem_analysis}


The task of rumour stance classification and veracity prediction is a difficult problem. The skewed nature of data from microblogs makes classifying minority classes difficult \cite{zubiaga16}. Further the majority of current research in this area has been targeted towards Twitter and the English language. While some related work has been carried out for other languages as well \cite{AnnotateItalian_TW-BS,gimenez17}, there is a lot left to be done for most languages. One of these is the Danish language, for which to our knowledge no research within rumour stance and veracity classification exists. General research in related areas, as well as approaches for Danish NLP exist, such as part-of-speech tagging and sentiment analysis \cite{nielsen19}. \\

Current research on the subject has put a lot of work into creating and extending datasets \cite{shu2017fake}. Labelled datasets from microblog platforms facilitate stance classification, which can be utilised for automatic generation of crowd stance for rumour veracity prediction \cite{dungs18}. Existing approaches and methods can be re-applied for Danish, however non-Danish data is not applicable. In other words, a labelled Danish dataset is needed in order to facilitate Danish rumour stance and veracity classification. 




The process of reaching the goals of this project are incremental and each relies on the former goal being met. Following the common approach, as introduced in section \ref{background:system_architecture}, the first step would be to create a stance annotated dataset for the Danish language. The dataset would be used for supervised stance classification and finding data spawned from rumours would be optimal, as this can be used for rumour veracity classification. The dataset would facilitate a stance classification model, which in turn could be used for rumour veracity classification. This calls for the need to gather relevant data from one or several social media platforms, such as Facebook, Twitter, and Reddit. One should be careful, however, and properly address the ``model organism problem" \cite{tufekci14}. 

\subsection{Rumour data}
As introduced in section \ref{background:system_architecture}, the first steps needed to build a dataset is rumour detection and rumour tracking. While rumour detection is very interesting it is difficult to rely on within the limited time of this thesis project. As such, it would be the simpler approach to use a priori rumour data. Then, for rumour tracking, a mechanism is needed to collect and filter relevant posts discussing the rumours. While one would setup some live monitoring tool for collecting emerging rumour data, a ``tracking" mechanism is not needed for historical data. Instead, one could gather a number of samples related to a rumour, and \textit{then} filter it.

Two particularly big social media platforms are Twitter and Facebook, which could be good choices as sources for the dataset. However, Reddit is also rather big in Denmark, which in contrast to the other two is an anonymous platform. Previous work has identified events known to contain rumours and searched for data on Twitter with keywords matching these events \cite{zubiaga16}. 
Another approach could be to fetch data matching predefined keywords unrelated to specific events, and then sequentially go through it to identify which would be relevant, i.e. rumourous. Social media platforms all have different structures and ways of defining subjects and conversations. 

The task of selecting rumours is difficult, as the data gathered might consist of both rumour and non-rumour data. Thus, one should have a framework for filtering this data. While non-rumour data can still be used to train a stance classifier, rumour data is needed for the purpose of veracity classification.

\subsubsection{Choosing sources}
\label{problem:choosing-sources}
\begin{figure}[h]
    \centering
    \includegraphics[width=0.5\textwidth]{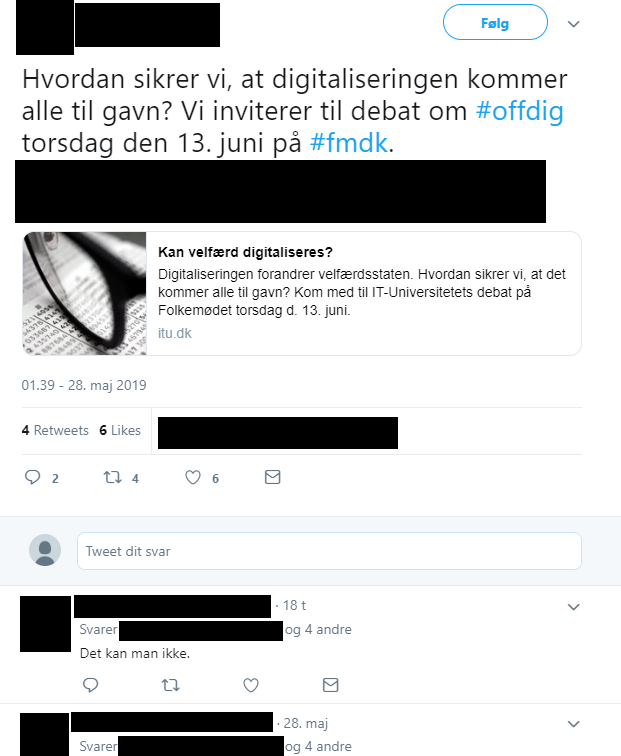}
    \caption{Example of a tweet on Twitter}
    \label{fig:twitter_platform}
\end{figure}

Twitter is a social media platform widely used as a data source for stance and rumour classification datasets \cite{qazvinian11,castillo11,zubiaga16}. Twitter is mainly based on short messages (once 140 characters, now 280\footnote{\label{fn:text_length}\url{https://sproutsocial.com/insights/social-media-character-counter} 24-05-2019}), which users can post/``tweet". Users are able to re-tweet tweets from other users, thereby sharing them and commenting on the initial tweet. This structure allows for conversations to spread and possibly spread rumours. Furthermore ``hashtags"(\#) are used to group similar content by including relevant keywords in the tweet, such as \#dkpol for a tweet concerned with Danish politics. Additionally people can refer to each other by including an `@' sign, followed by a user name. One can imagine that these properties facilitate much networking on Twitter. An example tweet from the Twitter platform is illustrated in Figure \ref{fig:twitter_platform}. In this example we see the source tweet in the top, including two hashtags, and an attachment. It is re-tweeted 4 times, and we see one direct reply and one nested reply.

\begin{figure}[h]
    \centering
    \includegraphics[width=0.5\textwidth]{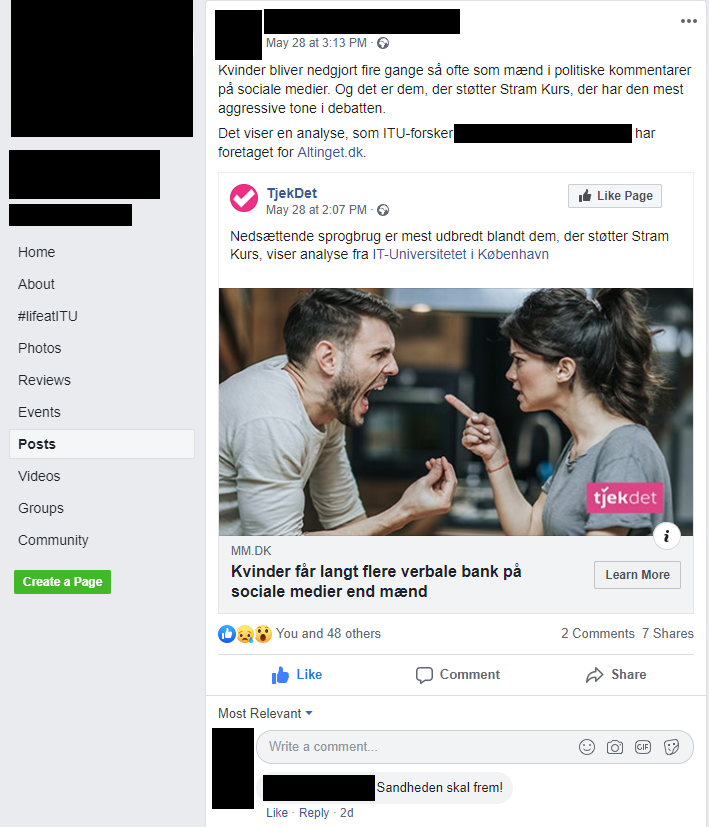}
    \caption{Example of a post on Facebook}
    \label{fig:facebook_platform}
\end{figure}

Facebook is another social media platform which has a posting mechanism resembling Twitter's and have additional concepts including pages, groups, personal albums, and live-chat. Just like Twitter, people can share other people's posts (like re-tweeting), as well as tag people and include hashtags. One major difference is the post character limit, which is 63,206 characters\footref{fn:text_length}. Figure \ref{fig:facebook_platform} illustrates an example post from a public profile on Facebook. This post has two replying comments and has been shared seven times. To our knowledge it is less common to see the use of hashtags on Facebook than on Twitter, but attachments such as links and photos appear frequently.

Both Twitter and Facebook are prominent candidates for a data source. In particular they seem compatible for a cross-platform dataset, in the sense that they share the same conversation structures and mechanisms. However, restrictions on both on them make it difficult to use either for our purpose. Twitter does not allow you to (freely) search for tweets older than 7 days\footnote{\url{https://developer.twitter.com/en/docs/tweets/search/overview} 24-05-2019}, which is very restrictive for finding rumourous data. Third-party libraries such as \texttt{twint}\footnote{\url{https://github.com/twintproject/twint}} does exist however, which circumvent the restrictive access through the API by scraping raw data from the website. There seems to be problems with gathering full conversation trees in a simple way with this approach though. A similar problem exist for Facebook, where one \textit{could} scrape public data, but all Facebook data is private unless made explicitly public, and even public data seems to require special permission\footnote{\url{https://developers.facebook.com/docs/public_feed/} 24-05-2019}.

Thus we turn to Reddit\footnote{\url{https://www.reddit.com/}}, which is a forum website with a fully available API\footnote{\url{https://www.reddit.com/dev/api}}. The site hosts smaller forums or communities in the form of ``Subreddits". A Subreddit typically has a specific theme or entity as the central point of relevance. Users can create posts, called ``submissions", with relevance to the specific Subreddit, where other users can post comments and engage in discussion. Users can downvote or upvote both submissions and other comments. The votes dictate the score of the comment or submission and is used to determine visibility. A low score can hide content such that users must click on it to see it, while a high score may rise the content in relevance and position on the forum. An important thing to note is the anonymity of the platform. While users are anonymous, they are still uniquely identified by their usernames. Additionally the character length for a post can be up to 10,000 characters\footnote{We could not find a source for this, but it is claimed in several Reddit submissions}. 

\begin{figure}[h]
    \centering
    \includegraphics[width=0.7\textwidth]{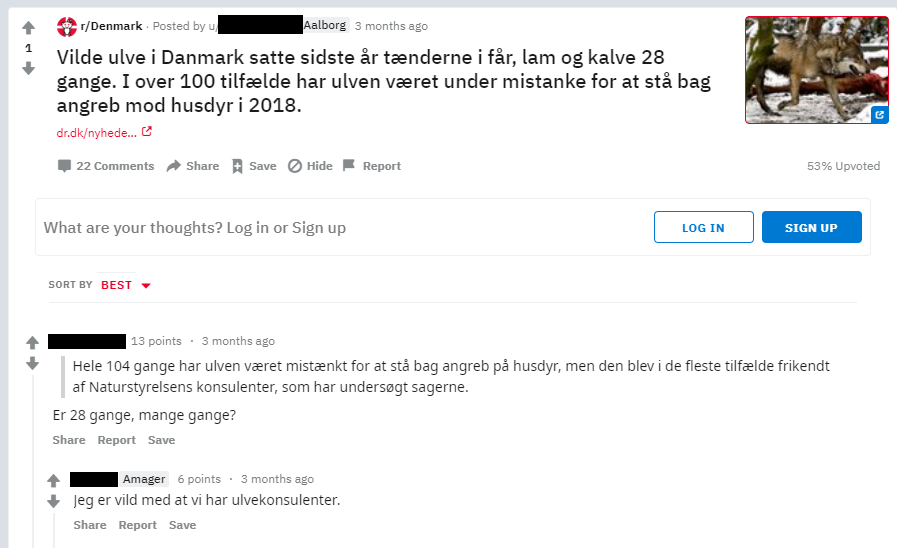}
    \caption{Example of a submission on Reddit with no post text, but a URL reference and an image attached}
    \label{fig:reddit_platform}
\end{figure}

An example of a submission from Reddit is illustrated in Figure \ref{fig:reddit_platform}. In this example the submission post contains a title, an image, and a URL reference, but no post text. It has a total of 22 comments and we see one top-level comment, and a nested reply.

\begin{figure}[h]
    \centering
    \includegraphics[width=0.7\textwidth]{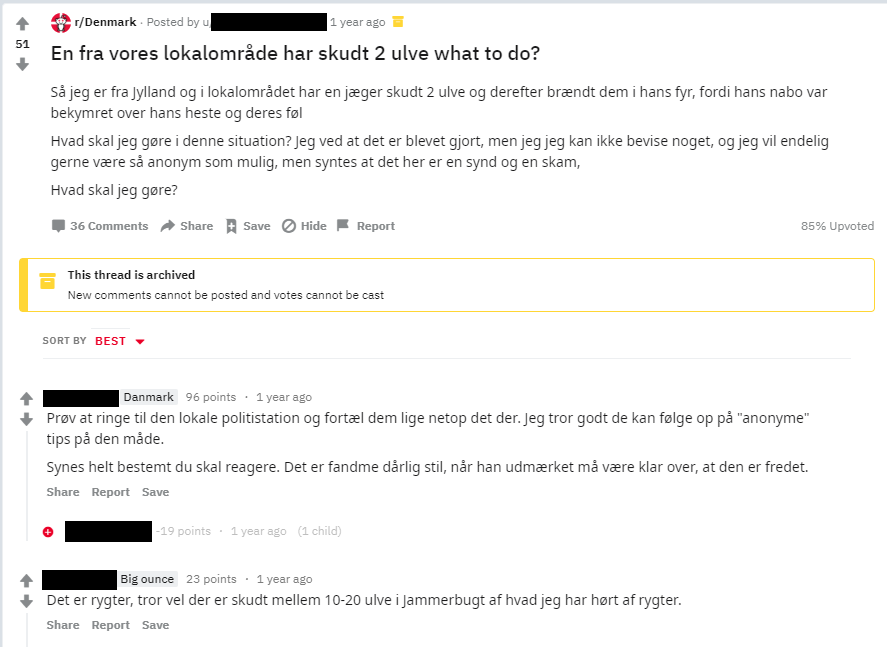}
    \caption{Example of a submission on Reddit with post text}
    \label{fig:reddit_platform2}
\end{figure}

Further, Figure \ref{fig:reddit_platform2} illustrates another Reddit example, where we do see a submission post text.

\clearpage
In both examples (Figures \ref{fig:reddit_platform} and \ref{fig:reddit_platform2}) we see that each comment has a score. In particular we see a negative score in Figure \ref{fig:reddit_platform2}, for the reply to the top-level comment, meaning that Reddit users have downvoted this comment. \\

Given time, data from another source should be included in the research, in order to deal with the ``model organism problem" \cite{tufekci14}. That is, using a single platform as source throughout research might have consequences such as bias, leaving out relevant information from other platforms, as well as human behaviour based on difference in self-awareness (e.g. anonymity vs public) \cite[4.5.1]{thesis-prep}. Most related work use Twitter, which makes the use of Reddit rather novel in this sense (it is also used in this year's RumourEval \cite{semeval_2019}).

\subsubsection{Platform structures}
\label{problem:platform_structures}
Section \ref{problem:choosing-sources} has compared three major social media platform including Twitter, Facebook, and Reddit. This section will further investigate the conversational structures implemented for each of the platforms, which may prove useful and important for choosing either of them as source(s) for the dataset.

\begin{figure}[h]
    \centering
    \begin{minipage}{.5\textwidth}
        \centering
        \includegraphics[width=0.9\textwidth]{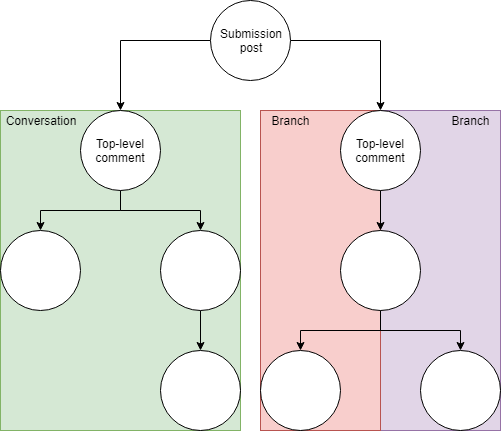}
    \end{minipage}%
    \begin{minipage}{0.5\textwidth}
        \centering
        \includegraphics[width=0.9\textwidth]{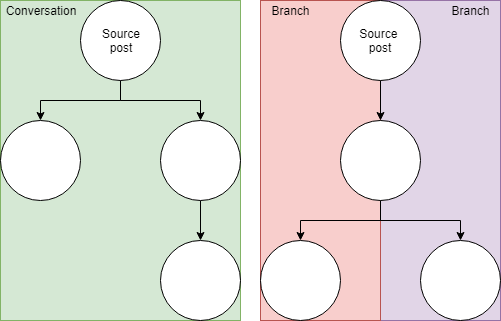}
    \end{minipage}
    \caption{Reddit structure (left) and Twitter/Facebook structure (right). Conversations are coloured green, while two individual branches are coloured respectively red and purple for each platform structure.}
    \label{fig:platform_structures}
\end{figure}

There is one big difference between Reddit and Twitter/Facebook in the way they are structured. Figure \ref{fig:platform_structures} demonstrates the way Reddit is structured (on the left) and how both Twitter and Facebook are structured (on the right). 

The Reddit structure comprises of a submission post, which contains at least a title text, which can spawn several top-level comments. These top-level comments, together with their respective nested replies, make up a \textit{conversation} (coloured green in Figure \ref{fig:platform_structures}). Further, within a conversation, at least one \textit{branch} is present, which is a sequence of replies from one comment with no replies to a top-level comment (two branches are coloured in respectively red and purple for each platform structure in Figure \ref{fig:platform_structures}). Twitter and Facebook implement the exact same structure with regards to conversations and branches, except that each conversation is isolated, whereas several conversations on Reddit are related to the submission post. That is, for Reddit, the concept of a \textit{source} post is actually the submission post, whereas it is the respective top-level posts for conversations on Twitter and Facebook. Note that branches within the same conversation share at least one post. 

\subsubsection{Data annotation}
\label{prob:data_annotation}
\begin{figure}[h]
    \centering
    \includegraphics[width=\textwidth]{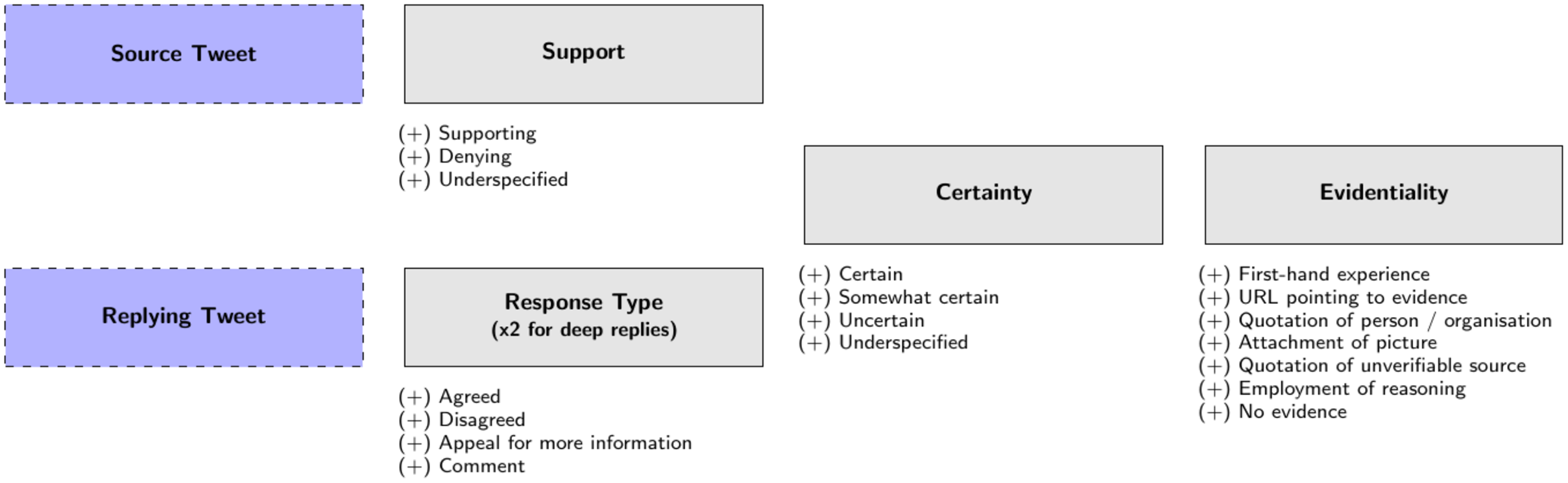}
    \caption{Annotation scheme from \cite{zubiaga16}}
    \label{fig:annotation_scheme}
\end{figure}

For the rumour stance classification component, the gathered dataset needs to be annotated, such that the classification model can perform supervised learning. Previous work has investigated how a dataset could be annotated such that the idea of a crowd stance could be used to either confirm or refute a rumour or \fake{} \cite{procter13}. A popular annotation scheme is defined as having the following purpose: ``[..] an annotation scheme suitable for capturing conversation properties of the Twitter threads in terms of such interactions and used it to obtain an annotated corpus using crowdsourcing"\cite[p. 8]{zubiaga16}. This annotation scheme is depicted in Figure \ref{fig:annotation_scheme} and illustrates how to annotate stance for respectively a source tweet and a replying tweet. Additionally a tweet is annotated for certainty and evidentiality. The certainty of a post describes how certain the author of the post is in their argument. The evidentiality is marked as what, if any, evidence the author of the post uses to support their argument. These two dimensions are used in \cite{zubiaga16} to discover correlations between certainty, evidentiality and rumour veracity.\\ 

The overall idea for the annotation scheme in \cite{zubiaga16} is to label replying posts with one of four classes: ``\textbf{S}upporting", ``\textbf{D}enying", ``\textbf{Q}uerying", and ``\textbf{C}ommenting" (SDQC). These denote the stance of a post towards some statement, which could initiate a rumour. Out of these four classes, the former two are the polarised ones, indicating whether one is for or against the statement. The latter two are neutral, in the sense that a query is asking for more information and ``commenting" is a label for a remark which expresses no opinion. 

The SDQC approach is based on properties identified in Twitter conversations that make it possible to detect rumours \cite{zubiaga16}. These properties include: (1) the phenomena that it takes at least two conversational turns to identify a rumour, (2) posts/comments on Twitter is sequentially ordered by time, (3) a Twitter conversation involves itself with a specific topic, and (4) that underlying features of a response post make up its content. Although the focus is on Twitter, the research is based on analysis of microblogs in general. This is important to consider when annotating data, since it may be structured differently, which may mean that this annotation scheme is not completely applicable, or at least not designed for that specific purpose. \\

For the case of Reddit, it seems that the characteristics of a Twitter conversations mentioned above also apply: a submission includes sequential turn taking regarding a specific topic, and; the production of the response posts is defined by its characteristics. A few things, however, are worth mentioning as differences between these platforms. It is common for a Reddit post to include only a URL pointing to a news event from e.g. \href{https://www.dr.dk/}{\texttt{dr.dk}} (see Figure \ref{fig:reddit_platform}) and that a post can contain up to 10,000 characters. The extra text space may make posts more nuanced, but less precise, which may influence the task of annotating stance. In contrast to Reddit, tweets on Twitter usually contain more than just references to news articles and has a limit of only 280 characters. 

As such, how a conversation is initiated can be different, which may impact how rumours are actually started. This is also mentioned in section \ref{problem:platform_structures}, which illustrates how conversations are structured differently for respectively Reddit and Twitter/Facebook. Thus, for annotating Reddit, one should decide how to follow the \cite{zubiaga16} annotation scheme, which is based on Twitter. This could be done in two ways: (1) is to regard the submission post as a source, and (2) is to regard each top-level comment as source for each conversation. In the first case, a whole submission is treated as equivalent to a Twitter conversation, while the latter follows the same structure. The problem with option (2) is that the top-level comments actually are responses to the Reddit submission post, and as such not really the sources of the conversations. This makes option (1) seem like the more reasonable choice.

\subsection{Stance and veracity classification}
\label{problem_stance_veracity}
The third and fourth steps to build a rumour veracity system are stance and veracity classification (see section \ref{background:system_architecture}). Stance classification is the task of classifying a piece of text to determine how it is oriented towards the veracity of a claim \cite{zubiaga18}. The task has been widely researched for the English language on Twitter \cite{derczynski17,semeval_2019}. As mentioned in section \ref{background:rumour_stance_classification} several different approaches have been applied to handle the problem of stance classification. The Branch-LSTM approach used in \cite{kochkina17} is interesting given the explicit focus on conversation branches. However the size of the dataset to be generated in this study (given the time) might not be sufficient for a deep learning approach, such as the LSTM, which requires a large dataset (further discussed in section \ref{stance_lstm_classifier}). As shown in \cite{AnnotateItalian_TW-BS} the process of annotating a large dataset can be difficult and time consuming. The fact that only two non-experts are available for the annotation task in this study could have an effect on the size of the dataset. As such a ``non-deep learning" approach and careful feature engineering, might facilitate good results even on a smaller dataset \cite{aker17}.

The findings for rumour veracity classification with crowd stance in \cite{dungs18} provide a strong motivation to follow this approach. By relinquishing language-specific features and relying solely on stance, a multilingual dataset might be applicable for rumour veracity classification. If the sequence of stance labels are successfully applicable across different languages, the amount of training data would be greatly increased, which could strengthen Danish veracity classification. Furthermore this makes it possible to avoid time-consuming feature extraction and model selection experiments, such as in \cite{enayet17} and \cite{butfit:semeval2019}, making it more feasible for this project. With this motivation in mind, the positive results of the Hidden Markov Model approach makes it a desirable choice for the veracity classification of the system.

\subsection{Dealing with \fake{} detection}
Section \ref{ralated_work} introduces \fake{} as a rumour which is intentionally and verifiably false, with malicious intend to manipulate or mislead the reader. \cite[sec. 3.1]{shu2017fake} defines the requirements for \fake{} detection as being: information about the author, the content of some news article and the user engagements which consist of a user, their post and a time stamp.

However to know whether the intend behind a rumour is malicious, information about social context for the author is required. Further the task of rumour veracity classification is mentioned as a very similar task as to \fake{} detection \cite[sec. 5.1]{shu2017fake}. Given the time frame of the project and the added complexity of searching for \fake{} rather than rumours, the \fake{} detection is deemed difficult to reach. The Fake News Challenge regards stance detection as a helpful building block to perform \fake{} detection \cite{fakenewschallenge17,hanselowski18}. As such a first step towards \fake{} detection could be to perform stance detection and further rumour veracity classification. \\

With this analysis of the tasks in order to reach the goal of rumour veracity classification, we dive into the methods and approaches for the system. First, in section \ref{technologies}, we include which technologies are used throughout the project.

%% file: technologies.tex
\section{Technologies}
\label{technologies}
This section briefly describes the different technologies, systems, and frameworks utilised within the thesis project to generate a labelled dataset and program classifiers for stance and veracity. The primary programming language used is Python, except for the annotation tool which is programmed in C\#.

\subsection{Data gathering}
For getting data from Reddit the libraries \texttt{praw}\footnote{\url{https://praw.readthedocs.io/en/latest/} v. 6.0.0. 22-02-2019} and \texttt{psaw}\footnote{\url{https://github.com/dmarx/psaw} v. 0.0.7. 22-02-2019} have been utilised to query the Reddit developer API. The API has facilitated the query and download functionality used in the process described in section \ref{data:gathering}.

\subsection{Annotation tool}
The annotation tool is described in section \ref{annotation:tool}, and is developed as an ASP.NET and C\# website with a MySQL database. These technologies were chosen to support rapid development for the project, as we knew the technologies well already. It is meant as a bare bones tool to facilitate a faster annotation process and highlight annotation conflicts.

\subsection{Machine learning models}
A number of frameworks were used for the machine learning models introduced in section \ref{system_description}, which are listed below.

\subsubsection{Stance classifiers}
For stance classification both traditional machine learning (ML) approaches and deep learning approaches are used. As such different technologies are used to support the different approaches:

\begin{itemize}
    \item \sklearn{} \\
    The Scikit Learn library\footnote{\url{https://scikit-learn.org/}} offers a wide variety of ML models, data mutation and testing functionality. All non-neural network models used for stance classification (introduced in section \ref{stance_ml_classifiers}) are implemented from the \sklearn{} API \cite{scikit-learn}.
    \item \texttt{PyTorch} \\
    PyTorch\footnote{\url{https://pytorch.org/}} is a library supporting development of neural networks. The LSTM deep learning model used for stance classification, introduced in section \ref{stance_lstm_classifier}, is implemented in \texttt{PyTorch}.
    \item \texttt{Google colab}\\
    Google colab\footnote{\url{https://colab.research.google.com/} 10-05-2019} is used for training and experimenting with the LSTM. Colab is essentially an online Jupyter Notebook environment. The solution offers a virtual runtime with access to Google's servers including GPU and TPU machines for 12 hours at a time. It is very useful for training neural network models and it makes large hyper-parameter searches more feasible.
\end{itemize}

\subsubsection{Rumour veracity classifier}
The models used for rumour veracity are programmed with the \texttt{hmmlearn} library\footnote{\url{https://hmmlearn.readthedocs.io/} 10-05-2019}. This library facilitates a number of implementations of Hidden Markov Models with respectively Gaussian and multinomial emissions. \\

With the technologies introduced, section \ref{data} will present the generated dataset used for stance classification and rumour veracity prediction.

%% file: data.tex
\section{Danish stance-annotated Reddit dataset}
\label{data}

This section introduces the \textbf{Da}nish \textbf{st}ance(\dataset{}) dataset on Reddit data, generated and annotated for this project. The dataset is publicly available at figshare \cite{lillie_middelboe_2019}. 

First, section \ref{data:gathering} presents the process of gathering the data and provides an overview of the content of the dataset as well as its volume. Second, section \ref{annotation} describes how the dataset is annotated in addition to providing statistics for the class label distribution following the SDQC annotation scheme (see section \ref{prob:data_annotation}).

\subsection{Gathering the data}
\label{data:gathering}
The data gathering process consists of two  approaches: to manually identify interesting submissions on Reddit, and; to issue queries to the Reddit developer API\footnote{\url{https://www.reddit.com/dev/api/}} on specific topics. An example of a topic could be ``Peter Madsen" referring to the submarine murder case, starting from August, 2017\footnote{\url{https://www.dr.dk/nyheder/tema/ubaadssagen} 26-05-2019}. A query would as such be constructed of the topic ``Peter Madsen" as search text, a time window and a minimum amount of Reddit upvotes. A minimum-upvotes filter is applied to limit the amount of data returned by the query. Moreover the temporal filters are to ensure a certain amount of relevance to the case, specifically \textit{when} the news event initially unfolded. Several submissions prior or subsequent to the given case may match a search term such as ``ubåd" (submarine). The list of queries are included in a CSV format in appendix \ref{app:reddit_queries}.

Information about the Reddit submissions returned by the queries as described above are dumped to a CSV file, with the IDs of the submissions needed for subsequent processing. Submissions are also manually identified, in which case the ID of the given submission is added to the CSV file. The submission IDs are used to download all posts from each submission and save them in a JSON format. The JSON data contains meta information about the posts and the users who wrote the them (see an abbreviated example in appendix \ref{app:extracted_reddit_data}). Events to look for were based on a list of ideas generated from browsing the media and our social network, which is reported (in Danish) in appendix \ref{app:data_event_ideas}. Four Danish Subreddits were browsed, including ``Denmark, denmark2, DKpol, and GammelDansk"\footnote{\url{https://www.reddit.com/r/Denmark/wiki/danish-subreddits} 27-05-2019}, although all relevant data turned out to be from the ``Denmark" Subreddit.

\subsubsection{Overview of Reddit data}
\label{data:overview_data}
Table \ref{tab:events_and_submissions} presents an overview of all the events selected and further annotated (which is further described in section \ref{annotation}). The submissions are grouped into events which they relate to. Furthermore the total number of submissions, branches, and posts are included, to illustrate how much data each event contains.

\begin{table}[h]
    \centering
    \begin{tabular}{lrrr}
        \textit{Event} & \textit{Submissions} & \textit{Branches} & \textit{Posts} \\
        \hline
        5G & 4 & 117 & 273\\
        Donald Trump & 3 & 89 & 246\\
        HPV vaccine & 7 & 122 & 255 \\
        ISIS & 2 & 68 & 169\\
        ``Kost"(diet) & 3 & 165 & 557\\
        MeToo & 1 & 29 & 60\\
        ``Overvågning"(surveillance) & 1 & 121 & 352\\
        Peter Madsen & 3 & 156 & 381\\
        ``Politik"(politics) & 3 & 126 & 323\\
        Togstrejke(train strike) & 2 & 49 & 101 \\
        ``Ulve i DK"(wolves in DK) & 4 & 119 & 290\\
        \hline
        \textit{Total} & 33 & 1,161 & 3,007 \\
        \hline
    \end{tabular}
    \caption{Overview of data events and submissions}
    \label{tab:events_and_submissions}
\end{table}

In total the dataset contains 3,007 Reddit posts distributed across 33 submissions respectively grouped into 16 events. Although the volume of this dataset is considered small when used for classification tasks (discussed further in section \ref{stance_lstm_classifier}), the size seems to be relatively big in comparison to other stance labelled dataset:

\begin{enumerate}
    \itemsep0em
    \item RumourEval 2019 dataset \cite{semeval_2019}: \\ Allegedly the largest stance-annotated dataset to date. SDQC labelled multi-platform (Twitter and Reddit) and multilingual (English, Danish, and Russian) dataset containing at least 297 source tweets and 7100 discussion tweets in English (final
    dataset with remaining data not published yet)
    \item IberEval 2017 dataset \cite{taule17}: \\  Favour/against/none stance (and gender) labelled Twitter dataset for respectively Spanish and Catalan. Each dataset consists of a total of 5,400 tweets, with 4,319 for training and 1,081 for testing
    \item PHEME dataset \cite{pheme-dataset}: \\ SDQC labelled Twitter dataset containing 4,842 tweets across 297 English and 33 German Twitter conversations
    \item Turkish tweets dataset \cite{kucuk17}: \\ Favour/against labelled Twitter dataset containing 700 tweets in Turkish
\end{enumerate}

However, only about half of the data in \dataset{} is annotated as being related to rumours (further described in section \ref{annotation}), which is relevant for the rumour veracity classification task. As a reference, for the PHEME dataset, all 330 conversations are deemed rumourous.


\subsection{Annotation}
\label{annotation}
As introduced in section \ref{prob:data_annotation}, one widely used annotation scheme for stance is the SDQC approach \cite{zubiaga16}. The scheme is depicted in Figure \ref{fig:annotation_scheme}, section \ref{prob:data_annotation}, and illustrates how to annotate stance on Twitter. However, a way to annotate the differently structured Reddit platform is by regarding a submission post as a source, which is discussed in the last part of section \ref{prob:data_annotation}.

In the process of annotating Reddit posts the annotation was initially configured to show the individual stance of the posts towards the possible rumourous event/topic itself (e.g. ``Donald Trump"). However, the intent with the annotation scheme is to more explicitly focus on the source post and the post which the given post is replying to (``parent post"). As such, a mechanism supporting this double-annotation was implemented, such that a post is both analysed with regards to the source post and the parent post instead of directly to the event (this is indicated as ``x2 for deep replies" in Figure \ref{fig:annotation_scheme}, section \ref{prob:data_annotation}). The double-annotation should facilitate a way to infer the stance for individual posts. For instance, if the source post supports a rumour, and a nested reply supports its parent post, which in turn denies the source, then the nested reply is implicitly denying the rumour.

In \cite{zubiaga15} they crowdsource the task of annotation and discusse how the task can be broken into micro-tasks such that annotators can concentrate on the same thing throughout the process. We have chosen to do the annotation ourselves as the crowdsourcing would require time and resources not available for the thesis project. As such we do not break the task into smaller tasks, rather each of us annotate a full post completely, i.e. on the three dimensions being support/response type, certainty, and evidentiality (see Figure \ref{fig:annotation_scheme}, section \ref{prob:data_annotation}). The result is a dataset which is not only annotated for stance, but also contains annotations for the certainty of posts and the evidentiality of posts. While these labels are not used in this study, they might be useful for future work. However, at times we actually found it useful to use these dimensions to reason about annotated stance, when resolving annotation conflicts between us.

Next, in section \ref{annotation:tool}, we describe how we have facilitated the annotation process with a custom built tool, as well as what we did to support and make use of the annotation scheme. 

\subsubsection{Process and tool}
\label{annotation:tool}
In order to support the annotation process we have developed a tool in C\# ASP.NET with a MySQL database. The tool is meant to make the annotation process more efficient and highlight annotation conflicts between annotators. Screenshots of relevant mechanisms in the web-based tool are presented along with their descriptions. \\

\begin{figure}[h]
    \centering
    \includegraphics[width=\textwidth]{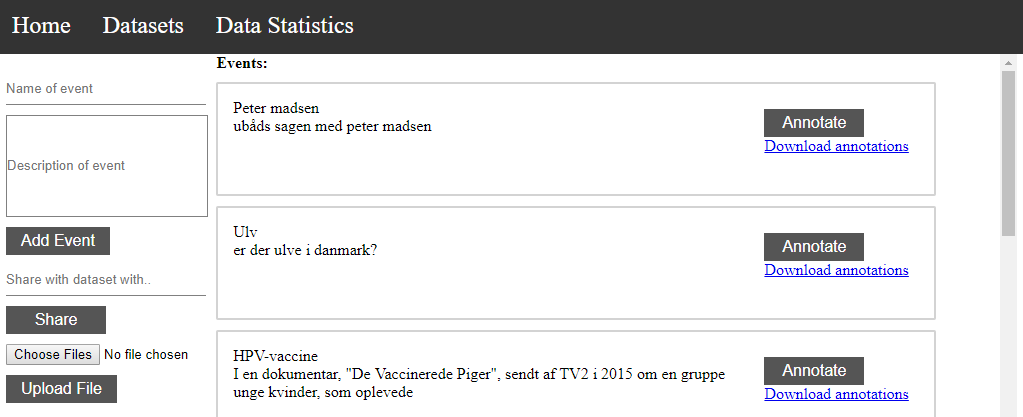}
    \caption{Overview of the events to annotate}
    \label{fig:event_annotate_overview}
\end{figure}

The tool supports separation of data, in that the user can create datasets, which can contain a number of events, each of which contains a number of submissions (source posts). This is illustrated in Figure \ref{fig:event_annotate_overview}, with the list of events to annotate on the right and utilities for creating new events and uploading data on the left. 

An event is meant to separate data from different topics and events, such that only submissions in the same event are displayed while annotating. In each event the generated JSON data from Reddit (see section \ref{data:gathering}) can be uploaded with the ``Choose Files" and "Upload File" buttons to the left in Figure \ref{fig:event_annotate_overview}. 

\begin{figure}[h]
    \centering
    \includegraphics[width=0.85\textwidth]{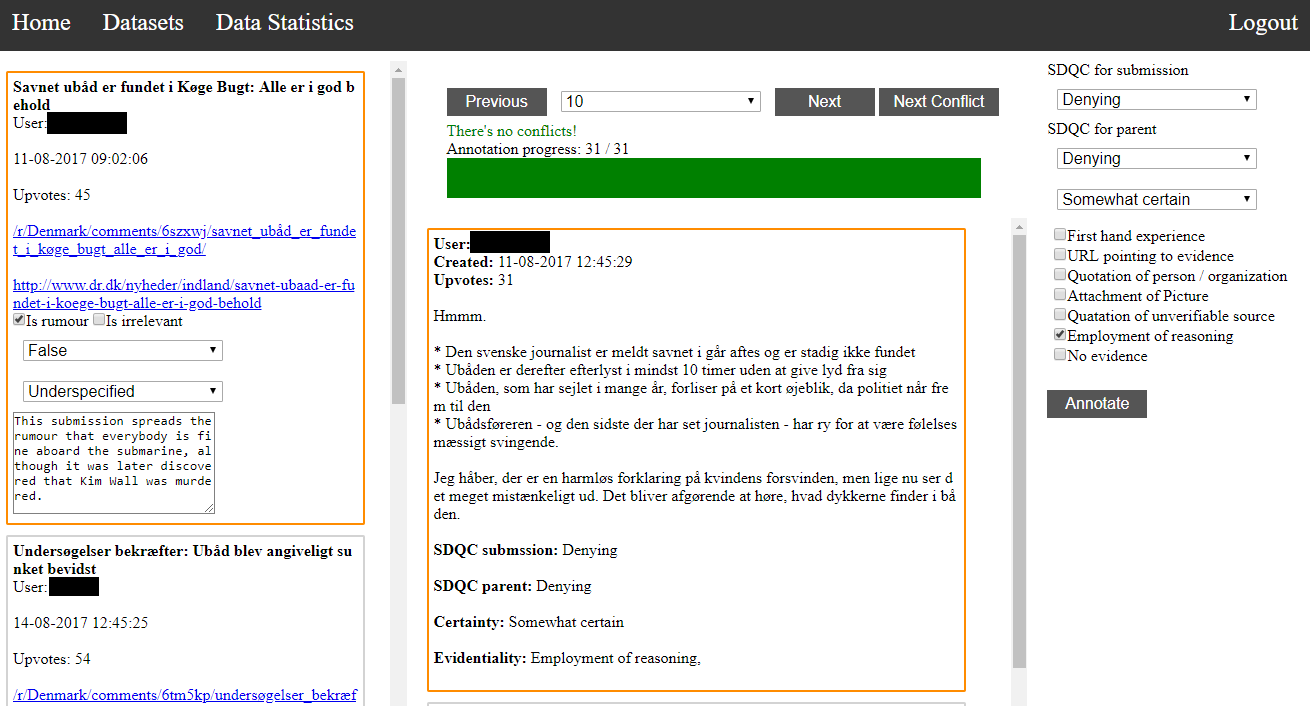}
    \caption{Annotation of a submission}
    \label{fig:submission_annotation}
\end{figure}

To initiate annotation of a submission within an event, one would click on the ``Annotate" button for the given event, after which the annotation page will be loaded, as depicted in Figure \ref{fig:submission_annotation}.

\clearpage
The annotation page in Figure \ref{fig:submission_annotation} supports a view of each submission within the event (on the left) and displays each branch of comments to the annotator (in the middle). The annotator can go through the branches sequentially by pressing the ``Next" button, or go to a specific one from the drop-down menu. The display of a whole branch allows the annotator to consider the context of the text of the individual posts when annotating. 

\begin{figure}[h]
    \centering
    \includegraphics[width=\textwidth]{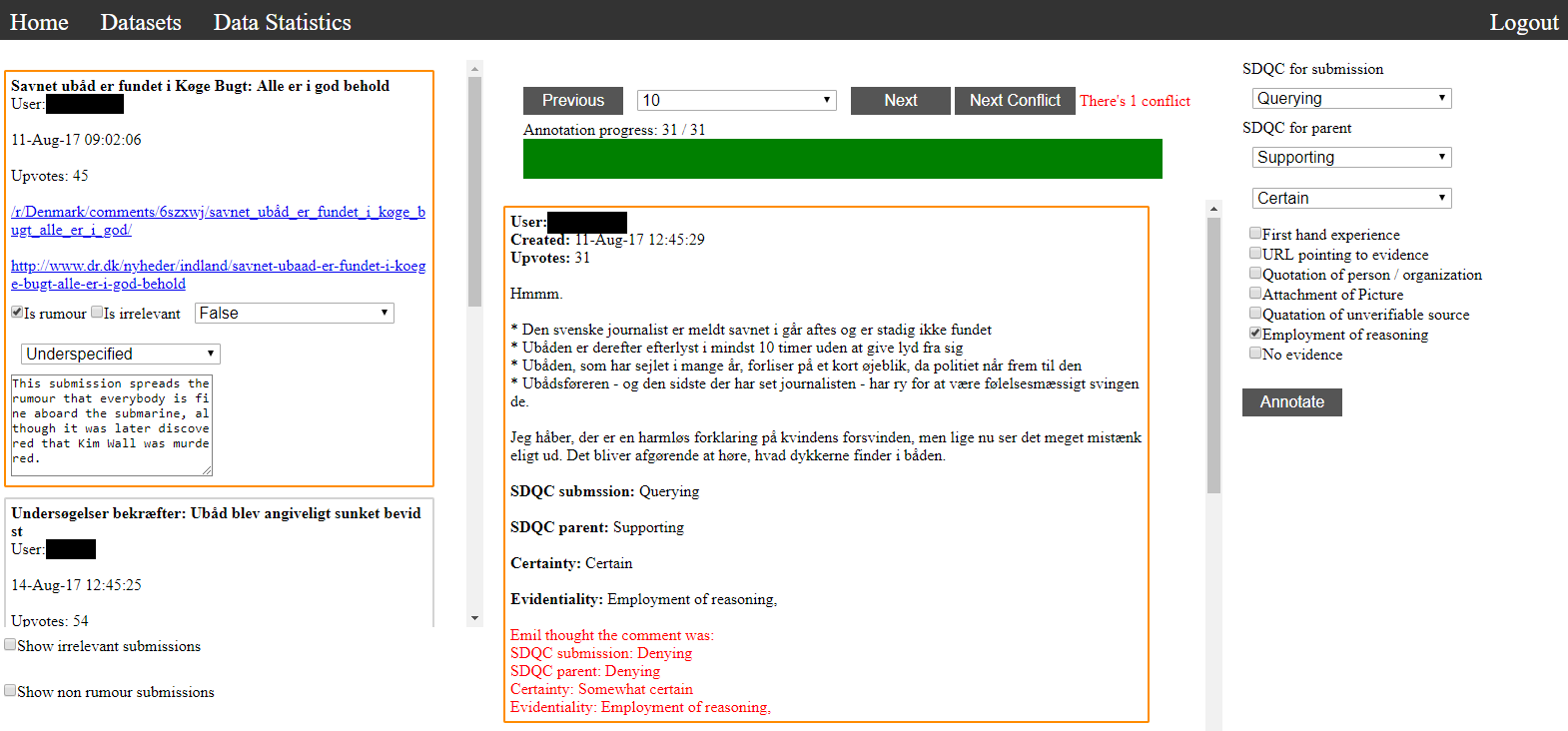}
    \caption{Resolving annotation conflicts}
    \label{fig:annotation_conflict}
\end{figure}

The annotator can annotate the selected post according to the annotation scheme with the options menu on the right of the annotation page. The number of conflicts between the annotators, if any, are also displayed in the top, with an option to go to/display the given post when clicking ``Next Conflict". An example is illustrated in Figure \ref{fig:annotation_conflict} where the annotation choices of the other annotator is displayed with text in red. For resolving such a conflict, we would sit down and discuss the given post and re-annotate it accordingly. 

Once satisfied with the annotations, they can be downloaded from the event overview page (see Figure \ref{fig:event_annotate_overview}), where only posts with no annotation conflicts are included in the downloaded file(s). The output is similar to the extracted Reddit data (see appendix \ref{app:extracted_reddit_data}), only with the following annotations appended: whether the submission is a rumour and its truth status, as well as SDQC labelled replying posts. Additionally a note on the submission is included, which is meant to describe what the statement and case is about in the submission itself (to the left in Figure \ref{fig:annotation_conflict}). \\

The stance of the source/submission post is taken into account when annotating the stance for replying posts of top-level posts (denoted ``SDQC for parent" in the annotation tool). As stance annotations are relative to some target, each post does not have one single stance annotation: each post is annotated for the stance targeted towards the submission and the stance targeted towards the direct parent of the post (also introduced in the beginning of section \ref{annotation}).

Further, a majority of submissions have no text, but a title and a link to an article, image or another website, with content related to the title of the submission. If this is the case and the title of the submission bears no significant stance, it is assumed that the author of the submission takes the same stance as the content which is attached to the submission. 

\begin{figure}[h]
    \centering
    \includegraphics[width=\textwidth]{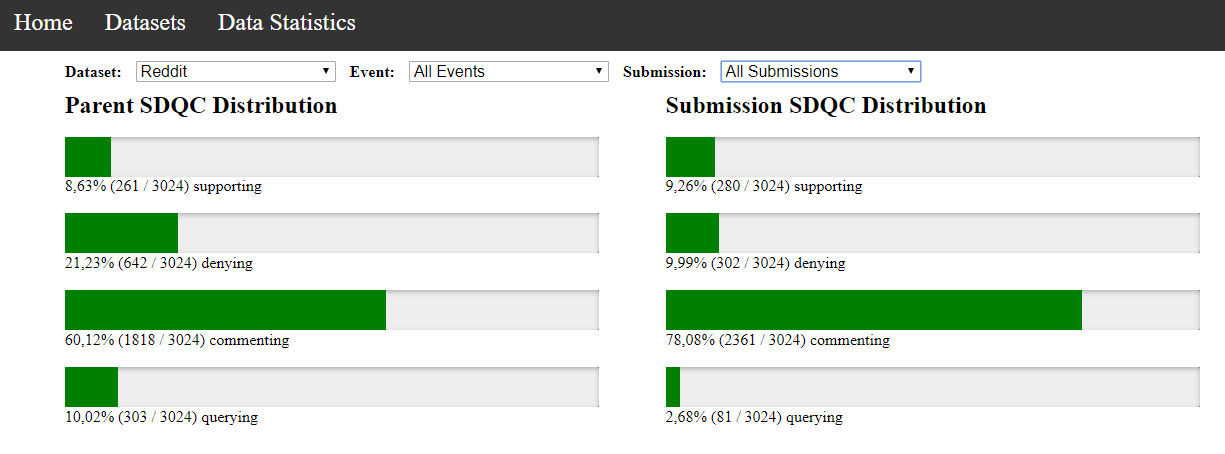}
    \caption{Annotation statistics (the numbers do not precisely reflect the final dataset, \dataset{})}
    \label{fig:annotation_stats}
\end{figure}

As the annotation progressed the tool also provided a way for us to track the distribution of SDQC annotations for respectively the submission and parent posts, as described above. This is illustrated in Figure \ref{fig:annotation_stats}, however note that these numbers do not reflect the final dataset, as some posts were deemed invalid (17 posts to be precise). The results of the annotations are presented next, in section \ref{annotated_dataset_overview}.

\subsubsection{Annotated dataset}
\label{annotated_dataset_overview}
This section will present an overview of the annotated dataset, including SDQC labels per event and the overall distribution.

The dataset, as presented in section \ref{data:overview_data}, contains 11 events, including 3,007 posts across 33 submissions, which have all been annotated for SDQC stance. Table \ref{tab:events_and_submissions_annotated} provides an overview of the class label distribution per event. 

\begin{table}[h]
    \centering
    \begin{tabular}{|l|r|r|r|r||r|}
        \hline
        \diagbox{\textit{Event}}{\textit{Label}} & S & D & Q & C & \textit{Total}\\
        \hline
        5G & 26 & 47 & 7 & 193 & 273\\
        \hline
        Donald Trump & 39 & 17 & 5 & 185 & 246 \\
        \hline
        HPV vaccine & 24 & 4 & 8 & 219 & 255\\
        \hline
        ISIS & 3 & 40 & 8 & 118 & 169\\
        \hline
        ``Kost"(diet) & 50 & 56 & 4 & 447 & 557 \\
        \hline
        MeToo & 1 & 8 & 3 & 48 & 60\\
        \hline
        ``Overvågning"(surveillance) & 41 & 20 & 13 & 278 & 352\\
        \hline
        Peter Madsen & 15 & 45 & 19 & 302 & 381\\
        \hline
        ``Politik"(politics) & 43 & 46 & 7 & 227 & 323\\
        \hline
        ``Togstrejke"(train strike) & 8 & 6 & 3 & 84 & 101\\
        \hline
        ``Ulve i DK"(wolves in DK) & 23 & 11 & 4 & 252 & 290\\
        \hline \hline
        \textit{Total} & 273 & 300 & 81 & 2,353 & 3,007 \\
        \hline
    \end{tabular}
    \caption{Overview of SDQC stance labels per event}
    \label{tab:events_and_submissions_annotated}
\end{table}

``Kost" is the dominating event with its 557 posts, whereas three events have between 300-400 posts, four events have 200-300 posts, two events have 100-200 posts, and ``MeToo" only has 60 posts. The ``querying" label is rare with a total of 81 annotations out of the 3,007 posts. The ``supporting" and ``denying" labels are almost equally distributed with a total of respectively 273 ``supporting" and ``300" denying posts. The ``commenting" class is the absolute dominant one, with a total of 2,353 annotations. While the ``commenting" class label is consistently the majority class for all of the events, there is variation with regards to the SDQ class labels within each event. Notably \q{MeToo} and \q{ISIS} have a very low amount of ``supporting" labels relative to the other events. \\

Table \ref{tab:dataset_sdqc} illustrates the relative SDQC distribtion for the whole dataset for both response types, being targeted towards respectively submission (source) and parent posts, i.e. the posts replied to. The two upper rows contain the actual numerical distributions and the two bottom rows contain the relative distribution. \\

\begin{table}[h]
    \centering
    \begin{tabular}{|l|r|r|r|r|}
    \hline
    \diagbox{\textit{Target}}{\textit{Label}} & S & D & Q & C \\
    \hline \hline
    Reddit submission post & 273 & 300 & 81 & 2,353 \\ \hline
    Reddit parent comment & 261 & 632 & 304 & 1,810 \\ \hline \hline
    Reddit submission post \% & 9.1 & 10 & 2.7 & 78.2 \\ \hline 
    Reddit parent comment \% & 8.7 & 21 & 10.1 & 60.2 \\ \hline
    \end{tabular}
    \caption{Relative SDQC stance label distribution for \dataset{} with regards to the source, being a ``submission post", and the post replied to, being a ``parent comment"}
    \label{tab:dataset_sdqc}
\end{table}

Finally the dataset is also annotated for rumours, and these as being either true, false or unverified. A total of 16 submissions were deemed as rumours, that is, the source post in each of these submissions initiates some rumourous statement, which spawns one or more \textit{conversations}. Each conversation has one or more \textit{branches}, being a sequence of nested replies from a comment with no replies until the top-level comment. The conversation structure is also described in Figure \ref{fig:platform_structures}, section \ref{problem:platform_structures}. The submissions are contained within nine of the events, and are listed in Table \ref{tab:submission_rumours}, including their title and annotated truth status. \\

\begin{table}[h]
    \centering
    \small
    \begin{tabular}{l|p{7cm}|c}
        \textit{Event} & \textit{Submission title} & \textit{Rumour status} \\
        \hline
         \multirow{3}{*}{5G} & 5G-teknologien er en miljøtrussel, som bør stoppes & Unverified \\ \cline{2-3}
         &Det er ikke alle, som glæder sig til 5G. & Unverified \\ \cline{2-3}
         &Uffe Elbæk er bekymret over de ``sundhedsmæssige konsekvenser" af 5G-netværket & Unverified \\ \hline
         \multirow{2}{*}{Donald Trump} & Hvorfor må DR skrive sådan noget åbenlyst falsk propaganda? & Unverified \\ \cline{2-3}
         &16-årig blev anholdt for at råbe `fuck Trump' til lovlig demonstration mod Trump & Unverified \\ \hline
         \multirow{2}{*}{ISIS} & 23-årig dansk pige har en dusør på \$1 million på hendes hovede efter at have dræbt mange ISIS militanter & Unverified \\ \cline{2-3}
         &Danish student `who killed 100 ISIS militants has \$1million bounty on her head but is treated as terrorist' (The Mirror) & Unverified \\
         \hline
         \multirow{2}{*}{Kost}&Bjørn Lomborg: Du kan være vegetar af mange gode grunde - men klimaet er ikke en af dem & Unverified \\ \cline{2-3}
         &Professor: Vegansk kost kan skade småbørns vækst & False \\ \hline
         MeToo &Björks FB post om Lars Von Trier (\#MeToo) & Unverified \\ \hline
         \multirow{3}{*}{Peter Madsen} & Savnet ubåd er fundet i Køge Bugt: Alle er i god behold & False\\ \cline{2-3}
         &Undersøgelser bekræfter: Ubåd blev angiveligt sunket bevidst & True \\ \cline{2-3}
         &Peter Madsen: Kim Wall døde i en ulykke på ubåden & False \\ \hline
         Politik & KORRUPT & True \\ \hline
         Togstrejke & De ansatte i DSB melder om arbejdsnedlæggelse 1. april. & True \\ \hline
         Ulve i DK &Den vedholdende konspirationsteori: Har nogen udsat ulve i Nordjylland? & Unverified \\ \hline
    \end{tabular}
    \caption{Overview of the rumour submissions and their veracity status}
    \label{tab:submission_rumours}
\end{table}

Out of the 16 rumourous submissions, three were true, three were false and the rest were unverified. They make up 220 Reddit conversations, or 596 branches, with a total of 1,489 posts, equal to about half of the dataset. The posts are distributed across the nine events as follows: 5G(233), Donald Trump(140), ISIS(169), ``Kost"(324), MeToo(60), Peter Madsen(381), ``Politik"(49), ``Togstrejke"(73), and ``Ulve i DK"(56). Thus ISIS, MeToo, and Peter Madsen are the only events which only contain rumourous conversations. \\

With the introduction and presentation of the Danish stance-labelled Reddit dataset, denoted \dataset{}, generated and annotated, we turn to the task of rumour stance classification and rumour veracity prediction. 

%% file: methods.tex
\section{Methods}
\label{system_description}
This section provides an introduction to the methods and approaches taken in this project to use the \dataset{} dataset described in section \ref{data} for stance classification and ultimately rumour veracity classification. Section \ref{preprocessing} describes how data preprocessing is performed, section \ref{stance_classification} introduces the machine-learning approaches for stance classification, as well as features, and section \ref{rumour_classification} describes the approach taken for rumour veracity classification with Hidden Markov Models.

\input{preprocessing.tex}
\input{methods_stance.tex}
\input{methods_veracity.tex}

%% file: preprocessing.tex
\subsection{Preprocessing of data}
\label{preprocessing}
This section describes how all the annotated data from Reddit (see section \ref{data}) is preprocessed for subsequent use in the stance classification task (section \ref{stance_classification}).

\subsubsection{Data representation}
\label{preprocessing:data_representation}
After the Reddit data has been annotated with the annotation tool as described in section \ref{annotation}, it can be downloaded. The data is downloaded through the annotation tool, which formats each submission into JSON files and groups them by event. Each JSON file is structured in the format sketched below, resembling the format of the extracted Reddit data (see appendix \ref{app:extracted_reddit_data}). All posts within a branch is sorted by its time of creation. 

\begin{verbatim}
{
  "redditSubmission": { ... },
  "branches": [
    [
      {post1},
      {post2},
      ...,
      {postN}
    ],
    ...,
  ]
}
\end{verbatim}

The ``redditSubmission" object contains the following relevant information: id, title, text, creation date and time, number of comments, possible URL referencing an article, URL referencing the submission, upvotes, and user info. Furthermore, this object includes the annotation data, including its stance towards the rumour; supporting, denying, underspecified, and rumour veracity status; true, false, unverified, and an accompanying description of the rumour. The user info mentioned above include the user id, user account creation time and date, as well as Reddit specific properties such as karma count and gold status, Reddit employee status(true/false), and a verified email flag.

The list of branches include nested lists of the posts contained within the branches. Each of these posts contains data similar to the data described above: comment ID, text, parent comment ID, URL, creation time and date, user info, and Reddit specific properties including upvotes and replies, as well as flags such as ``is submitter" and ``is deleted". Each post also has annotation details including SDQC to the rumour/statement introduced in the submission and SDQC with regards to the parent post. 

The download functionality exposed through the annotation tool takes care of cleaning the data, such that deleted posts and non-annotated posts are ignored. Additionally all posts created after a deleted post in a branch are ignored, as keeping them would otherwise break the natural flow of conversation with the valid posts above the deleted one.

\subsubsection{Data preprocessing}
\label{preprocessing_norm}
Once the data is downloaded it can easily be loaded and represented as Python objects with the JSON decoder\footnote{\url{https://docs.python.org/3/library/json.html}}, keeping the structure from the JSON files. As a preprocessing step, all post texts are lower-cased and then tokenised with the NLTK library \cite{nltk_book}, and finally all punctuation is removed, not including cases such as commas and periods in numbers, as well as periods in abbreviations. Furthermore URLs are replaced with the tag ``urlurlurl" and quotes with the tag ``refrefref". 

Throughout loading and preprocessing of the data, minimum and maximum values are recorded for discrete values in the data, in order to allow normalisation in the feature extraction step (see section \ref{features}).

It is important to note that the source post, i.e. the submission post, is not subsequently used in feature extraction and classification. This is because it is separate from the other posts in a submission, unlike e.g. conversations on Twitter (see discussion in sections \ref{problem:platform_structures} and \ref{prob:data_annotation}). It is only used for annotation of top-level comments in Reddit conversations, its rumour veracity status, as well as the cosine similarity between that and other posts (further explained in section \ref{features}). \\

With the \dataset{} data preprocessed, feature extraction can be performed and subsequently work as input to a stance classifier, which is described next, in section \ref{stance_classification}. 

%% file: methods_stance.tex
\subsection{Rumour stance classification}
\label{stance_classification}
This section presents the methods applied for rumour stance classification on the preprocessed \dataset{} data.

First, the scoring metrics used for the experiments in this project are introduced in section \ref{scoring_metrics}. A variety of Machine Learning (ML) models are utilised to perform stance classification, which are described next in sections \ref{stance_lstm_classifier} and \ref{stance_ml_classifiers}. As input, these ML models require feature vectors. A description of the features generated from feature extraction of the preprocessed \dataset{} is presented in section \ref{features}, followed by a feature vector overview in section \ref{feature_vector_overview}.


\subsubsection{Scoring metrics}
\label{scoring_metrics}
Most of the related work report results with accuracy as scoring metric \cite{derczynski17, aker17}, which expresses the ratio of number of correct predictions to the total number of input samples. However, this becomes quite uninteresting if the input samples have imbalanced class distributions, which is the case for our dataset (see section \ref{annotated_dataset_overview}). What \textit{is} interesting to measure is how well the models are at predicting the correct class labels. As such, in addition to reporting accuracy we will also use the $F_1$ scoring metric, which is the harmonic mean between precision and recall. In particular we will use an unweighted macro-averaged $F_1$ score. 

In order to get an idea of the differences between the scoring metrics, we will briefly describe what they stand for as defined in \cite[8.5.1]{data_mining_book}. The goal with the metrics is to evaluate classification performance. That is, we have labelled data, being \dataset{}, and we want to be able to classify unseen data, such that it is classified with the correct class label. In our case however, knowing if the classification results on unseen data really \textit{is} correct can be difficult to determine, but we investigate this further with an example in section \ref{veracity_example}.

For classification, four terms are typically used to denote the outcome of classification, being true positives(TP), true negatives(TN), false positives(FP), and false negatives(FN), which are summarised in table \ref{tab:confusion_matrix}. Note that we will present several such confusion matrices throughout the experiments in section \ref{experiments}.

\begin{table}[h]
    \centering
    \begin{tabular}{@{}cc|cc@{}}
\multicolumn{1}{c}{} &\multicolumn{1}{c}{} &\multicolumn{2}{c}{Predicted} \\ 
\multicolumn{1}{c}{} & 
\multicolumn{1}{c|}{} & 
\multicolumn{1}{c}{Yes} & 
\multicolumn{1}{c}{No} \\ 
\cline{2-4}
\multirow[c]{2}{*}{\rotatebox[origin=tr]{90}{Actual}}
& Yes  & \textit{TP} & \textit{FN}   \\[1.5ex]
& No  & \textit{FP}   & \textit{TN} \\ 
\cline{2-4}
\end{tabular}
    \caption{Confusion matrix of true positives(TP), true negatives(TN), false positives(FP), and false negatives(FN)}
    \label{tab:confusion_matrix}
\end{table}

Now, accuracy is defined as the true positives and true negatives out of all predictions, or formally as defined in equation \ref{math:accuracy}:

\begin{align}
    accuracy = \frac{TP+TN}{TP+FN+FP+TN}
    \label{math:accuracy}
\end{align}

In order to understand \fone{}, first we must understand \textit{precision} and \textit{recall}. The former is a measure of exactness, i.e. how many predictions labelled as positives (TP+FP) are actually such. The latter is a measure of completeness, i.e. how many actual positive prediction (TP+FN) are labelled as such. This can be expressed formally as in equations \ref{math:precision} and \ref{math:recall} below:

\begin{align}
    precision = \frac{TP}{TP+FP}
    \label{math:precision}
\end{align}

\begin{align}
    recall = \frac{TP}{TP+FN}
    \label{math:recall}
\end{align}

Precision and recall tend to have an inverse relationship, in that increasing one will come at the cost of reducing the other. This is where the $F$ measure applies, which combines precision and recall in a single measure. A typical $F$ measure is \fone{}, which is defined as:

\begin{align}
    F_1 = \frac{2\times precision\times recall}{precision + recall}
    \label{math:fone}
\end{align}

As mentioned, this is a \textit{harmonic mean}\footnote{\url{https://en.wikipedia.org/wiki/Harmonic_mean} 31-05-2019} of precision and recall, which gives equal weight to each measure. One can also use the $F_\beta$ measure, which assigns $\beta$ times as much weight to recall as to precision, which may be desirable in some cases.

Finally, for multi-class classification, such as for SDQC, we want to include an averaged \fone{} score. In this case we will use the \textit{macro} \fone{} measure\footnote{\url{https://scikit-learn.org/stable/modules/generated/sklearn.metrics.f1_score.html}}, which is an unweighted mean of \fone{} score \textit{per} class label, hence treating all classes equally. In contrast, \textit{micro} \fone{} is calculated as the mean of aggregated contributions for all classes. 

Where appropriate we will also report standard deviation for averaged results, which quantifies the amount of variation in the set of predicted class labels\footnote{\url{https://en.wikipedia.org/wiki/Standard_deviation} 31-05-2019}.

\subsubsection{LSTM classifier}
\label{stance_lstm_classifier}
The LSTM model is widely used for tasks where the sequence of data and earlier elements in sequences are of importance \cite{neural_primer}. The temporal sequence of tweets was one of the motivations for \cite{kochkina17} to use the LSTM model for branches of tweets, as well as for the bidirectional conditional LSTM for \cite{augenstein16}. As such this section introduces the LSTM method and how we propose to use it for stance classification. \\ 

The Long-Short Term Memory deep learning method is a variant of a Recurrent Neural Network (RNN). RNNs allow representing arbitrarily sized structured inputs in a fixed-size vector, while paying attention to the structured properties of the input \cite[p. 46]{neural_primer}. This property can be desirable when dealing with language data, which are sequences of letters, words, and sentences, and is typically the motivation for implementing an RNN architecture. The RNN model works as follows: it takes as input an ordered list of input vectors $x_1,\dots, x_n$ and an initial state vector $s_0$, and returns an ordered list of state vectors $s_1, \dots, s_n$, as well as an ordered list of output vectors $y_1, \dots, y_n$ \cite[p. 46]{neural_primer}. This can formally be described as:

\begin{align}
	RNN(s_0,x_{1:n}) = s_{1:n}, y_{1:n}
\end{align}

The output of an RNN is determined by two abstract functions $R$ and $O$. $R$ is a recursively defined function which computes a new state vector $s_{i}$ from the previous state $s_{i-1}$, and an input vector $x_i$. $O$ is a function which maps a state vector $s_i$ to an output vector $y_i$. $R$ and $O$ are mathematically defined as follows:

\begin{equation}
	\begin{aligned}
	s_i &= R(s_{i-1}, x_i) \\
	y_i &= O(s_i)
	\end{aligned}
\end{equation}

\begin{figure}[h]
    \centering
    \begin{minipage}{.3\textwidth}
        \centering
        \includegraphics[width=\textwidth]{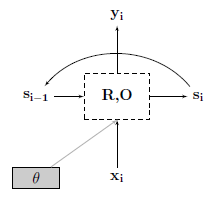}
    \end{minipage}%
    \begin{minipage}{.7\textwidth}
        \centering
        \includegraphics[width=\textwidth]{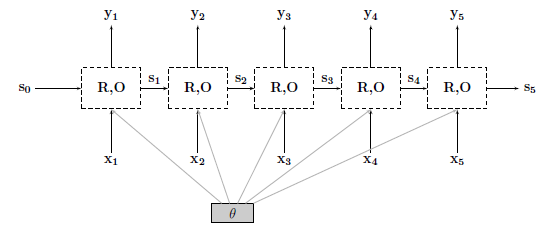}
    \end{minipage}
    \caption{Graphical representation of a recursive definition of an RNN (left) and an unrolled RNN for a fixed-size sequence (right). For time step $i$, $x_i$ denotes the input vector, $s_i$ the state, and $y_i$ the output. \textsc{r} and \textsc{o} are functions, while $\theta$ denotes the parameters, which are shared across all time steps. The unrolled version (right) is an example for a finite sized input sequence ${x_1, \dots, x_5}$. Source: \cite[p. 47]{neural_primer}}
    \label{fig:rnn}
\end{figure}

That is, an RNN state is represented by $s_i$ and $y_i$ after observing the inputs $x_{1:i}$. This means that an RNN state is based on the history of inputs $x_1, \dots, x_i$. In contrast, the Markov assumption describes models, such as the HMM, where future states of a stochastic process depends only upon the present state\footnote{\url{https://en.wikipedia.org/wiki/Markov_property} 01-06-2019}. The RNN model is illustrated in Figure \ref{fig:rnn}, with the recursive definition for arbitrarily sized input on the left, and an example of a fixed sized input on the right.

The abstract functions $R$ and $O$ can be implemented in various ways, thus instantiating different versions of an RNN. The simplest RNN formulation, Simple-RNN (S-RNN) \cite[p. 55]{neural_primer}, computes a state $s_i$ as a linear combination of the input $x_i$ and previous state $s_{i-1}$, passed through a non-linear activation function (typically ReLU), while the output $y_i$ evaluates to $s_i$.

The S-RNN model however suffers from the \textit{vanishing gradients} problem, making it unable to capture long-range dependencies \cite[p. 55]{neural_primer}. The LSTM architecture is designed to solve this by introducing the idea of ``memory cells" to preserve gradients across time. For each input state, a \textit{gate} is used to decide how much of the new input should be written to the memory cell (input gate), and how much of the current content of the memory cell should be forgotten (forget gate). A third and final gate (output gate) determines the output based on the content of the memory cell at a given time. The LSTM architecture is defined formally in \cite[p. 56]{neural_primer} as in equation \ref{math:lstm}: 

\begin{equation}
\label{math:lstm}
\begin{aligned}
s_j = R_{LSTM}(s_{j-1},x_j) &= [c_j;h_j] \\
c_j &= c_{j-1}\odot f+g\odot i \\
h_j &= tanh(c_j)\odot o \\
i &= \sigma(x_jW^{xi} + h_{j-1}W^{hi}) \\
f &= \sigma(x_jW^{xf} + h_{j-1}W^{hf}) \\
o &= \sigma(x_jW^{xo} + h_{j-1}W^{ho}) \\
g &= tanh(x_jW^{xg} + h_{j-1}W^{hg}) \\
y_j = O_{LSTM} &= h_j 
\end{aligned}
\end{equation}

$$s_j \in \R{}^{2\cdot d_h}, x_i \in \R{}^{d_x},$$
$$c_j,h_j,i,f,o,g \in \R{}^{d_h},$$
$$W^{xo} \in \R{}^{d_x\times d_h}, W^{ho} \in \R{}^{d_h\times d_h}$$ \\

The mechanisms of the LSTM as described in \cite[p. 56]{neural_primer} work as follows: 

A new state at time $j$ is composed of two vectors, $c_j$ and $h_j$, where the former is the memory cell and the latter is the output component. The three gates are denoted as respectively \textbf{i}, \textbf{f}, and \textbf{o}, controlling for \textbf{i}nput, \textbf{f}orget, and \textbf{o}utput as described above. The gate values are computed based on linear combinations of the current input $x_j$ and the previous state $h_{j-1}$, passed through a sigmoid activation function ($\sigma$). \textbf{g} is an update candidate, which is computed as a linear combination of $x_j$ and $h_{j-1}$, passed through a tanh activation function. When these four values are computed, the memory cell $c_j$ is updated accordingly, by determining how much of the previous memory to keep ($ c_{j-1}\odot f$) and how much of the proposed memory to keep ($g\odot i$). Note that $\odot$ denotes component-wise product. The output $y_j$ is then the value of $h_j$, which is determined by the value of the updated memory cell passed through a tanh non-linearity and controlled by the output gate, \textbf{o}. \\

The LSTM approach in \cite{augenstein16} was introduced as a \sota{} system in section \ref{background:rumour_stance_classification}, performing very well in an unsupervised target-stance classification task. This approach implements a bidirectional variant of an LSTM, denoted ``Bi-LSTM". While the LSTM as described above would allow us to compute a function of the \textit{i}th word $x_i$ based on the words $x_{1:i}$ in a sentence $x_1, \dots, x_n$, the bidirectional implementation also considers the words $x_{i:1}$ \cite[p. 52]{neural_primer}. Put differently, the input sequence is also considered in reverse order, thus allowing the LSTM to look arbitrarily far at both the past \textit{and} the future of the given input data. The bidirectional architecture works by maintaining respectively a \textit{forward state} and a \textit{backward} state, generated by two different LSTM models. One LSTM receives the input sequence in one order and the other receives it in reverse. The state representation at a given time step is then composed of both the forward \textit{and} the backward states. 

While the results from both the Bi-LSTM in \cite{augenstein16} and Branch-LSTM in \cite{kochkina17} achieves \sota{} performance, they both note that their deep learning approaches suffer from the lack of a larger training dataset. As such we suspect that we would observe the same tendency for the \dataset{} dataset, which is relatively small with its 3,007 Reddit posts (see section \ref{data:overview_data}). However, as the LSTM approach still manages to achieve \sota{} performance, we have opted to include an LSTM implementation for the stance classification task.

Specifically, the LSTM classifier used for stance classification in this project consists of a number of LSTM layers, and a number of ReLU layers followed by a dropout layer and a softmax layer to perform classifications. The configurations considered and overall approach is inspired by the Branch-LSTM classifier in \cite{kochkina17}, except that we do not input data grouped sequentially by branches, but one by one. As such we do not implement any extra features than the LSTM method described in this section.

\subsubsection{Classic classifiers}
\label{stance_ml_classifiers}
The remaining ML approaches utilised are facilitated by the \sklearn{} \cite{scikit-learn} library which provides a wide variety of machine learning implementations. Specifically we implement non-neural network models, and will as such denote these models as either ``\sklearn{} classifiers" or ``classic classifiers". It is the intention to use non-neural network models in contrast to the LSTM deep learning approach above, as research shows that this approach can do very well \cite{derczynski17}, particularly Decision Tree and Random Forest classifiers \cite{aker17}. Furthermore Support Vector Machine (SVM) and Logistic Regression have proven to be efficient \cite{enayet17,derczynski17}. The models are listed below, prefixed with a label, which we will use to denote them throughout the paper: \\

\paragraph{\textit{logit}: Logistic Regression}
fits a logistic function to determine the probability of some label being true given training data\footnote{\url{https://en.wikipedia.org/wiki/Logistic_regression} 02-06-2019}. Given that this is a binary classifier, a logistic regression is made for each label present in the data, and predicts the label with the highest probability of being true. \textit{logit} is implemented with the \texttt{LogisticRegression} model\footnote{\url{https://scikit-learn.org/stable/modules/generated/sklearn.linear\_model.LogisticRegression.html}}.

\paragraph{\textit{tree}: Decision Tree}
classifier builds a tree structure from training data, where the labels are isolated given partitions on the attributes in the training data \cite[p. 330-346]{data_mining_book}. A commonly used strategy for identifying the appropriate attribute to split on is \textit{information gain}. The decision tree will try to choose the split of attributes such that the class label is isolated most efficiently. New data entries choose a path through the tree given the trained attribute splits and the class label in the leaf is the prediction. \textit{tree} is implemented with the \texttt{DecisionTreeClassifier} model\footnote{\url{https://scikit-learn.org/stable/modules/generated/sklearn.tree.DecisionTreeClassifier.html}}.

\paragraph{\textit{svm}: Support Vector Machine} 
searches for the optimal separating hyperplane, a boundary which separates the data of one class label from another \cite[sec. 9.3, p. 408-415]{data_mining_book}. If the input is not linearly separable the SVM applies a non-linear mapping function on the input data to lift it into a higher dimensional space. The mapping into higher dimensions is applied until the data is linearly separable by a hyperplane in the new dimensions. The complexity of the SVM is not correlated to the dimensionality of the data, but the number of support vectors found. This means the SVM does not tend to be prone to overfitting, which is a desired trait given the skewedness of \dataset{}. \textit{svm} is implemented with the \texttt{LinearSVC} model\footnote{\url{https://scikit-learn.org/stable/modules/generated/sklearn.svm.LinearSVC.html}} using a linear kernel.

\paragraph{\textit{rf}: Random Forest}
is an ensemble method in which \textit{k} Decision Trees are build, each given a random subset of the attributes \cite[p.382-383]{data_mining_book}. Given new data the ensemble of trees each vote and the most popular prediction is the output of the Random Forest. The method is robust in that outliers in the data only have a small effect on the model. Further the model is not prone to overfitting as long as the number of trees in the ensemble is large. \textit{rf} is implemented with the \texttt{RandomForestClassifier} model\footnote{\url{https://scikit-learn.org/stable/modules/generated/sklearn.ensemble.RandomForestClassifier.html}}. \\

As baseline models, a simple majority voter as well as a stratified classifier have been used from \sklearn{}\footnote{\url{https://scikit-learn.org/stable/modules/generated/sklearn.dummy.DummyClassifier.html}}. They are defined as:

\paragraph{\textit{mj}: Majority vote}
classifier is a simple classifier, which identifies the most frequently occurring class label in the training set, and always predicts that given label. As such it can achieve deceptively high accuracy scores on skewed datasets.

\paragraph{\textit{sc}: Stratified classification}
generates predictions such that they respect the distribution of class labels in the training data. In other words, this baseline classifier creates random samples respecting the distribution found in the training data. For example given a training set with 90\% 0's and 10\% 1's, the baseline will pick 0 at a 90\% rate and 1 at a 10\% rate when \q{classifying} unseen data.

\subsubsection{Features}
\label{features}
Words and sentences can be difficult for computers to process and understand, therefore a numerical representation of each post is needed. This is achieved through feature extraction, in which properties about the post are represented by numbers. The numerical representation of a post as some vector is much easier to work with and reason about in a computational context.

In order to represent the features of the preprocessed data numerically we employ eight feature categories, which are grouped by how they relate: text, lexicon, sentiment, Reddit, most frequent words, BOW, POS, and word embeddings. Note that only the Reddit specific feature are domain-dependent, while the others should apply for the general case. The choices of features are a compilation of select features from various \sota{} systems \cite{aker17,kochkina17,enayet17}, except for the Reddit specific ones. Most of the features are binary, taking either a 0 or a 1 as value, and those that are not are min-max normalised 
\cite[p. 114]{data_mining_book}, except for the word embeddings.

\paragraph{Text features} are extracted from the syntax of the text in the data and include the following listings:

Binary values:
\begin{itemize}
    \itemsep0em
    \item Presence of period, `.' 
    \item Presence of exclamation mark, `!' 
    \item Presence of question mark `?' or 'hv'-words (`wh' question words, such as `what' and `why')
    \item Presence of three sequential periods, `\dots'
\end{itemize}

Discrete values:
\begin{itemize}
    \itemsep0em
    \item Length of raw text (no preprocessing)
    \item Count of URL references
    \item The maximum length of a capital character sequence
    \item Count of three sequential periods, `...'
    \item Count of question mark, `?'
    \item Count of exclamation mark, `!'
    \item Number of words
\end{itemize}

Continuous values:
\begin{itemize}
    \itemsep0em
    \item Ratio of capital letters to non-capital letters
    \item Average word length
\end{itemize}

\paragraph{Lexicon features} are extracted by looking up occurrences of items in four predefined lexicon/dictionaries: negation words, swear words, positive smileys, and negative smileys. The negation words are translated from the English list used in \cite{kochkina17}, as no list could be found for this purpose elsewhere. The swear words are generated from various sources aside from ourselves: \texttt{youswear.com}\footnote{\url{http://www.youswear.com/index.asp?language=Danish} 06-05-2019}, \texttt{livsstil.tv2.dk}\footnote{\url{http://livsstil.tv2.dk/2017-06-10-taet-oploeb-her-er-brugernes-favorit-bandeord} 06-05-2019}, \texttt{dansk-og-svensk.dk}\footnote{\url{https://www.dansk-og-svensk.dk/danskt_lexikon2/Bandeord/svenske_danske_bandeord.htm} 06-05-2019}, and  \texttt{dagens.dk}\footnote{\url{https://www.dagens.dk/nyheder/se-listen-her-er-de-allervaerste-bandeord} 06-05-2019}. The smiley lists were compiled from Wikipedia using the western style emoticons\footnote{\url{https://en.wikipedia.org/wiki/List_of_emoticons} 25-02-2019}. 



\paragraph{Reddit-specific features} are features, which are specific to the domain of Reddit, and include information about the user who posted a submission or reply to a submission, as well as meta information about the post. For the user features, these include the following, where only the first one is non-binary:
\begin{itemize}
    \itemsep0em
    \item Karma - The score awarded from upvotes on the users comments and submissions. On Reddit this value does not have an upper bound.
    \item Gold Status - Whether the user has gold status\footnote{Paid premium Reddit membership, which can also be awarded to you by the quality of your post. See \url{https://www.dailydot.com/debug/what-is-reddit-gold/} 27-05-2019}
    \item Is Employee - Whether the user is a Reddit employee
    \item Verified Email - Whether the user has a verified email
    \item Is submitter - Whether the user is submitter of the given submission
\end{itemize}

Furthermore the syntax when posting to Reddit allows for enriching the display of the text. For example `$>$' will display the subsequent text as a quote and `**' will enable bold text. Others are more subtle, such as the `/s' tag, which as an unwritten rule marks the comment from the user as intentionally sarcastic. The Reddit specific syntax features include the following, where the first two are binary values:

\begin{itemize}
    \itemsep0em
    \item Sarcasm - Whether the user expresses sarcasm with the `/s' tag
    \item Edited - If the text has been denoted as edited with the `edit:' tag
    \item Quote count - Count of quotes denoted with `$>$'
    \item Reply count - The number of replies to the given post
    \item Upvotes - How many upvotes(or downvotes) a post has received
\end{itemize}


\paragraph{Bag of words} (BOW) is a set of unique words appearing in all the text data, and the features generated from this are binary features of whether the given post include a word from the set or not.

\paragraph{Most frequent words} are extracted as being the \textit{n} most occurring words in posts grouped per annotation class. In order to filter out general frequent words, such as `er' (`is') and `og' (`and'), words that appear in all \textit{n} most frequent words per class are removed. This way the lists more precisely captures words related to each specific class. 

The ``most frequent words" (MFW) feature is implemented such that we consider the 100 most frequent words per class, and \textit{then} filter out words occurring in all four lists, resulting in 33 words per class, and a total of 132 words/features. The filtering is performed in order to remove potential stopwords. 

Considered too many words would make MFW be more similar to BOW, and conversely; considered too few words would make MFW too specific. As such, the choice of generating 33 words per class seemed like a good trade-off, capturing unique words per class, without including the more general stopwords. The generated list of most frequent words per class is included in appendix \ref{appendix:most_frequent_words}. As an example, two of the most occuring words for the ``querying" class are ``hvorfor" and ``hvordan", which is interesting as these are question words. However, we also observe event-specific words such as ``CO2", ``5G", and ``B12", specifically tying some of the most frequent words to \dataset{}.

\paragraph{Sentiment} values are computed with the \texttt{Afinn} library, which is used to perform sentiment analysis on a number of different languages, including Danish \cite{afinn}. It takes as input a piece of text and rates the overall sentiment score of the text, where negative sentiment gives low or negative values and positive sentiment gives higher values. The sentiment score provided by this library on the text of a post is used as a continuous normalised feature.

\paragraph{Part-of-speech} (POS) tags are used to tag a given text and denote with binary features whether the text include each tag from the POS set. The \texttt{polyglot} library is used for the POS tagging, which include support for the Danish language \cite{polyglot:2013:ACL-CoNLL}. The POS tags consists of the following 17 tags: \\

\begin{minipage}{0.5\textwidth}
		\begin{itemize}
        \itemsep0em
        \item[\textbf{ADJ}] adjective
        \item[\textbf{ADP}] adposition
        \item[\textbf{ADV}] adverb
        \item[\textbf{AUX}] auxiliary verb
        \item[\textbf{CONJ}] coordinating conjunction
        \item[\textbf{DET}] determiner
        \item[\textbf{INTJ}] interjection
        \item[\textbf{NOUN}] noun
    \end{itemize}
\end{minipage}
\begin{minipage}{0.5\textwidth}
		\begin{itemize}
		    \itemsep0em
            \item[\textbf{NUM}] numeral
		    \item[\textbf{PART}] particle
		    \item[\textbf{PRON}] pronoun
		    \item[\textbf{PROPN}] proper noun
		    \item[\textbf{PUNCT}] punctuation
		    \item[\textbf{SCONJ}] subordinating conjunction
		    \item[\textbf{SYM}] symbol
		    \item[\textbf{VERB}] verb
		    \item[\textbf{X}] other
		\end{itemize}
\end{minipage}\\

\paragraph{Word embeddings} are a way to represent words as vectors of real numbers, which have a number of benefits. One benefit is the ability to compare and group words with the same meaning, even if the letters and structures of the words are not alike. Finding nearest neighbouring words is possible on the GloVe dataset \cite{pennington2014glove}, where a query for the word \q{frog} among others returns \q{leptodactylidae} and \q{eleutherodactylus}, which are words used to describe certain types of frogs\footnote{\url{https://nlp.stanford.edu/projects/glove/} 30-05-2019}. Even though neither of the words are alike the query word \q{frog}, they refer to the same entity and are used in the same contexts. \\ 

Word embeddings have been employed as an average of word vectors for each word in a text \cite{kochkina17}. Various algorithms for using dense word embeddings for representing the words in a text have been considered. First, pre-trained word embeddings with \texttt{fastText} for the Danish language have been downloaded and used \cite{grave2018learning}. The algorithm is developed by Facebook AI Research and is applicable for text classification, being fast and on par with state of the art \cite{joulin2016bag}. Pre-trained word vectors for 157 languages are distributed from the \texttt{fastText} website\footnote{\url{https://fasttext.cc/docs/en/crawl-vectors.html} 22-02-2019}, which are trained on \textit{Common Crawl}\footnote{\url{http://commoncrawl.org/}} and \textit{Wikipedia}\footnote{\url{https://www.wikipedia.org/}}.

Second, \texttt{word2vec} \cite{mikolov13word2vec} have been used to ``manually" train word embeddings for the Danish language. The algorithm learns vector representations of words by processing a text corpus with either the CBOW or skip-gram model\footnote{\url{https://code.google.com/archive/p/word2vec/} 22-02-2019}. A Danish text corpus has been acquired from ``Det Danske Sprog- og Litteraturselskab" (DSL)\footnote{\url{https://dsl.dk/}}, being their biggest corpus, ``Korpus 2010", consisting of 45 million tokens of written LGP (Language for General Purposes)\footnote{\url{https://korpus.dsl.dk/resources.html}}. The sentences used from the corpus contain no punctuation or uppercase letters, and has subsequently been tokenised and fed into the \texttt{word2vec} algorithm through the \texttt{gensim} Python framework \cite{rehurek_lrec-gensim}. For experimental purposes the corpus has also been used to train fastText vectors from scratch, instead of the pretrained ones. The framework allows to test with different parameters, such as vector lengths, windows sizes, and training algorithms in order to find the optimal settings, but the default ones have been used. Once trained the word embeddings can be saved to disk, such that they can be retrieved more efficiently when used in word representations in the classification task. This avoids the need to train the model all over again. 

Finally the word embeddings model has been further trained on the preprocessed text from the Reddit dataset. A vector length of 300 is used, as in \cite{kochkina17}. In addition to the averaged word vector, three features are computed from the word embeddings, namely cosine similarity to the parent post, source post, and concatenation of branch posts, which has been deemed relevant in other research \cite{kochkina17}.





\subsubsection{Feature vector overview}
\label{feature_vector_overview}
Table \ref{tab:feature_vector} presents an overview of the total feature vector, including a rough categorisation of their meaning as introduced in this section. Note that the word embeddings are actually 300 long, but the extra 3 features are the cosine similarities between different word embeddings with regards to parent, source, and branch word tokens.

\begin{table}[h]
    \centering
    \begin{tabular}{l|r}
        \textit{Category} & \textit{Length} \\
        \hline
        Text & 13 \\
        Lexicon & 4 \\
        Sentiment & 1 \\
        Reddit & 10 \\
        Most frequent words & 132 \\
        BOW & 13,663 \\
        POS & 17 \\
        Word embeddings & 303 \\
        \hline
        \textbf{Total} & 14,143 \\
    \end{tabular}
    \caption{Feature vector overview}
    \label{tab:feature_vector}
\end{table}

\subsubsection{Testing approach}
\label{testing_setup}
Throughout the stance classification experiments conducted in section \ref{experiments} we will make use of some common techniques for testing, which will be briefly introduced here. \\

First off, where applicable we will make use of the concept of splitting the data into a training and test sample, which allows us to evaluate the models on \q{new}, unseen data, and avoid overfitting. For this purpose we will be using the \texttt{train\_test\_split(..)} function from \sklearn{}\footnote{\url{https://scikit-learn.org/stable/modules/generated/sklearn.model_selection.train_test_split.html} 13-05-2019}. The function splits the lists of indexed features vectors and class labels properly according to a given representation of the proportion of the dataset to include in the test split. If nothing else is specified, we are using a test split size of 0.2, which makes up a sample of 602 data points, leaving 2,405 data points in the training sample. Furthermore we use the ``stratify" option in the train-test split, making sure both samples have the same distribution of class labels.

Furthermore, where applicable, we use \textit{k}-fold cross validation(CV) to evaluate generalisation strength of the models. Again, we use the \sklearn{} implementation\footnote{\url{https://scikit-learn.org/stable/modules/generated/sklearn.model_selection.cross_validate.html} 13-05-2019}. In this case we also enforce stratification. If nothing else is specified we employ 5-fold CV.

Finally, in the experiments, if nothing else is stated we use all of the features introduced in section \ref{features}, and the \texttt{word2vec} word vectors as representing word embeddings.

%% file: methods_veracity.tex
\subsection{Rumour veracity classification}
\label{rumour_classification}
As stated in section \ref{problem_analysis} the approach for rumour veracity classification is based on a Hidden Markov Model (HMM). 
This section briefly describes the workings of a HMM and how it is used for the task of rumour veracity classification with crowd stance, following the work in \cite{dungs18}.

\subsubsection{Hidden Markov Models}
\label{methods_hmm}
A Hidden Markov Model is a probabilistic model which works on sequences of input. The model consists of a number of hidden states $S = \{s_1, s_2, .. , s_N\}$ and a number of possible observations $E = \{e_1, e_2, .. , e_M\}$, where \textit{N} is the number of possible hidden states and \textit{M} is the number of possible observations. The \textit{transition probabilities} between hidden states in \textit{S} can be described by a transition matrix $T \in \R{}^{N\times N}$, such that $T_{ij}$ describes the probability of transitioning from state $S_i$ to $S_j$. The \textit{emission probabilities} can be described as an emission matrix $O \in \R{}^{N\times M}$, where $O_{ij}$ describes the probability to see observation $E_j$ when in state $S_i$. Lastly the probability of \textit{starting} in a given state \textit{s} can be described as a randomly initialised vector $P \in \R{}^N$. As such the Hidden Markov Model can be more formally described as:
\begin{align}
    \hmm{} = \{S, E, T, O, P\}
\end{align}

The chosen implementation in the \texttt{hmmlearn} library uses a Gaussian distribution\footnote{\url{https://hmmlearn.readthedocs.io/en/latest/api.html\#hmmlearn.hmm.GaussianHMM} 28-05-2019}, where the randomly initialised probabilities are adjusted by the Baum-Welch algorithm \cite{baum-welch_paper} on the training data. When $\hmm{}$ has been ``trained" with Baum-Welch, the Viterbi algorithm \cite{viterbi_paper} is used to determine the most likely sequence of hidden states and the total probability of some sequence of observations $E' = \{e_1, e_2, .., e_k\}$ with length \textit{K}. \\

The approach used in \cite{dungs18} relinquishes, among other, textual features to rely solely on sequences of stance labels in their model $\hmm{}$ and sequence of stance labels \textit{and} time stamps in the ``Multi-spaced" HMM $\lambda'$. Inspired by $\lambda'$, we implement a variation of $\hmm{}$, denoted $\mshmm{}$. $\lambda'$ initialises a random real number for each stance and a weight which is learned given a distribution function over the time stamps\footnote{See \cite[4.3]{dungs18} for a description of this more complex HMM version}. $\mshmm{}$ works much like $\hmm{}$, however normalised time stamps are included as a feature. This was done as temporal properties were observed to boost performance in the \cite{dungs18} results. The more complex multi-spaced HMM ($\lambda'$) was however deemed out of scope for this project given time and resource constraints, although it would be interesting to apply.

For classification a HMM is built for each label with a varying state space size ranging from $ 1 \leq n \leq 15$. The prediction for some sequence \textit{Q} is determined by which model outputs the greater probability for \textit{Q}.

\subsubsection{Testing approach}
\label{methods_hmm_testing_approach}
The fact that the models rely solely on stance labels and time stamps opens up the opportunity for stance labelled data to be used across languages and platforms. This is especially interesting as the \dataset{} dataset only contains 16 rumours, with 1,496 posts across 220 conversations. As such it should be possible to use the PHEME dataset \cite{pheme-dataset} in conjunction with \dataset{}. The \pheme{} dataset is a popular and widely used Twitter dataset, such as in RumourEval 2017 \cite{derczynski17}. The dataset contains 4,842 tweets across 297 English and 33 German Twitter conversations, out of which 159 are true, 68 are false and 103 are labelled as unverified \cite{zubiaga16}. The data used for training is however a subset of \pheme{}, as described by  \cite{dungs18}. Here only 5 events yielded 5 or more rumours with 5 or more tweets in the conversation. The same choice was applied here to align with their approach. 

First, in order to investigate the Danish data isolated, 3-fold cross validation will be performed solely on the Danish data. Further, to see how well the data can be used in conjunction, the \pheme{} dataset will be utilised in two ways: (1) as training data for the models, with the Danish data as test set, and (2) mixed with the Danish data in 3-fold cross validation.

As a baseline throughout the experiments, a simple stratified baseline will be used, denoted as \textit{VB}. The baseline notes the average distribution of stance labels as a four-tuple for respectively true and false (and unverified where relevant) rumours. When predicting rumour veracity, \textit{VB} calculates the distribution of stance labels in a given sequence in the testing data and chooses the truth value with the most similar class label distribution.

%% file: experiments_stance.tex
\section{Stance classification experiments}
\label{experiments}
This section reports on various experiments carried out in order to reach the best performing models for the stance classification task. These experiments constitutes model selection techniques including feature selection, parameter search, and data sampling. However for feasibility reasons, for the case of the LSTM the parameter search has only been carried out.

\subsection{Feature selection}
\label{feature_selection}
The features generated to represent a data point in our dataset are compiled from various research, as introduced in section \ref{features}. However, those features worked great for \textit{their} data, which is not a given will be the same case for our data. As such this section reports on various experiments with the goal of selecting those features from which our models benefit the most, while still considering generalisation strength with regards to domain and platform. 

\subsubsection{Ablation study}
Aside from experimenting with all the features, an ablation study for the feature categories has been carried out. This is done by holding one feature category out at a time while the rest remains. This should unveil the effectiveness of each feature category. Further it allows possible discrepancies in data preference for the models to be revealed, as some models might have better results on different data. Optimally all combinations of features would be tested, but this becomes quite unfeasible with 8 feature \textit{categories}, covering several \textit{individual} features.

\begin{figure}[h]
    \centering
    \includegraphics[width=\textwidth]{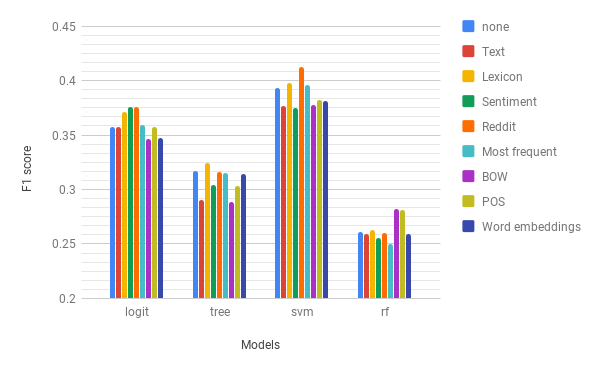}
    \caption{Feature importance per model measured with averaged $F_1$ macro through 5-fold cross validation}
    \label{fig:feature_importance}
\end{figure}

This experiment has been carried out with default versions of the \sklearn{} models introduced in section \ref{stance_ml_classifiers}, leaving the configuration to the default values defined by the library implementations. The results are shown in figure \ref{fig:feature_importance}, where tests are run with 5-fold cross validation with macro $F_1$ as scoring metric\footnote{As well as accuracy, but those results deviated too little in order to use it for this analysis}. \\

The blue pins are results where no features have been removed and are thus the pins to compare the others to. As such, a higher result than this means that we achieve better results by removing the given feature, and vice versa. We see that it is the general case that by removing lexicon features we achieve a higher macro $F_1$ score. Both for \textit{logit} and \textit{svm}, the Reddit features also do not seem to do any good, while they do not change much for \textit{tree} and \textit{rf} (the classic classifiers introduced in section \ref{stance_ml_classifiers}). It is difficult to say anything in general about the remaining features, as they don't seem to show an overall trend, such as the sentiment features for Logistic Regression, which has opposite effect on the other models. As such it seems that we might benefit across the line by excluding lexicon and Reddit features. Table \ref{tab:features_removed} shows that by running the same cross validation test with the default models, but with both lexicon and Reddit features removed, we see an expected improvement in macro $F_1$ score for almost all of the models.

\begin{table}[h]
    \centering
    \begin{tabular}{c|r|r}
        Model & All features & Lexicon+Reddit removed\\
        \hline
        \textit{logit} & 0.36 & 0.38\\
        \textit{tree} & 0.32 & 0.32\\
        \textit{svm} & 0.39 & \textbf{0.41}\\
        \textit{rf} & 0.26 & 0.25\\
        \hline
    \end{tabular}
    \caption{$F_1$ score of the default models with respectively no features removed and lexicon and Reddit features removed}
    \label{tab:features_removed}
\end{table}

\subsubsection{Removing low-variance features}
Furthermore, a second strategy for testing the feature importance has been to reduce the number of features through a simple but effective feature selection method, which further improves running time for the classifiers. The feature vectors are quite long, as can be seen in Table \ref{tab:feature_vector}, section \ref{feature_vector_overview}, with a total length of 14,143, with BOW being the main contributor with its 13,663 values. As the BOW feature consists of only 0's and 1's, one can imagine that there might be low variance for some of the variables. \texttt{VarianceThreshold}(\texttt{VT})\footnote{\url{https://scikit-learn.org/stable/modules/generated/sklearn.feature_selection.VarianceThreshold}} from \sklearn{} has been used to eliminate low-variance features, that is, features which under a predefined threshold only occur few times in the samples.

The benefits of removing features which rarely change can be compared to the concept of \q{information gain} used in Decision Trees. The information gain is a metric which describes how \q{pure} a split of labels would be given some variable. If all vectors across all labels for example have the same value in some dimension \textit{d}, the feature provides no partitioning information about the class labels. In that case the feature is more likely to be noise than helpful information.

Table \ref{tab:feature_selection} shows the number of features eliminated with variance thresholds of respectively 0\%, 0.1\%, and 1\%. With only a 0.1\% threshold, the number of features are reduced to 3,288, equivalent to a reduction of nearly 77\% of the features. Further, we see that it indeed is the BOW features having very low variance. What is also interesting is the fact that the lexicon features are removed altogether with a variance threshold of 1\%, as well as removal of one Reddit feature and one POS feature with a 0\% variance threshold. It is however not that surprising for the lexicon and Reddit features, as the previous feature selection experiment showed that these two categories have negative impact to some degree for each of the models. \\


\begin{table}[h]
    \centering
    \begin{tabular}{l|r|r|r|r}
        \textit{Category} & All & 0\% & 0.1\% & 1\% \\
        \hline
        Text & 13 & 13 & 10 & 4 \\
        Lexicon & 4 & 4 & 3 & 0 \\
        Sentiment & 1 & 1 & 1 & 1\\
        Reddit & 10 & 9 & 9 & 6\\
        Most frequent words & 132 & 132 & 132 & 129 \\
        BOW & 13,663 & 13,663 & 2,814 & 381\\
        POS & 17 & 16 & 16 & 16 \\
        Word embeddings & 303 & 303 & 303 & 303\\
        \hline
        \textbf{Total} & 14,143 & 14,141 & 3,288 & 840 \\
    \end{tabular}
    \caption{Low variance feature removal by feature category}
    \label{tab:feature_selection}
\end{table}

\begin{figure}[h]
    \centering
    \includegraphics[width=0.9\textwidth]{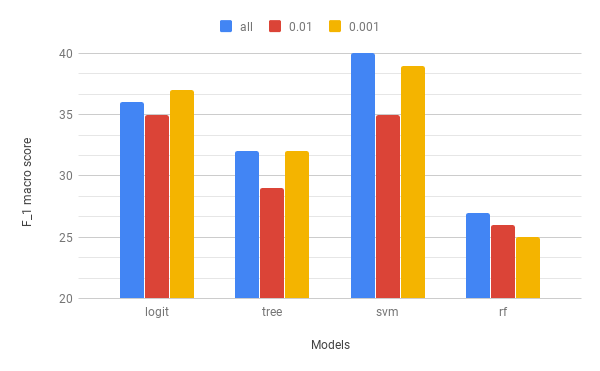}
    \caption{Macro $F_1$ score of default models with \texttt{VarianceThreshold} feature reduction}
    \label{fig:simple_models_vt_f1}
\end{figure}

In order to test the impact of removing low-variance features, this experiment was run with the default \sklearn{} models. Figure \ref{fig:simple_models_vt_f1} illustrates the macro $F_1$ score of the different models through 5-fold cross validation with respectively no feature reduction, 1\%, and 0.1\% thresholds. From this one can see that removing features with variance less than 1\% does have some influence, while 0.1\% is \textit{almost} identical to removing no features. 

When combining \texttt{VarianceThreshold} with removal of lexicon and Reddit features, the benefit is not uniform across the classifiers. \textit{logit} and \textit{rf} are the only models which get better by using both feature selection strategies, while \textit{tree} and \textit{svm} get worse performance, as can be seen in Table \ref{tab:features_vt_and_lr}.

\begin{table}[h]
    \centering
    \begin{tabular}{c|r|r}
        Model & Lexicon+Reddit & Lexicon+Reddit+\texttt{VT}\\
        \hline
        \textit{logit} & 0.38 & \textbf{0.40}\\
        \textit{tree} & 0.32 & 0.31\\
        \textit{svm} & \textbf{0.41} & 0.38\\
        \textit{rf} & 0.25 & 0.26\\
        \hline
    \end{tabular}
    \caption{$F_1$ scores for the default models with lexicon and Reddit features removed, with and without \texttt{VarianceThreshold}(\texttt{VT})}
    \label{tab:features_vt_and_lr}
\end{table}

\subsubsection{BOW vs Most frequent words}
\label{bow_vs_mfw}
Using the variance threshold approach (\texttt{VT}) with the BOW feature is essentially the same as removing least frequent words, which roughly translates to keeping the most frequent words. Thus, experiments were carried out in order to determine if either BOW or the \q{Most frequent words}(MFW) features should be left out entirely. Table \ref{tab:features_bow_vs_mfw} shows that leaving MFW out and using \texttt{VT} improves performance for \textit{logit} and just leaving MFW out leaves \textit{svm} unchanged. These experiments were only performed on the \textit{logit} and \textit{svm} models, as these have performed best so far in the experiments reported. \\

\begin{table}[h]
    \centering
    \begin{tabular}{l|r|r}
        Features removed & \textit{logit} & \textit{svm} \\
        \hline \hline
        None & 0.36 & 0.39 \\
        \hline
        BOW & 0.36 & 0.37 \\
        MFW & 0.38 & \textbf{0.39} \\
        MFW+(All-\texttt{VT}) & \textbf{0.39} & 0.38 \\
        MFW+(BOW-\texttt{VT}) & \textbf{0.39} & 0.38 \\
        \hline
    \end{tabular}
    \caption{$F_1$ scores of the default models with respectively BOW and MFW features left out, in combination with \texttt{VarianceThreshold}(\texttt{VT}) applied on respectively all features(All-\texttt{VT}) and BOW features(BOW-\texttt{VT})}
    \label{tab:features_bow_vs_mfw}
\end{table}


Thus, we can conclude that we can \q{safely} remove the MFW features without \textit{decreasing} performance. This is especially valueable, as it turns out that the MFW features actually are quite specific for our dataset. This is evident from the generated words, as listed in appendix \ref{appendix:most_frequent_words}. We see words such as \q{B12}, \q{CO2}, and \q{5G}, which is due to the concentration of events in the dataset, which for these three cases are respectively \q{Kost}(diet) for the former two, and \q{5G} for the latter (see Table \ref{tab:events_and_submissions_annotated} for the overview of events). Thus, using this list/dictionary of words might have unwanted consequences if applied to unseen data. 

\subsubsection{Comparing word embeddings}
As introduced in section \ref{features}, \texttt{word2vec} (\texttt{w2v}) and \texttt{fastText} are used as word embeddings, the latter both as pre-trained word vectors (\texttt{ft}) as well as trained on the DSL corpus and Reddit data (\texttt{ft'}). This section compares the performance when each of these are used. As setup, we use the two best performing models from the feature selection section above, being \textit{logit} and \textit{svm}, where lexicon and Reddit features are removed, and the former is combined with \texttt{VarianceThreshold}. Again, we use 5-fold cross validation and macro $F_1$ as scoring metric. Table \ref{tab:word_embeddings} indicates that we achieve no gain in performance by employing other algorithms than \texttt{word2vec} on the DSL(+Reddit) corpus.

\begin{table}[h]
    \centering
    \begin{tabular}{c|r|r|r}
        Model & \texttt{w2v} & \texttt{ft} & \texttt{ft'}\\
        \hline
        \textit{logit} & \textbf{0.40} & 0.36 & 0.38\\
        \textit{svm} & \textbf{0.41} & 0.41 & 0.39\\
        \hline
    \end{tabular}
    \caption{Comparison of the different word embeddings. word2vec (\texttt{w2v}) and fastText (\texttt{ft'}) are trained on the DSL+Reddit corpus, while fastText (\texttt{ft}) is with pre-trained word vectors.}
    \label{tab:word_embeddings}
\end{table}

\subsubsection{Best feature configurations in summary}
\label{feature_comparison_summary}
The experiments presented in this section have investigated the importance of the individual feature categories as a step in finding the best model for rumour stance classification. Although the feature selection methods are non-exhaustive they have revealed interesting properties about the features. Table \ref{tab:feature_selection_all} gives an overview of the results for the feature selection in this section and highlight the best ones. The underlined results are the very best results obtained, while the ones in bold are the best results, when lexicon, Reddit and ``Most frequent words" features are removed, which is desirable as described below.

First of all, throughout the experiments the Logistic Regression (\textit{logit}) and Support Vector Machine (\textit{svm}) models have been superior to the Decision Tree (\textit{tree}) and Random Forest (\textit{rf}) classifiers, showing results around the 0.4 macro $F_1$ score mark. This could be due to the skewedness of the data, as both \textit{logit} and \textit{svm} are known to be robust in this regard (see section \ref{stance_ml_classifiers}). While the Random Forest model is also known to be robust to skewedness and outliers, this is dependent on the quality of the Decision Tree classifiers which make up the ensemble of the \textit{forest}. As such the results of the Decision Tree model indicate that the success of a Random Forest with more trees might be limited.  

Second, we conclude that the lexicon and Reddit features do not make positive contributions to the performance. Furthermore it seems that \textit{logit} especially can benefit from removing low-variance features. Finally we exclude the ``Most frequent words" features, as (1) they show to be similar to the BOW features with mentioned low-variance feature reduction applied, and (2) they are too domain-specific, including some words only relevant to events included in the dataset, such as ``B12", ``CO2", and ``5G". 

\begin{table}[h]
    \centering
    \begin{tabular}{l|r|r}
        Features removed & \textit{logit} & \textit{svm} \\
        \hline \hline
        None & 0.36 & 0.39 \\
        \hline
        Lexicon+Reddit & 0.38 & \underline{0.41} \\
        Lexicon+Reddit+(All-\texttt{VT}) & \underline{0.40} & 0.38 \\
        \hline
        BOW & 0.36 & 0.37 \\
        MFW & 0.38 & 0.39 \\
        MFW+(All-\texttt{VT}) & 0.39 & 0.38 \\
        MFW+(BOW-\texttt{VT}) & 0.39 & 0.38 \\
        \hline
        Lexicon+Reddit+BOW & 0.35 & 0.36 \\
        Lexicon+Reddit+MFW & 0.38 & \textbf{0.40} \\
        Lexicon+Reddit+MFW+(All-\texttt{VT}) & \textbf{0.39} & 0.37 \\
        Lexicon+Reddit+MFW+(BOW-\texttt{VT}) & \textbf{0.39} & 0.38 \\
        \hline
    \end{tabular}
    \caption{Macro $F_1$ scores of the default models with respectively BOW and MFW features left out, in combination with Lexicon+Reddit features removed and \texttt{VarianceThreshold}(\texttt{VT}) applied on respectively all features(All-\texttt{VT}) and BOW features(BOW-\texttt{VT})}
    \label{tab:feature_selection_all}
\end{table}

\subsection{Parameter search}
This section reports on the findings for doing parameter search for respectively the LSTM model, Logistic Regression (\textit{logit}), and Support Vector Machine (\textit{svm}). Although parameter search has been carried out for Decision Tree (\textit{tree}) and Random Forest (\textit{rf}), the results are not included, as they still perform sub-optimally compared to \textit{logit} and \textit{svm} (see section \ref{feature_comparison_summary}).

\subsubsection{LSTM parameters search}
The hyper-parameter space for the LSTM model is presented in Table \ref{tab:lstm_hyper_params}. These are searched with a grid-search strategy, exhaustively running through all combinations of parameters in the parameter space. Another common approach for parameter search is the random-search, which can be preferable to grid-search in some cases. The random-search approach might uncover optimal parameters which are not present in the grid-search parameter space. However \cite{kochkina17} which inspired the use of the LSTM model already uncovered an effective set of hyper-parameters for this problem and model. Although it was another dataset, no motivation was found to deviate from the approach which yielded such strong stance classification results.

\begin{table}[h]
    \centering
    \begin{tabular}{|l|r|}
        \hline
        \textit{Parameter} & \textit{Value set} \\
        \hline
        LSTM Layers & \{1, 2\} \\
        \hline
        LSTM Units & \{100, 200, 300\} \\
        \hline
        ReLU Layers & \{1, 2\} \\
        \hline
        ReLU Units & \{100, 200, 300, 400, 500\}\\
        \hline
        Epochs & \{50\} \\
        \hline
        Dropout & \{0.00, 0.25, 0.50\} \\
        \hline
        L2 Regularisation strength & \{0, 1e-3\} \\
        \hline
    \end{tabular}
    \caption{Hyper-parameter space for LSTM classifier}
    \label{tab:lstm_hyper_params}
\end{table}

The LSTM grid search is performed through Google Colab (see section \ref{technologies}). This makes the brute grid search much more feasible to do, as \texttt{PyTorch} supports GPU utilisation\footnote{In particular \texttt{PyTorch} supports CUDA with GPU compute compatibility 3 or higher, at the time of writing}. The parameters for the five best scoring models are reported in Table \ref{tab:lstm_param}. Note that the tests performed for the LSTM are with all features enabled. 

\begin{table}[h]
    \centering
    \begin{tabular}{rrrrrr|rr}
        LSTM-L & LSTM-U & ReLU-L & ReLU-U & Dropout & L2 & $F_1$ & Acc. \\
        \hline
        1 & 100 & 1 & 400 & 0 & 0 & \textbf{0.39} & 0.73\\
        1 & 300 & 1 & 300 & 0.25 & 0 & 0.36 & 0.73\\
        1 & 300 & 1 & 500 & 0.25 & 0 & 0.35 & \textbf{0.76}\\
        1 & 200 & 1 & 200 & 0 & 1e-3 & 0.34 & 0.68\\
        1 & 200 & 1 & 500 & 0 & 0 & 0.34 & 0.68\\
         \hline
    \end{tabular}
    \caption{Parameter configurations for the five best performing LSTM models on macro-averaged $F_1$}
    \label{tab:lstm_param}
\end{table}

\begin{figure}[h]
    \centering
    \includegraphics[width=0.7\textwidth]{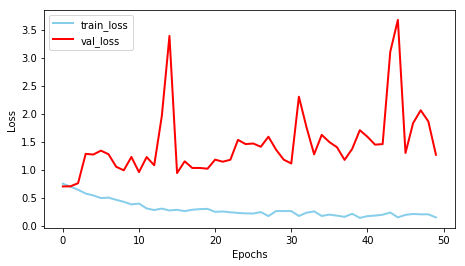}
    \caption{Loss graph for the best LSTM parameter search run}
    \label{fig:lstm_loss}
\end{figure}

With a macro $F_1$ score of 0.39 and accuracy of 0.73, the best parameter combination is with respectively one LSTM layer (LSTM-L) and one ReLU layer (ReLU-L), respectively 100 LSTM units (LSTM-U) and 400 ReLU units (ReLU-U), and no dropout or regularisation. We clearly see the tendency of single layers and a high number of ReLU units across all of the results. For each epoch, the model is also evaluated on a development set, in order to keep track of the training loss. Figure \ref{fig:lstm_loss} illustrates that the model quite early in the training epochs start to overfit. Applying 0.5 dropout with this parameter configuration, the model still overfits, even though the large ``spikes" are reduced.

\begin{figure}[h]
    \centering
    \includegraphics[width=0.8\textwidth]{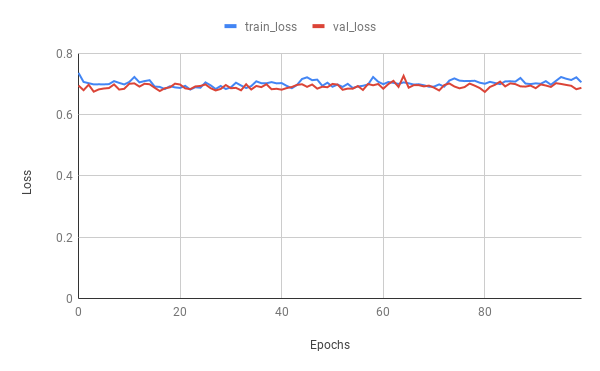}
    \caption{Loss graph for the LSTM without overfit}
    \label{fig:lstm_loss_smooth}
\end{figure}

Applying dropout and regularisation to the LSTM model \textit{did} result in less overfitting for some experiments, however with the cost of poorer results. Some general tendencies were seen in the loss graphs for the search space. An example is the loss graph illustrated in Figure \ref{fig:lstm_loss_smooth}, which has one LSTM layer, 100 LSTM units, two ReLU layers, and 100 ReLU units, but with 0.5 dropout and 0.001 L2 regularisation applied. In this case the model did not overfit to the training data, but was on par with the validation set. This loss graph represents a general tendency, where the LSTM achieves macro \fone{} scores on the test set in the 0.20-0.30 range, and neither validation loss or training loss declining. This could indicate that the skewed label distribution makes it difficult to minimise the loss on the training set while still maintaining good results on the validation and test sets.

\subsubsection{\textit{logit} and \textit{svm} parameter search}
The non-neural network models presented in \ref{stance_ml_classifiers} as the ``classic classifiers" have a number of different parameters, which are documented in the \sklearn{} API\footnote{\url{https://scikit-learn.org/stable/modules/classes.html} 26-04-2019}. These parameters have been searched through various strategies, including standard grid-search and a randomised search. 


The \textit{logit} and \textit{svm} models are quite similar as they both learn a linear function and uses the ``liblinear" optimisation algorithm. Thus, they are actually tuned on the same parameter space, as shown in Table \ref{tab:logit_svm_parameters}. Initially a randomised search was carried out, using the \texttt{RandomizedSearchCV}\footnote{\url{https://scikit-learn.org/stable/modules/generated/sklearn.model_selection.RandomizedSearchCV.html} 06-05-2019} implementation for \sklearn{}, which generates a predefined number of random parameter configurations, performs cross validation, and evaluates on a held out testing sample. With 3-fold CV and 10 random samples we would have 30 train-test iterations, allowing us to get a first impression of the behaviour. With random number generators we learned which values might be valuable to include in a grid-search for the non-nominal parameter settings. Additionally we learned that L2 regularisation was superior to L1 for both models. 3-fold CV was also performed with the grid-search with evaluation on a test sample, using \texttt{GridSearchCV}\footnote{\url{https://scikit-learn.org/stable/modules/generated/sklearn.model_selection.GridSearchCV.html} 16-05-2019}, resulting in 72 train-test iterations, leaving out parameter combinations with L1.

\begin{table}[h]
    \centering
    \begin{tabular}{|l|r|}
        \hline
        \textit{Parameter} & \textit{Value set} \\
        \hline
        Penalty & \{`L1', `L2'\} \\
        \hline
        C & \{1, 10, 50, 100, 500, 1000\} \\
        \hline
        Class weight & \{`balanced', \texttt{None}\} \\
        \hline
        Dual & \{\texttt{True}, \texttt{False}\} \\
        \hline
    \end{tabular}
    \caption{Parameter space for \textit{logit} and \textit{svm}}
    \label{tab:logit_svm_parameters}
\end{table}

The parameters are defined as follows\footnote{\url{https://scikit-learn.org/stable/modules/generated/sklearn.linear_model.LogisticRegression.html} 16-05-2019}: ``Penalty" specifies the norm used in penalisation and `C' specifies the inverse of regularisation strength, smaller values specifying stronger regularisation. The ``class weight" specifies the weights applied to the classes, where \texttt{None} means all classes have weight one, and ``balanced” uses the values of the true class labels to automatically adjust weights inversely proportional to class frequencies in the input data. Finally ``dual" specifies whether to solve the dual or primal optimisation problem.

Note that experiments were actually carried out with an SVM using a non-linear kernel (``RBF") as well\footnote{Using the \sklearn{} model from: \url{https://scikit-learn.org/stable/modules/generated/sklearn.svm.SVC.html} 16-05-2019}, however the results were even worse than \textit{tree} and \textit{rf}, which is why they were discarded. \\

The feature configuration was based on the ones marked as most suited in section \ref{feature_selection} (see section \ref{feature_comparison_summary} for a summary), leaving out lexicon, Reddit, and ``Most frequent words" features, as well as removing low variance features for the case of \textit{logit}. The optimal parameters found, based on the best evaluation score throughout the grid-search are presented in Table \ref{tab:optimal_parameters}. 

\begin{table}[h]
    \centering
    \begin{tabular}{l|l|r|r}
        Model & Optimal parameters & $F_1$ & Accuracy\\
        \hline
        \textit{logit} & C=1, class\_weight=``balanced", dual=\texttt{True} & \textbf{0.4473} & 0.7409\\
        \textit{svm} & C=10, class\_weight=\texttt{None}, dual=\texttt{True} & 0.4253 & \textbf{0.7508}\\
        \hline
    \end{tabular}
    \caption{Optimal parameters for \textit{logit} and \textit{svm} based on macro $F_1$ evaluation score}
    \label{tab:optimal_parameters}
\end{table}

Note that the data is in fact split two times, first in a train-test split, leaving out the test sample for evaluation and then the training sample is split into train-development sets throughout cross validation in the grid-search. Additionally, different from the feature selection experiments, 3-fold CV was used. Because of these factors the high results are not truly comparable to the ones obtained so far. 

\subsection{Stance classification results}
\label{experiments_stance_results}
Running the Logistic Regression (\textit{logit}) and Support Vector Machine (\textit{svm}) classifiers through 5-fold CV with the optimal parameters, should give the best representable results achieved so far. Table \ref{tab:stance_results} presents the final results for the tuned models, denoted as \textit{logit}' and \textit{svm}', as well as the default models using default parameters and all features, and the baseline models. Even though it performs poorly, the parameter-tuned LSTM is also included, performing the same CV, with all features (\textit{LSTM}) and with lexicon, Reddit, and MFW features removed (\textit{LSTM}').

\begin{table}[h]
    \centering
    \begin{tabular}{l|rr|rr}
        Model & Macro-$F_1$ & std. dev. & Accuracy & std. dev \\
        \hline
        \textit{MV} & 0.2195 & (+/- 0.00) & \underline{0.7825} & (+/- 0.00) \\
        \textit{SC} & 0.2544 & (+/- 0.04) & 0.6255 & (+/- 0.01) \\
        \hline
        \textit{logit} & 0.3778 & (+/- 0.06) & 0.7812 & (+/- 0.02)\\
        \textit{svm} & 0.3982 & (+/- 0.04) & 0.7496 & (+/- 0.02)\\
        \textit{LSTM} & 0.2802  & (+/- 0.04) & 0.7605 & (+/- 0.03) \\
        \hline
        \textit{logit}' & 0.4112 & (+/- 0.07) & 0.7549 & (+/- 0.04)\\
        \textit{svm}' & \textbf{0.4212} & (+/- 0.06) & 0.7572 & (+/- 0.02)\\
        \textit{LSTM}' & 0.3060 & (+/- 0.05) & 0.7163 & (+/- 0.16) \\
        \hline
    \end{tabular}
    \caption{5-fold cross validation results for \textit{logit}, \textit{svm}, LSTM, and baselines with macro $F_1$ and accuracy, including standard deviation(std. dev.).}
    \label{tab:stance_results}
\end{table}

We see that \textit{svm}' is the best performing model, achieving a macro $F_1$ score of 0.42, an improvement of 0.02 over the default model. It is however only marginally better than \textit{logit}', taking the deviation into account. Note that the accuracy is worse than the \textit{MV} baseline, and \textit{logit}' has even decreased its accuracy. The reason for this could be that the models have been tuned for macro $F_1$, as discussed in section \ref{scoring_metrics}. Tables \ref{tab:sdqc_f1} and \ref{tab:sdqc_acc} demonstrates how the models \textit{really} improve over the baselines by more fairly looking at respectively the $F_1$ and accuracy per class. As expected we see that \textit{MV} only predicts ``commenting" classes and that \textit{SC} follows the class label distribution of the dataset, while \textit{logit}' and \textit{svm}' are able to predict the under-represented classes. Because of the low-volume data in \dataset{} we did not expect the LSTM to perform very well, which is evident from the best macro $F_1$ score of 0.3060. For this reason we focus on \textit{logit} and \textit{svm} in the remainder of the experiments.\\

\begin{table}[h]
    \centering
    \begin{tabular}{c|rrrr}
        \diagbox{Model}{Class} & S & D & Q & C \\
        \hline
        \textit{MV} & 0.00 & 0.00 & 0.00 & 0.88 \\
        \textit{SC} & 0.11 & 0.10 & 0.04 & 0.80 \\
        \hline
        \textit{logit}' & 0.31 & 0.31 & 0.16 & 0.86 \\
        \textit{svm}' & 0.29 & 0.32 & 0.22 & 0.86 \\
        \hline
    \end{tabular}
    \caption{$F_1$ score per class}
    \label{tab:sdqc_f1}
\end{table}

\begin{table}[h]
    \centering
    \begin{tabular}{c|rrrr}
        \diagbox{Model}{Class} & S & D & Q & C \\
        \hline
        \textit{MV} & 0.00 & 0.00 & 0.00 & 1.00 \\
        \textit{SC} & 0.11 & 0.09 & 0.04 & 0.81 \\
        \hline
        \textit{logit}' & 0.27 & 0.28 & 0.12 & 0.89 \\
        \textit{svm}' & 0.24 & 0.28 & 0.19 & 0.90 \\
        \hline
    \end{tabular}
    \caption{Accuracy score per class}
    \label{tab:sdqc_acc}
\end{table}

\begin{figure}[h]
    \centering
    \begin{minipage}{.5\textwidth}
        \centering
        \includegraphics[width=\textwidth]{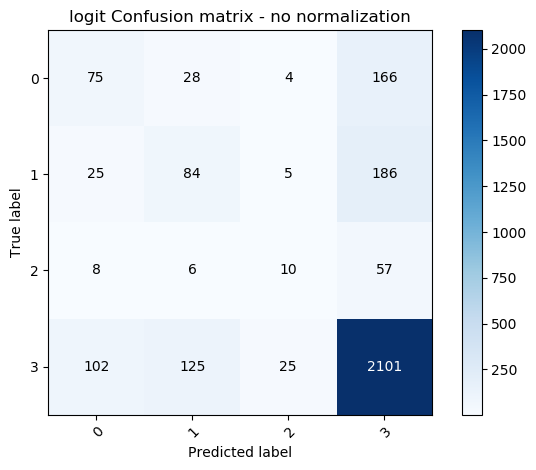}
    \end{minipage}%
    \begin{minipage}{0.5\textwidth}
        \centering
        \includegraphics[width=\textwidth]{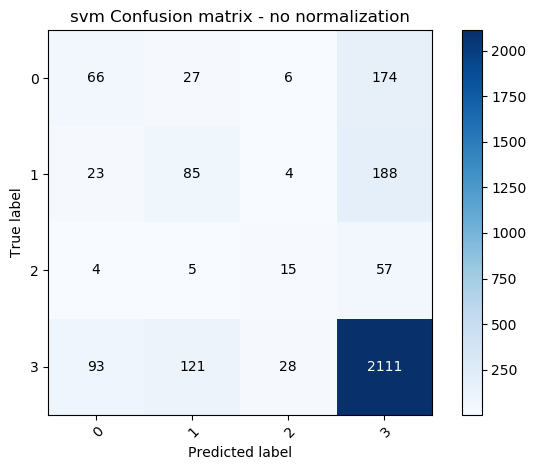}
    \end{minipage}
    \caption{Confusion matrices for the tuned classifiers, respectively \textit{logit}'(left) and \textit{svm}'(right). 0, 1, 2, 3 refer to the class labels S, D, Q, C in that order.}
    \label{fig:cms}
\end{figure}

Figure \ref{fig:cms} visualises the confusion matrices for both \textit{logit}' and \textit{svm}', where the numbers 0, 1, 2, 3 refer to the class labels S, D, Q, C in that order.

\textit{logit}' seems to be better at classifying the ``supporting" class, and \textit{svm}' seems to be better at classifying the ``querying" class, while they are equally good at classifying ``denying" and ``commenting" classes.

Additionally, investigating their ability to learn from the dataset provided can be interpreted through a learning curve. Figures \ref{fig:logit_learningcurves} and \ref{fig:svm_learningcurves} in appendix \ref{app:learning_curves} demonstrates the models' training and CV test scores through different sample sizes of the dataset. It becomes obvious that much more data is needed in order for the models to learn optimally. 

\subsection{Improving results with data sub- and super-sampling}
\label{subsec:data_sampling}
The skewed data with regards to the class labels (see Table \ref{tab:dataset_sdqc}) motivated data sampling as an experiment to smooth the class distribution. This was achieved by respectively reducing and amplifying relevant data points, with the goal and expectation of observing better results. Given the small size of \dataset{} the skewed distribution means that there is a very small amount of data points for the minority classes. As such the classification models have less data to train on and learn these class labels from, which makes it difficult to classify them. This is equivalent to saying that the large amount of ``commenting" posts makes it more likely for the classifiers to classify SDQ class labels as ``commenting".

\paragraph{Sub-sampling} of the data is done by removing all branches, which have pure \q{commenting} labels, effectively reducing the total size of the dataset from 3,007 to 2,313 data points, but improving the distribution of ``supporting", ``denying", and ``querying" classes.

\paragraph{Super-sampling} of the data is done by first splitting the data in stratified train and test set. Considering only the train set, for each post which is labelled as either S, D, or Q, it is duplicated, which allows us to alter the text (and thereby the features) to create a synthetic replica of the post representing one of the non-neutral class labels. The initial data split is performed as we do not want a classifier which is really good at predicting something almost identical to what it has already seen. This would be the case by having the original in one sample and the synthetic partner in the other (and vice versa).

The text is altered by trying to replace a fraction of the words by synonyms stored in a dictionary\footnote{\url{https://korpus.dsl.dk/e-resources/Synonyms\%20from\%20DDO.html} 08-05-19}, and then only allowing a post to pass as a super-sample candidate, if this succeeds. On one hand we do not want an example where the sentences are complete nonsense; conversely, we do not want to just replace a couple of words, leaving the synthetic post to be close to identical to the original one. Thus, iterative experiments with different thresholds of respectively 25\%, 37.5\%, and 50\% were carried out, resulting in the choice of 37.5\% providing a good ``original-to-synthetic" balance. Experiments have also been carried out with respectively \texttt{word2vec} and \texttt{fastText} word embeddings, replacing words with their most similar word in the vocabulary. The results were, however, quite strange, in particular when using pre-trained \texttt{fastText} word vectors. Examples for each of the three cases with replacement are included in appendix \ref{appendix:super_sampling}. \\

Table \ref{tab:sampling_cnt} provides an overview of the SDQC stats with sub- and super-sampling, as well as a combination of the two, where first the former is applied, and then the latter\footnote{The numbers from \q{Super} to \q{Sub+Super} are just a bit off, which is due to a stratified train-test split after removing data points with sub-sampling}. Additionally Table \ref{tab:sampling_pct} illustrates the shift in the class labels' relative contribution with the different sampling techniques.

\begin{table}[h]
    \centering
    \begin{tabular}{|l|r|r|r|r|r|}
    \hline
    \diagbox{\textit{Sampling}}{\textit{Label}} & S & D & Q & C & \textit{Total} \\
    \hline
    None & 273 & 300 & 81 & 2,353 & 3,007 \\ 
    \hline 
    Sub & 273 & 300 & 81 & \textbf{1,659} & 2,313 \\
    \hline
    Super & \textbf{412} & \textbf{462} & \textbf{124} & 2,353 & 3,351 \\
    \hline
    Sub+Super & \textbf{416} & \textbf{458} & \textbf{125} & \textbf{1,659} & 2,658 \\
    \hline
    \end{tabular}
    \caption{Stance label distribution count with sub-sampling, super-sampling, and their combination}
    \label{tab:sampling_cnt}
\end{table}

\begin{table}[h]
    \centering
    \begin{tabular}{|l|r|r|r|r|}
    \hline
    \diagbox{\textit{Sampling}}{\textit{Label}} & S & D & Q & C \\
    \hline
    None & 0.091 & 0.100 & 0.027 & 0.782 \\ 
    \hline 
    Sub & 0.118 & 0.130 & 0.035 & 0.717 \\
    \hline
    Super & 0.123 & 0.138 & 0.037 & 0.702 \\
    \hline
    Sub+Super & 0.157 & 0.172 & 0.047 & 0.624 \\
    \hline
    \end{tabular}
    \caption{Relative stance label distribution with sub-sampling, super-sampling, and their combination}
    \label{tab:sampling_pct}
\end{table}

For sub-sampling we see that comments are reduced from 2,353 to 1,659 improving SDQ contributions from 0.091, 0.1, 0.27 to 0.118, 0.13, and 0.035, respectively. This distribution is almost identical to super-sampling, but in this case we increase the (SDQ) data points with a total of 344, instead of reducing the ``commenting" class. Finally, the best distribution with regards to the SDQC class labels are achieved by first sub-sampling and then super-sampling, with distributions of 0.157(S), 0.172(D), 0.047(Q), and 0.624(C) and a total number of 2,658 data points. This distribution looks almost similar to the SDQC distribution in the \pheme{} dataset \cite{pheme-dataset}, but shrinks \dataset{} by around 12\%.\\

Running experiments with the different sampling techniques described in this section, the parameter-optimised \textit{svm} with lexicon and Reddit features removed (see section \ref{experiments_stance_results}), is evaluated through stratified CV on the different configurations, which are reported in Table \ref{tab:svm_sample}. As we do not want the original feature vector and the synthetic partner to be in each of their train/test set, we make sure they do not split up. 

\begin{table}[h]
    \centering
    \begin{tabular}{l|rr|rr}
        \textit{Sample} & Macro-$F_1$. & std. dev. & Accuracy & std. dev \\
        \hline
        None & 0.4212 & (+/- 0.06) & \textbf{0.7572} & (+/- 0.02) \\
        \hline
        Sub & 0.4050 & (+/- 0.06) & 0.6922 & (+/- 0.03) \\
        \hline
        Super & 0.4418 & (+/- 0.05) & 0.7106 & (+/- 0.02) \\
        \hline
        Sub+Super & \textbf{0.4807} & (+/- 0.09) & 0.6658 & (+/- 0.03) \\
        \hline
    \end{tabular}
    \caption{\textit{svm}' sample results}
    \label{tab:svm_sample}
\end{table}

First off, even though the sub- and super-sampling techniques yielded similar SDQC distributions, we see a clear advantage for the SVM with more data points, scoring 0.4050 with ``Sub" and 0.4418 with \q{Super}, the latter improving over the original result with $\sim0.02$. More interesting is the result for the combined sub- and super-sampling dataset, where the SVM scores a 0.4807 macro $F_1$, really improving with the much more balanced class distribution. With regards to the accuracy, we see a correlation with the number of data points and the number of ``commenting" classes: the less data points of the majority class, the lower the total number of correct predictions. This indicates that the model gets worse at learning that class, which makes sense since we drastically reduce the number of data points with sub-sampling. For the case of super-sampling, we would expect the accuracy to be more or less the same as with the original data set, which is not really the case with its 0.7106 compared to 0.7572. However this is more promising than for the super-sample cases. 

Completely separating the original posts and synthetic ones from the test folds with super-sampling yields different results, as illustrated in Table \ref{tab:svm_sample_train}. 

\begin{table}[h]
    \centering
    \begin{tabular}{l|rr|rr}
        \textit{Sample} & Macro-$F_1$. & std. dev. & Accuracy & std. dev \\
        \hline
        Super & 0.3910 & (+/- 0.09) & \textbf{0.8000} & (+/- 0.03) \\
        \hline
        Sub+Super &  \textbf{0.4412} & (+/- 0.10) & 0.7721 & (+/- 0.02)\\
        \hline
    \end{tabular}
    \caption{\textit{svm}' sample results with super-sample only in train}
    \label{tab:svm_sample_train}
\end{table}

It is interesting to see that we do get better performance with combined sub- and super-sampling, both with macro $F_1$(0.4412) and accuracy(0.7721), when retaining the original posts and synthetic ones only in the training set. \\

The results presented in this section show how \textit{smoothing} the class balance improves macro \fone{} for the stance classifier. Finally the results show be close to \sota{} results for the Branch-LSTM model \cite{kochkina17}, although the results are not directly comparable, as the results are obtained on separate datasets. This concludes the stance classification experiments, and we thus present the rumour veracity experiments next, in section \ref{sec:rumour_experiments}.

%% file: experiments_veracity.tex
\section{Rumour veracity prediction experiments}
\label{sec:rumour_experiments}
This section presents a number of experimental approaches, which have been applied in order to find the optimal solution to the task of rumour veracity prediction with the HMM approach introduced in section \ref{rumour_classification}.\\ 

Throughout the experiments exhaustive search is performed for the HMM models $\hmm{}$ and $\mshmm{}$ to identify the optimal state space size \texttt{N}, from 1 to 15, as introduced in section \ref{methods_hmm}. Further, as described in section \ref{methods_hmm_testing_approach}, the language-agnostic and platform-agnostic HMM approach allows us to use other stance-labelled datasets.

\subsection{Using data across languages and platforms}
We propose to utilise the PHEME dataset from \cite{zubiaga16} based on the idea of the HMM approach relying only on stance and posting times. However, as there might be some discrepancies, this is investigated in this section.

\begin{figure}[h]
    \centering
    \includegraphics[width=0.5\textwidth]{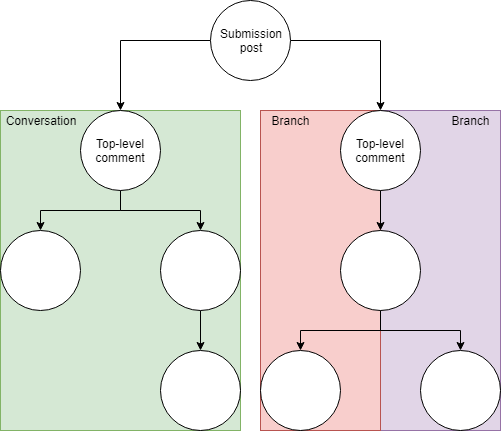}
    \caption{The structure of a Reddit submissions}
    \label{fig:reddit_submission_structure}
\end{figure}

Experiments are needed to determine the optimal partitioning and structure of the data. By using multi-platform datasets, discrepancies in the data structure arise, which should be kept in mind. While a submission text on the Reddit platform is the actual source of a rumour and a post, they do not always contain stance towards the rumour. Furthermore several of the submissions in \dataset{} are much larger conversation trees than the Twitter conversations if all the comments are grouped. The structure of a Reddit submission is illustrated in Figure \ref{fig:reddit_submission_structure}, as introduced in section \ref{problem:platform_structures}.

Apart from the structure-specific differences between Reddit and Twitter, the difference in language might also cause issues in the compatibility of the two datasets. First of all, Reddit is an anonymous platform, while Twitter is not, which might cause differences in posting tendencies. The language of the post texts in the two dataset also differ. The \pheme{} dataset \cite{pheme-dataset} contains English (and some German) text and \dataset{} contains only Danish text. These factors along with the structural differences in conversation lengths might introduce discrepancies, which hopefully will be discovered in the experiments presented in section \ref{experiments_veracity}. \\

The platform differences between Reddit and Twitter discussed above might influence the results obtained when training on the \pheme{} dataset \cite{pheme-dataset}. As such the experiments will be performed on three different structures of the \dataset{} data:

\paragraph{\textit{SAS}: Submission as a source}
is treated as a singular rumour, in which all replies and nested replies to the submission have been flattened to a single list and sorted from earliest to latest post. This generates few rumour entries with quite long sequences of stance. Some challenges with this structure of the data is the low amount of entries. There are only 16 rumour submissions, which makes for low amounts of training and test entries. Furthermore it might be difficult to apply the \pheme{} rumour data to this structure. The \pheme{} conversation data generally consist of lower amount of comments than the submissions in \dataset{}.

\paragraph{\textit{TCAS}: Top level comment as source} 
regards each conversation tree within a submission as a rumour. A conversation tree is the tree of posts which spawn from a top-level post. This generates more data entries and is more alike the \pheme{} Twitter conversation structure where each conversation is spawned from a source-tweet.

\paragraph{\textit{BAS}: Branch top level comment as source}
treats each individual branch in the rumour submissions as a rumour itself. This generates a lot of data entries, with a lower average length than the two structures above. Further it also implicates duplicates of the stance labels, since multiple replies to a single post will create different branches with shared parent posts. This approach might however prove useful for early detection, given the shorter average length of the branches. \\

\begin{figure}[h]
    \centering
    \includegraphics[width=\textwidth]{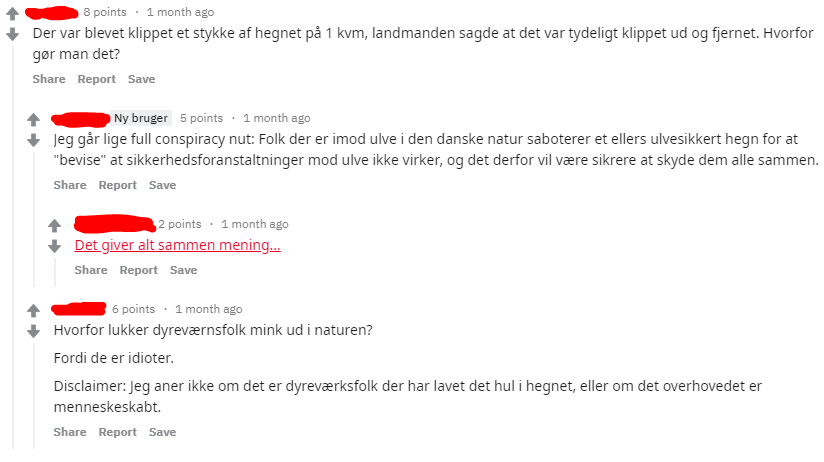}
    \caption{Reddit conversation tree example}
    \label{fig:reddit_example}
\end{figure}

Figure \ref{fig:reddit_example} displays an example of a short Reddit conversation. The conversation consists of 4 comments, of which one is a top-level comment. The conversation contains 2 branches respectively of length 3 and 2. As such this would yield 2 branches for the BAS structure, 1 for the TCAS structure and only be a part of the SAS structure, which consists of an entire submission.

As first discussed in section \ref{problem_stance_veracity} the sole reliance on stance labels and time stamps in the HMM approach from \cite{dungs18} opens an opportunity to possibly utilise data from other languages. This will be explored by training the HMM on the \pheme{} rumour stance labels and testing the results on the different structure iterations of \dataset{} introduced above. Further experiments will be carried out in which the data from \pheme{} and \dataset{} are used in conjunction with each other, i.e. mixed across training and testing set. 

\subsection{Veracity classification}
\label{experiments_veracity}
This subsection will present the experiments performed in regard to veracity classification. The results obtained from the experiments will be provided as the experiments are presented. \dataset{} contains only 3 true and 3 false rumour submissions, where the remaining 10 are unverified. This could very well reflect reality, as it can be difficult to obtain the actual truth value of a rumour. To investigate how to handle the unverified rumours, they are approached in 3 different ways. One is to see them as false rumours, given they have not been confirmed to be true yet. Another is to see them as true, since they have not been proven false yet. The results obtained from these two interpretations of the unverified rumours might reveal whether the stance tendencies in unverified rumours are more alike false or true rumours. The last approach taken is to move away from binary classification and do three-way classification instead, trying to predict respectively false, true and unverified rumours. \\

Each approach will involve three experiments, described next. 3 fold cross validation on \dataset{} will investigate how well the unverified approach can be expected to perform for \dataset{} data alone. Another experiment will investigate how well the data from PHEME can be used to classify data on \dataset{}. The experiment will train on PHEME data and test on all of \dataset{}. The third and last type of experiment will be 3 fold cross validation on the conjunction of \dataset{} and PHEME data. This experiment will shed some light on how well the data can be used across language, not only from the PHEME dataset to \dataset{}, but the other way around as well.

The results for `unverified as false', `unverified as true' and `three-way classification' are presented and analysed upon throughout this section. However, for readability, only the results for the best performing approach of `unverified as false' are included, while the results for the two other approaches are included in appendix \ref{app_veracity}.

\subsubsection{Treating unverified rumours as false}
\label{subsub:unv_false}
Table \ref{tab:danish_veracity} contains results from solely training and testing on \dataset{}. The accuracy and \fone{} columns shows the average results, including deviation, across folds. Notably the results for the SAS structure is the same across the two models. This might be a result of the small sample size this structure presents in regards to rumour count. Further $\hmm{}$ has higher \fone{}, but lower accuracy than $\mshmm{}$ on the TCAS structure. On the BAS structure $\mshmm{}$ shows superior results with a higher accuracy and \fone{}. \\

\begin{table}[h]
    \centering
    \begin{tabular}{l|c|c|c}
        Structure & Model & Acc. & $F_1$ \\ \hline
        \multirow{3}{*}{SAS}&$\hmm{}$ & 0.81 (+/- 0.03) & 0.45 (+/- 0.01) \\
        &$\mshmm{}$ & 0.81 (+/- 0.03) & 0.45 (+/- 0.01) \\ 
        & VB & 0.39 (+/- 0.58) & 0.36 (+/- 0.57) \\ \hline
        \multirow{3}{*}{TCAS}&$\hmm{}$ & 0.73 (+/- 0.02) & 0.63 (+/- 0.06) \\
        &$\mshmm{}$ & 0.79 (+/- 0.04) & 0.61 (+/- 0.07) \\ 
        & VB & 0.35 (+/- 0.13) & 0.35 (+/- 0.13) \\ \hline
        \multirow{3}{*}{BAS}&$\hmm{}$ & 0.78 (+/- 0.03) & 0.66 (+/- 0.02) \\
        &$\mshmm{}$ & \textbf{0.83} (+/- 0.02) & \textbf{0.68} (+/- 0.04) \\
        & VB & 0.43 (+/- 0.07) & 0.42 (+/- 0.07) \\ \hline
    \end{tabular}
    \caption{Danish veracity results on 3-fold cross validation}
    \label{tab:danish_veracity}
\end{table}

The confusion matrix seen in Table  \ref{tab:veracity_danish_cm} shows the distribution of sampling and gold labels for the best performing model and structure in Table \ref{tab:danish_veracity}. The majority label is the false label, which the model classifies correctly at a high rate. Roughly half of the true rumours are classified correctly in this case.\\

\begin{table}[h]
    \centering
    \begin{tabular}{|l|r|r|}
    \hline
    \diagbox{\textit{Predicted}}{\textit{Actual}} & False & True \\
    \hline
    False & 450 & 47 \\ 
    \hline 
    True & 56 & 43 \\
    \hline
    \end{tabular}
    \caption{Truth value distribution with BAS structure 3-fold cross validation for $\mshmm{}$}
    \label{tab:veracity_danish_cm}
\end{table}

Next, the results for training on the PHEME dataset and testing on \dataset{} can be seen in Table \ref{tab:pheme_danish_veracity}. Interestingly $\hmm{}$ generally shows better results than $\mshmm{}$ in this testing setup. This could indicate that the included time stamps used in $\mshmm{}$ does not generalise well from the PHEME data to \dataset{}. This could be caused by multiple factors such as the discrepancies between the Twitter and Reddit platform or the different languages for each dataset. The reusability of the PHEME data does however show promising result when relying solely on stance labels. The best results are seen on the SAS structure, with an accuracy of 0.88 and an \fone{} score of 0.71. As such it seems the PHEME dataset can be used across platforms and languages quite well, given stance labels.

\begin{table}[h]
    \centering
    \begin{tabular}{l|c|c|c}
        Structure & Model & Acc. & $F_1$ \\ \hline
        \multirow{3}{*}{SAS} & $\hmm{}$ & \textbf{0.88} & \textbf{0.71} \\
        & $\mshmm{}$ & 0.75 & 0.67 \\ 
        & VB & 0.81 & 0.45 \\ \hline
        \multirow{3}{*}{TCAS} & $\hmm{}$ & 0.77 & \textbf{0.66} \\
        & $\mshmm{}$ & \textbf{0.81} & 0.59 \\ 
        & VB & 0.80 & 0.62 \\ \hline
        \multirow{3}{*}{BAS} & $\hmm{}$ & \textbf{0.82} & \textbf{0.67} \\ 
        & $\mshmm{}$ & 0.67 & 0.57 \\ 
        & VB & 0.77 & 0.53 \\ \hline
    \end{tabular}
    \caption{Training on the PHEME dataset and testing on \dataset{}}
    \label{tab:pheme_danish_veracity}
\end{table}

To investigate the results found in Table \ref{tab:pheme_danish_veracity} for the model $\hmm{}$, see confusion matrix in Table \ref{tab:pheme_danish_cm}. Even though the three structures have different sizes and properties, the results share some traits. For these tests the model is much better at identifying false rumours, than true rumours. The worst performing model test on the TCAS structure correctly identifies over 85\% of false rumours. However more than half of the true rumours are incorrectly identified as false rumours for both TCAS and BAS tests. The tendency for the model to guess false rumours very precisely is also seen in Table \ref{tab:veracity_danish_cm}. \\

\begin{table}[h]
    \centering
    \begin{tabular}{|l|l|r|r|}
    \hline
    Structure & \diagbox{\textit{Predicted}}{\textit{Actual}} & False & True \\
    \hline
    \hline
    \multirow{2}{*}{SAS} & False & 13 & 0 \\ 
    \cline{2-4} 
    & True & 2 & 1 \\
    \hline
    \hline
    \multirow{2}{*}{TCAS} & False & 149 & 26 \\ 
    \cline{2-4}
    & True & 24 & 21 \\
    \hline
    \hline
    \multirow{2}{*}{BAS} & False & 447 & 50 \\ 
    \cline{2-4}
    & True & 57 & 42 \\
    \hline
    \end{tabular}
    \caption{Truth value distribution with PHEME data training and SAS structure testing}
    \label{tab:pheme_danish_cm}
\end{table}

While it can be difficult to compare these results to \cite{dungs18} because of the difference in datasets, there is some things to note. The \fone{} score is generally lower for the results here than achieved in \cite{dungs18}. However the accuracy scores are higher, although this metric can be misleading as earlier stated (see section \ref{scoring_metrics}), given the skewed label distribution of \dataset{}. The lower \fone{} scores here could also be the reflections of language and platform discrepancies between the \pheme{} dataset and \dataset{}. The different platforms or languages might entice different conversational dynamics, resulting in different sequencing of stance or different time stamp tendencies. \\

Finally, Table \ref{tab:mix_veracity} shows the results from doing 3-fold cross validation on a mix of the PHEME dataset and the different structures of \dataset{}. The last row section where structure is \q{None} refers to the results from doing cross-validation solely on the PHEME data. The best result is achieved by $\mshmm{}$ on the BAS structure, with an accuracy of 0.67 and an \fone{} of 0.62. The results for all the models on the TCAS and BAS structures improve over None. This indicates that the stance approach is transferable across language and platform using the \cite{dungs18} approach.

\begin{table}[h]
    \centering
    \begin{tabular}{l|c|c|c}
        Structure & Model& Acc. & $F_1$ \\ \hline
        \multirow{3}{*}{SAS}&$\hmm{}$ & 0.53 (+/- 0.09) & 0.53 (+/- 0.10) \\
        &$\mshmm{}$ & 0.55 (+/- 0.09) & 0.55 (+/- 0.10) \\ 
        & VB & 0.37 (+/- 0.03) & 0.31 (+/- 0.07) \\ \hline
        \multirow{3}{*}{TCAS}&$\hmm{}$ & 0.60 (+/- 0.07) & 0.58 (+/- 0.08) \\
        &$\mshmm{}$ & 0.64 (+/- 0.05) & 0.61 (+/- 0.05) \\ 
        & VB & 0.53 (+/- 0.04) & 0.38 (+/- 0.03) \\ \hline
        \multirow{3}{*}{BAS}&$\hmm{}$ & 0.60 (+/- 0.05) & 0.58 (+/- 0.05) \\
        &$\mshmm{}$ & \textbf{0.67} (+/- 0.03) & \textbf{0.62} (+/- 0.04) \\ 
        & VB & 0.49 (+/- 0.10) & 0.40 (+/- 0.01) \\ \hline \hline
        \multirow{3}{*}{None}&$\hmm{}$ & 0.55 (+/- 0.05) & 0.54 (+/- 0.07) \\
        &$\mshmm{}$ & 0.57 (+/- 0.08) & 0.55 (+/- 0.10) \\ 
        & VB & 0.43 (+/- 0.03) & 0.33 (+/- 0.08) \\ \hline
    \end{tabular}
    \caption{Training and testing on mix of PHEME data and different \dataset{} structures for unverified false}
    \label{tab:mix_veracity}
\end{table}

\subsubsection{Treating unverified rumours as true}
\label{subsub:unv_true}
In the following experiments the unverified rumours have been interpreted as true rumours. Comparisons between these results and the `Unverified as false' experiments in section \ref{subsub:unv_false} above, might reveal interesting properties about the data. The results for interpreting the unverified rumours as true, which can be seen in appendix \ref{app_veracity:unv_true}, were not as promising as interpreting them as false. The 3-fold cross validation experiment generally provided lower scores with the highest accuracy at 0.74 achieved with the $\mshmm{}$ and $\hmm{}$ models on the SAS structure. The highest \fone{} score is achieved by $\mshmm{}$ on the BAS structure, reaching 0.62. \\

The results for training on \pheme{} an testing on \dataset{} were not as good either. The highest accuracy achieved is 0.81 reached by the $\mshmm{}$ model on the SAS structure. The highest \fone{} score is 0.59, achieved on the SAS structure as well by the $\hmm{}$ model. \\

The results of doing 3-fold cross validation on the union of the PHEME dataset and \dataset{} are interesting. The accuracy is not improved by the addition of any \dataset{} data, with the highest reached being 0.84 both on PHEME data alone and with the addition of SAS. The best \fone{} is 0.62, achieved with the combination of PHEME data and BAS. Interestingly it was achieved by the baseline \textit{VB} with $\hmm{}$ and $\mshmm{}$ falling behind. This indicates that the sequence of stance labels might be different from the PHEME data and \dataset{} with the BAS structure, however the overall distribution of the stance labels are more alike than the sequence.

\subsubsection{Three-way rumour veracity classification}
Rather than interpreting the unverified rumours as either true or false, the experiments in this section investigate the results of doing three-way classification of true, false \textit{and} unverified rumours. The results of the experiments are presented in tables in appendix \ref{app_veracity:3way}. The results are generally worse than the results obtained from the binary classification experiments presented above in sections \ref{subsub:unv_false} and \ref{subsub:unv_true}. This was expected as this approach in contrast performs three-way classification. While the results are not comparable to the previous two experiment approaches, they are interesting as they give insight into classifying unverified rumours. \\

First, for three-way classification through 3-fold cross validation on \dataset{}, the highest accuracy is 0.61 scored on the SAS structure by the $\mshmm{}$ model however the deviation for the results on the SAS structure is very high. The second highest accuracy of 0.57 also contains the highest macro \fone{} result of 0.53 with a much lower deviation, for the $\mshmm{}$ model on the BAS structure.

Second, training on \pheme{} data and testing on \dataset{} for three-way classification also gave poor results. The highest accuracy is 0.62, achieved by the baseline \textit{VB} on the SAS structure. Further the best macro \fone{} scores are achieved by the \textit{VB} on the TCAS structure and by $\mshmm{}$ on the SAS structure. This could indicate that the unverified rumour structures are not very compatible across the \pheme{} dataset and \dataset{}.

Finally, the results of doing 3-fold cross validation on a mix of PHEME data and \dataset{} for three-way classification shows the difficulty of the task. The best results are achieved on the TCAS structure with an accuracy of 0.53 for the $\hmm{}$ model and a macro \fone{} score of 0.42 for the $\mshmm{}$ model. \\

The results for three-way classification were not very promising. The task is harder than binary classification and it seems the HMM method does not translate well to this task, given only stance and/or time stamps as features. The results achieved by interpreting unverified rumours as particularly false proved much more promising.

\subsection{Connecting stance classification and veracity prediction}
While the results so far show promising results, these all rely on gold labels made by humans. In order to show how well the results will translate to unseen data, the stance classifier and veracity classifier should be connected. This approach was introduced in section \ref{background:system_architecture} as being a possible way of doing rumour veracity prediction. Thus, instead of gold labels, the labels should be generated by the stance classifier for all rumour data and then used for the veracity classification component. Ultimately this would prove the system to have practical applications.

The stance labels will be generated by the SVM model (\textit{logit}) which showed promising results with a macro \fone{} score of 0.4212 and an accuracy of 0.7572 on cross validation tests. The automatic stance labels will be obtained by training on all of \dataset{} except one rumour submission, for which stance will be classified. This will be performed for each rumour submission. The SDQC distribution and performance across the rumours are presented in Table \ref{tab:auto_stance}. 

\begin{table}[h]
    \small
    \centering
    \begin{tabular}{c|l|rrrr|rr}
        \textit{Event} & \textit{Title (abbreviated)} & S & D & Q & C & $F_1$ & Acc. \\
        \hline
        \multirow{3}{*}{5G} & 5G-teknologien \dots & 7$|$7 & 6$|$5 & 2$|$0 & 27$|$30 & 0.37 & 0.62 \\
        & Det er ikke alle \dots & 7$|$2 & 6$|$4 & 3$|$0 & 57$|$67 & 0.21 & 0.73 \\
        & Uffe Elbæk er \dots & 11$|$8 & 30$|$7 & 0$|$3 & 82$|$105 & 0.32 & 0.67 \\
        \hline
        \multirow{2}{*}{\shortstack{Donald\\ Trump}} & Hvorfor må DR \dots & 10$|$4 & 6$|$3 & 0$|$1 & 28$|$36 & 0.30 & 0.64 \\
        & 16-årig blev \dots & 15$|$1 & 5$|$13 & 5$|$1 & 71$|$81 & 0.24 & 0.69 \\
        \hline
        \multirow{2}{*}{ISIS} & 23-årig dansk \dots & 2$|$3 & 31$|$19 & 8$|$2 & 104$|$121 & 0.39 & 0.76 \\
        & Danish student \dots & 1$|$0 & 9$|$3 & 0$|$3 & 14$|$18 & 0.33 & 0.67 \\
        \hline
        \multirow{2}{*}{Kost} & Bjørn Lomborg \dots & 5$|$5 & 15$|$6 & 2$|$0 & 75$|$86 & 0.28 & 0.73 \\
        & Professor: \dots & 16$|$21 & 25$|$17 & 0$|$0 & 186$|$189 & 0.42 & 0.74 \\
        \hline
        \multirow{1}{*}{MeToo} & Bjørks FB post \dots & 1$|$2 & 8$|$8 & 3$|$2 & 48$|$48 & 0.51 & 0.78 \\
        \hline
        \multirow{3}{*}{\shortstack{Peter\\ Madsen}} & Savnet ubåd \dots & 0$|$0 & 11$|$1 & 3$|$0 & 17$|$30 & 0.23 & 0.52 \\
        & Undersøgelser \dots & 4$|$5 & 0$|$10 & 6$|$4 & 71$|$62 & 0.38 & 0.79 \\
        & Peter Madsen: \dots & 11$|$13 & 34$|$20 & 10$|$6 & 214$|$230 & 0.35 & 0.74 \\
        \hline
        \multirow{1}{*}{Politik} & KORRUPT \dots & 12$|$0 & 0$|$1 & 6$|$5 & 31$|$43 & 0.33 & 0.65 \\
        \hline
        \multirow{1}{*}{Togstrejke} & De ansatte \dots & 7$|$1 & 3$|$4 & 1$|$0 & 62$|$68 & 0.30 & 0.82 \\
        \hline
        \multirow{1}{*}{Ulve i DK} & Den vedholdende \dots & 1$|$1 & 3$|$1 & 1$|$0 & 50$|$53 & 0.23 & 0.87 \\
        \hline
        \hline
        \multirow{1}{*}{\textit{Overall}} & & 110 & 192 & 50 & 1137 & 0.38 & 0.73 \\
        \hline
    \end{tabular}
    \caption{SDQC distributions and automatic stance labels and results per rumour (see Table \ref{tab:submission_rumours}), as well as overall SDQC, macro $F_1$ and accuracy when combining the predictions across the rumours. For SDQC, the numbers left of the `$|$' are the actual class labels, while the numbers right of the `$|$' are the predicted class labels.}
    \label{tab:auto_stance}
\end{table}

\begin{figure}[h]
    \centering
    \includegraphics[width=0.7\textwidth]{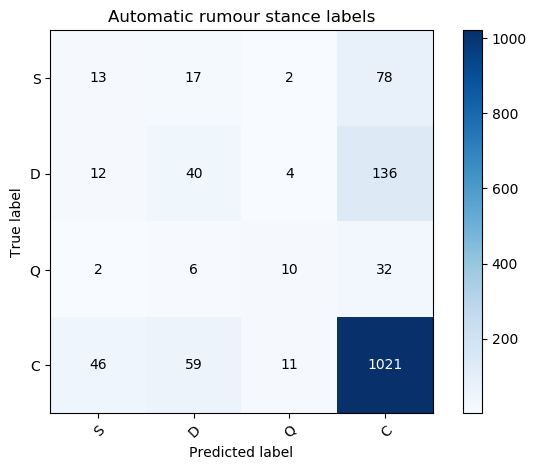}
    \caption{Confusion matrix for the combined automatic rumour stance class labels}
    \label{fig:auto_stance_cm}
\end{figure}

The results of the automatic stance labels on the rumour data are diverse. For instance, the classification with highest macro \fone{} is 0.51 (MeToo), while the lowest macro \fone{} is 0.23 (``Ulve i DK"). For MeToo, with a test size of only 60, just one class label is misclassified, being a ``querying" classified as ``supporting". For ``Ulve i DK", the single ``querying" class label and two of the ``denying" class labels are incorrectly classified as ``commenting". Overall, we see that it is difficult to classify the ``denying" class, such as for ``Uffe Elbæk \dots" and ``23-årig dansk \dots", and in particular it is the tendancy to misclassify SDQ as the majority class ``commenting". This is also evident from the confusion matrix for the combined classification, which is depicted in Figure \ref{fig:auto_stance_cm}.

\subsubsection{Binary classification}
Table \ref{tab:combine_3way_unv_false} shows the results of training on PHEME data and testing on automatic stance labels for \dataset{}. When comparing to the original results in Table \ref{tab:pheme_danish_veracity}, section \ref{subsub:unv_false}, the automatic predictions obtains a lower \fone{} score while they maintain or improve on the accuracy. This could indicate that the majority class is being predicted more often, while the minority is predicted less precisely. The error between the automatic stance labels and the gold labels are generally that there are more ``commenting" class labels in the automatic stance labels. As such the results indicate that there is a correlation between more ``commenting" posts and the ``unverified" rumour class. Some cases do also lose accuracy, however the greatest losses are generally observed in the \fone{} scores while the accuracy scores are nearer the gold label predictions. \\

\begin{table}[h]
    \small
    \centering
    \begin{tabular}{l|c|c|c}
        Structure & Model & Acc. & $F_1$ \\ \hline
        \multirow{3}{*}{SAS}&$\hmm{}$ & 0.81 & 0.64 \\
        &$\mshmm{}$ & 0.75 & \textbf{0.67} \\
        & VB & 0.81 & 0.45 \\ \hline
        \multirow{3}{*}{TCAS}&$\hmm{}$ & 0.79 & 0.56 \\
        &$\mshmm{}$ & 0.68 & 0.55 \\ 
        & VB & 0.76 & 0.43 \\ \hline
        \multirow{3}{*}{BAS}&$\hmm{}$ & \textbf{0.82} & 0.58 \\
        &$\mshmm{}$ & 0.76 & 0.56 \\ 
        & VB & 0.76 & 0.48 \\ \hline
    \end{tabular}
    \caption{Training on the PHEME dataset and testing on automatic stance labels generated for \dataset{} where unverified is false}
    \label{tab:combine_3way_unv_false}
\end{table}

Table \ref{tab:combine_3way_unv_true} shows the results of training on the PHEME dataset and testing on automatic stance labels for \dataset{} where unverified is interpreted as true. The tendencies are much the same as the previous table. The results with the automatic stance labels are generally worse than the gold label predictions shown in appendix \ref{app_veracity:unv_true}, Table \ref{tab:pheme_danish_veracity_unv_true}. The \fone{} scores are however the main difference, as the accuracy does not loose as much performance.

\begin{table}[h]
    \small
    \centering
    \begin{tabular}{l|c|c|c}
        Structure & Model & Acc. & $F_1$ \\ \hline
        \multirow{3}{*}{SAS}&$\hmm{}$ & \textbf{0.81} & 0.45 \\
        &$\mshmm{}$ & 0.79 & \textbf{0.59} \\
        & VB & \textbf{0.81} & 0.45 \\ \hline
        \multirow{3}{*}{TCAS}&$\hmm{}$ & 0.72 & 0.45 \\
        &$\mshmm{}$ & 0.75 & 0.46 \\ 
        & VB & 0.66 & 0.43 \\ \hline
        \multirow{3}{*}{BAS}&$\hmm{}$ & 0.63 & 0.49 \\
        &$\mshmm{}$ & 0.59 & 0.48 \\ 
        & VB & 0.60 & 0.51 \\ \hline
    \end{tabular}
    \caption{Training on the PHEME dataset and testing on automatic stance labels generated for \dataset{} where unverified is true}
    \label{tab:combine_3way_unv_true}
\end{table}
\subsubsection{Three-way classification}
Table \ref{tab:combine_3way} displays the results for three-way classification on the automatic stance labels. When comparing the results to the gold stance label results in Table \ref{tab:pheme_danish_veracity_3way}, appendix \ref{app_veracity:3way}, the tendencies are much alike the ones observed for unverified false and true.

\begin{table}[h]
    \small
    \centering
    \begin{tabular}{l|c|c|c}
        Structure & Model & Acc. & $F_1$ \\ \hline
        \multirow{3}{*}{SAS}&$\hmm{}$ & \textbf{0.62} & 0.26 \\
        &$\mshmm{}$ & 0.56 & \textbf{0.39} \\ 
        & VB & 0.62 & 0.26 \\ \hline
        \multirow{3}{*}{TCAS}&$\hmm{}$ & 0.48 & 0.29 \\
        &$\mshmm{}$ & 0.55 & 0.33 \\ 
        & VB & 0.51 & 0.23 \\ \hline
        \multirow{3}{*}{BAS}&$\hmm{}$ & 0.44 & 0.33 \\
        &$\mshmm{}$ & 0.44 & 0.34 \\ 
        & VB & 0.45 & 0.30 \\ \hline
    \end{tabular}
    \caption{Training on the PHEME dataset and testing on automatic stance labels generated for \dataset{} for three-way classification}
    \label{tab:combine_3way}
\end{table}

\subsubsection{Rumour veracity prediction example}
\label{veracity_example}
To display the use of the conducted research of this paper in a practical way, a small proof of concept command line tool has been developed\footnote{\url{https://github.com/danish-stance-detectors/RumourResolution}}. The tool contains a pretrained Support Vector Machine trained on \dataset{} to perform stance classification. Further it contains a Hidden Markov Model $\hmm{}$ pretrained on \pheme{} data for applying veracity classification with ``unverified" rumour class interpreted as ``false". Finally a number of data fetching, wrapping and preprocessing functionality has been included to enable ``live" rumour veracity resolution of Reddit submissions. The tool can download all comments from a specific submission, perform stance classification on the comments and then veracity classification on the submission.

While this is merely an example, a Reddit submission discussing a possible wolf attack on a pack of sheep has been highlighted\footnote{\url{https://www.reddit.com/r/Denmark/comments/b9b009/25_f\%C3\%A5r_d\%C3\%B8de_i_muligt_ulveangreb_der_var_klippet/}  accessed 30-05-2019}. The submission has 19 comments and the news article provided in the submission post could not conclude whether the attack was done by wolves or not. The command line tool classified the then unverified rumour as ``true". 

A later Reddit submission links to a news article in which wolves were concluded as the culprit of the attack described above, given DNA tests\footnote{\url{https://www.reddit.com/r/Denmark/comments/bn16w8/ulve_stod_bag_26_f\%C3\%A5rs_d\%C3\%B8d_i_vestjylland/} accessed 30-05-2019}. This article confirms the first unverified rumour to be true and as such shows some of the functionality of the veracity classification research done in this project. This is one single example and the tool will not be correct for all rumours. It does however show how the research conducted in this project can be practically applicable and may assist in classifying the veracity of unverified rumours.

\subsection{Evaluation of veracity classification experiments}
\label{veracity:eval}
The experiments and results hereof reported in section \ref{sec:rumour_experiments} investigates whether the stance label approach is applicable across language and platform. Further the experiments investigate how different interpretations of unverified rumours reflects in the rumour resolution classification. Lastly experiments are done to research the robustness of the rumour resolution system. In other words: how well the gold label results translate to new data with automatically generated stance labels. \\

The best results for this approach were seen with the $\lambda$ model with the unverified rumours interpreted as false. The approach scored above 0.80 accuracy and between 0.66 and 0.71 \fone{} score across the different \dataset{} structures. These results are promising and show that sequences of stance labels can be reused across languages. The results for 3-fold cross validation on the mix of \pheme{} data and \dataset{} were not quite as strong. This indicates that in spite of the good results when testing solely on \dataset{}, there are some discrepancies between the datasets. 

This is also seen in the results for the Hidden Markov Model $\mshmm{}$. It performs well when testing on data from one dataset, but does not perform as well as $\lambda$ across datasets. This indicates the existence of discrepancies between the two datasets specifically in relation to the posting time tendencies. \\

The experiments on interpreting the unverified rumours as respectively true and false favour treating them as false. This indicates that the unverified rumours are more alike the false rumours found in the training data. In other words this could indicate that the crowd stance for an unverified rumour is more alike a false rumour than a true rumour. \\

The experiments for performing veracity classification on automatic stance labels display interesting results. A drop in \fone{} performance when using automatic stance labels is observed, however the accuracy is generally not impacted negatively. The best result for the automatic binary classification are achieved by regarding ``unverified" as ``false". In this case, we observe an accuracy of 0.82 (on BAS) while the highest \fone{} achieved was 0.67 (on SAS). However the SAS structure is a small data sample of only 16 submissions, and thus may not be representative. The highest \fone{} score aside from the SAS result is 0.58 \fone{} (on BAS) with the $\hmm{}$ model.

The relatively low drop in performance with automatic stance labels is promising, showing that the system has practical applications, as it seems to generalise well to new data.

%% file: discussion.tex
\section{Discussion}
\label{discussion}


The results for Danish stance classification and rumour veracity classification presented in \ref{veracity:eval} are promising. However a number of things must be kept in mind when reviewing these. The \dataset{} dataset contains 3,007 stance data points, which is less than optimal. More data would probably facilitate better results and increase the chance of the dataset being representative for new unseen data. Further as shown in \cite{zubiaga16} and \cite{AnnotateItalian_TW-BS} the process of annotating a dataset is resource intensive and difficult. \dataset{} has been gathered and annotated by us, two people, in a time period of three months. Neither of us are experts in linguistics or journalism as some annotators in \cite{zubiaga16}. This might affect the correctness of the labels. The low amount of resources available for making \dataset{} also affected the final size of the dataset. If more resources had been available in the form of more annotators or experts, the dataset would probably be larger. Further the certainty in the correctness of each ground truth label would increase and might lessen the skewedness of the dataset towards the ``commenting" label. 

A larger dataset could have added more rumours for which the actual truth value was known. 10 out of the 16 rumour submissions gathered in \dataset{} were annotated as being \q{unverified}. This meant that in order to have a meaningful amount of rumour data, the \q{unverified} rumours were needed. Either the veracity prediction had to be considered as a three-way classification task or the \q{unverified} rumours interpreted as either \q{true} or \q{false}. While the results of the veracity classification system are promising, the confidence in the results could be strengthened by more rumour data with a known veracity value.  \\

Time and resource constraints also meant that Reddit is the only platform from which data was gathered for \dataset{}. This might cause issues in relation to the model organism problem introduced in \cite{tufekci14}. The model organism problem states that basing research in data from a single platform might warp the results and usefulness of the research towards the specific platform. Even though platform specific features as mentioned in \ref{features} can be left out, it would be valuable to extend \dataset{} with data from other platforms to mitigate the model organism problem. 

However the experiments carried out in section \ref{experiments_veracity} show promising results for applying the stance labels across platforms. This indicates that the data in \dataset{} is applicable in a broader context and not restrained by the Reddit platform. This indication could be confirmed or refuted with the addition of data from other platforms and experiments on it.\\

The approaches taken to stance classification have been inspired from several \sota{} approaches to the problem. These include the deep learning LSTM approach by \cite{kochkina17} and the idea of focusing on feature engineering with non-neural network classifiers by \cite{aker17}. The better performance is achieved by latter approach with a Linear Support Vector Machine model. It is however important to keep in mind that the LSTM approach was somewhat expected to have difficulty to perform well due to the small size of \dataset{}. It would have been interesting to apply another powerful approach, such as the Bi-LSTM implemented in \cite{augenstein16}. 

Further the BERT approach employed by \cite{butfit:semeval2019} in RumourEval 2019 \cite{semeval_2019} achieves very promising results, outperforming earlier \sota{} approaches. The BERT method would however have been difficult to reproduce given the comprehensive pre-training step: the model was pre-trained on two English corpora consisting of more than 3 million words combined. Further the promising results were obtained on a dataset than twice as large as \dataset{}. However given more time and possibly more data it could have been interesting to apply the BERT model on \dataset{}. \\



The data sampling presented in section \ref{subsec:data_sampling} was as stated an attempt to make the class distribution more evenly aligned. While the approaches do facilitate a much less skewed class distribution (see table \ref{tab:sampling_pct}), the artificial data only display improvements in one area. Table \ref{tab:svm_sample} shows that the \textit{svm} model improves with super-sampling, while sub-sampling decreases performance. This indicates two things: (1) the model gets better with more data, and (2) given more data, the model actually takes advantage of an improved SDQC class label distribution. This also correlates with the learning curves included in appendix \ref{app:learning_curves}, figure \ref{fig:svm_learningcurves}, which indicates that stance classification could become better with an extended \dataset{} dataset. This is especially true if the SDQC distribution becomes more even. A stronger stance classifier would also strengthen the results seen in the veracity classification for automatic stance labels, since it relies on the correctness of the automatic labels.

The skewed distribution of the data further entices the use of a two-step approach as implemented by \cite{two_step_stance}. This approach suggests making a classifier to filter the \q{commenting} and non-``commenting" class labels, and then use another classifier on the minority class labels. The two-step classifier was implemented as a way to tackle the skewed class label distribution of the RumourEval dataset \cite{derczynski17}. Therefore this approach is especially interesting for the case of \dataset{} given the particularly skewed nature of the class label distribution.

%% file: conclusion.tex
\section{Conclusion}
\label{conclusion}
This thesis project has investigated \sota{} approaches for stance classification and rumour veracity prediction. The approaches have been analysed and from these the necessary steps to perform Danish rumour resolution have been dictated. A dataset of Danish posts from Reddit has been generated, including a number of rumourous submissions. The posts have been annotated according to the widely researched SDQC annotation scheme \cite{zubiaga16}. The dataset constitutes the first Danish stance-annotated dataset \dataset{} \cite{lillie_middelboe_2019}, which enables Danish stance classification. The best stance classification results were achieved by a Linear Support Vector Machine classifier, which outperformed a number of other models including an LSTM deep learning model.

Rumour veracity classification relying solely on stance labels and time stamps have been applied to \dataset{}, utilising a Hidden Markov Model as described in \cite{dungs18}. Experiments utilising the PHEME dataset \cite{pheme-dataset} have been conducted to investigate the effectiveness of using data across languages and platforms. Results from using the stance classifier and veracity classifier in connection are promising, indicating the system can assist in rumour resolution for new unseen data. \\

As such this thesis project contributes with a Danish stance-annotated dataset consisting of 3,007 Reddit posts. Further a Linear Support Vector Machine has been deemed an effective stance classifier, scoring an accuracy of 0.76 and a macro \fone{} score of 0.42. Finally a Hidden Markov Model has been used to classify veracity of Danish rumours obtaining an accuracy of 0.82 and an \fone{} score of 0.67 when training on data from another platform and language. A performance of 0.83 in accuracy and 0.68 in \fone{} is observed when relying only on the \dataset{} dataset. When using automatic stance labels for the HMM, only a small drop in performance is observed, showing that the implemented system can have practical applications.

%% file: dataset.tex
\section{Dataset}
\subsection{Reddit queries}
\label{app:reddit_queries}
Note that the list is copy-pasted from a CSV file, hence the format.\\

\noindent topic,query,after,before,score \\
peter-madsen,"peter madsen",2017-8-1,2018-1-1,50\\
peter-madsen,ubåd,2017-8-1,2018-1-1,10\\
peter-madsen,"Kim Wall",2017-8-1,2018-1-1,50\\
ulve,ulv,2012-1-1,2019-1-1,5\\
ulve,ulvesagen,2012-1-1,2019-1-1,0\\
hpv,"hpv vaccine",2015-1-1,2019-1-1,5\\
hpv,"hpv-vaccine",2015-1-1,2019-1-1,5\\
hpv,"de vaccinerede piger",2015-1-1,2019-1-1,5\\
anna-mee,"Anna Mee",2017-1-1,2019-1-1,20\\
anna-mee,"Paradise Papers",2017-1-1,2019-1-1,5\\
klima,klima,2015-1-1,2019-1-1,5\\
metoo,metoo,2016-1-1,2019-1-1,10\\
kost,kost,2018-1-1,2019-1-1,5\\
kost,veganer,2018-1-1,2019-1-1,5\\
kost,vegetar,2018-1-1,2019-1-1,0\\
trump,trump,2016-1-1,2019-1-3,50\\
trump,"donald trump",2016-1-1,2019-1-3,50\\

\clearpage
\subsection{Extracted Reddit data}
\label{app:extracted_reddit_data}
\begin{verbatim}
{
    "title": "5G-teknologien er en miljøtrussel, som bør stoppes",
    "text": "",
    "submission_id": "ax70y5",
    "created": "2019-03-04 13:17:12",
    "num_comments": 43,
    "url": "/r/Denmark/comments/ax70y5/5gteknologien_er_en_miljøtrussel...",
    "text_url": "https://www.information.dk/debat/2019/02/5g-teknologien-...",
    "upvotes": 0,
    "is_video": false,
    "user": {
        "id": "5khw3",
        "karma": 9789,
        "created": "2011-07-26 08:25:10",
        "gold_status": false,
        "is_employee": false,
        "has_verified_email": false
    },
    "subreddit": {
        "name": "Denmark",
        "subreddit_id": "t5_2qjto",
        "created": "2008-07-08 19:19:11",
        "subscribers": 110799
    },
    "comments": [
        {
            "comment_id": "ehrpmiv",
            "text": ">Pernille Schriver, Vibeke Frøkjær Jensen og ...",
            "is_deleted": false,
            "created": "2019-03-04 15:01:17",
            "is_submitter": false,
            "submission_id": "t3_ax70y5",
            "parent_id": "t3_ax70y5",
            "comment_url": "/r/Denmark/comments/ax70y5/5gteknologien_...",
            "upvotes": 15,
            "replies": 1,
            "user": { ... }
        },
        ...
    ]
}
\end{verbatim}

\clearpage
\subsection{Event ideas}
\label{app:data_event_ideas}
\begin{itemize}
    \itemsep0em
    \item Britta Nielsen svindel
    \item Danske Bank hvidvask
    \item Minister, kontor for 130k
    \item 25 år og borgmester
    \item Wozniaki har gigt ? - vinder første grandslam
    \item Bendtner og taxa
    \item Håndbold-mænd, vinder VM i Rio
    \item Løkke og jakkesætsskandale
    \item Udbytteskatskandale
    \item Stein Bagger
    \item Dong Energy
    \item Ubådssagen
    \item Ulve i Danmark
    \item Skatteskandalen
    \item MeToo, Peter Ålbæk
    \item Klimadebat - Varm/kold sommer
    \item Amatørlandsholdet
    \item Facebook hack
    \item Dantaxa skattely
    \item Maersk hack
    \item HPV vaccine
    \item Kostråd
    \item Togulykke
    \item Fodboldtransfer
    \item Anna Mee Allerslev
    \item Korrupt politiker: Esben Lunde Larsen
    \item Burkaforbud
    \item Post Nord
\end{itemize}

%% file: experiment_data.tex
\section{Experiments data}
\label{appendix:experiment_data}

\subsection{Most frequent words}
\label{appendix:most_frequent_words}
Below are the generated words for the ``Most frequent words" features, when considering the 100 most frequent words per SDQC class, and then filtering out words appearing for all classes. Note that the words ``urlurlurl" and ``refrefref" are replacements for respectively URLs and quotes(see section \ref{features}).

\subsubsection{Supporting}
urlurlurl, refrefref, da, fordi, børn, alle, nok, ville, hele, over, må, synes, mange, flere, derfor, ting, se, mennesker, gør, efter, samme, kommer, andet, alt, været, går, blive, andre, tage, langt, gøre, gå, får.

\subsubsection{Denying}
refrefref, nok, alle, mange, fordi, da, andre, over, siger, havde, får, urlurlurl, ham, hende, b12, min, skulle, giver, børn, andet, alt, været, selvfølgelig, uden, samme, hvordan, tilskud, kommer, gør, kun, kost, flere, penge.

\subsubsection{Querying}
hvorfor, måske, hvordan, skulle, hele, ville, ulykke, sådan, politiet, nogle, været, vores, vel, programmet, må, kun, hende, havde, gøre, denne, the, står, sikkert, side, se, journalisten, ham, gjort, fald, co2, blev, 5g, år.

\subsubsection{Commenting}
refrefref, da, urlurlurl, fordi, nok, alle, gør, dig, andre, ville, mange, se, kun, får, over, må, din, samme, kommer, min, havde, siger, gøre, måske, efter, nogle, uden, altså, mener, spise, mennesker, ingen, andet.

\clearpage
\subsection{Super-sampling}
\label{appendix:super_sampling}
Examples of word replacement in super-sampling. Three strategies have been performed, being replacement with most similar word from respectively word2vec and fastText word vectors, and replacement by looking up synonyms in a dictionary. Note that for the word vector cases, only preprocessed tokens are replaced, while with synonyms, replacement is performed on the non-preprocessed text. 

\subsubsection{Example 1}
\paragraph{Original:} ``tankegangen er at de bredeste skuldre kan bære de tungeste læs der er ikke nogen mening i at give de svageste et lige så tungt læs som de stærkeste det bliver de svage bare kvast af"
\paragraph{Word2vec:} ``tankegangen er at de bredeste skuldre kunne udholde de tungeste læs hist er ej nogen betydning i at forære de svageste et sidestykke udså tungt læs ligesom de stærkeste det bliver de svage bare kvast af"
\paragraph{FastText:} ``evidenstankegangen er atr dissekere bredest skuldrene kunne udholde dissekere tungest læs der er ikke nogen mening i at give de svageste et lige så tungt læs som de stærkeste det bliver de svage bare kvast af"
\paragraph{Synonyms:} ``Tankegangen er at de bredeste skuldre kan udholde de tungeste læs. Hist er ej nogen som helst formål i at fremføre de svageste et netop så tungt læs som de stærkeste, det bliver de svage når blot visk af."

\subsubsection{Example 2}
\paragraph{Original:} ``sku da cool nok at have næsten samme record som chris kyle og endda en større bounty chris med 4 tours i irak og 150ish kills fik hele 80.000 dowwahs på sit hovede"
\paragraph{Word2vec:} ``sku da nøgtern sikkert at kolonihave næsten ens collection ligesom chris smide og ovenikøbet en større bounty chris med 4 tours i irak og 150ish kills fik hele 80.000 dowwahs på sit hovede"
\paragraph{FastText:} ``skuvoy pvda nøgtern sikkert atr kolonihave gæsten ens record som chris kyle og endda en større bounty chris med 4 tours i irak og 150ish kills fik hele 80.000 dowwahs på sit hovede"
\paragraph{Synonyms:} ``Sku da sej yderligere at kolonihave næsten samme record som Chris Smide, og ovenikøbet en større bounty! (Chris, i kraft af 4 tours i Irak og 150ish kills, fik totalitet 80.000 dowwahs inden for sit hovede)"

\subsection{Parameter search}
\label{app:parameter_search}
\subsubsection{Learning curves}
\label{app:learning_curves}

\begin{figure}[h]
    \centering
    \begin{minipage}{.5\textwidth}
        \centering
        \includegraphics[width=\textwidth]{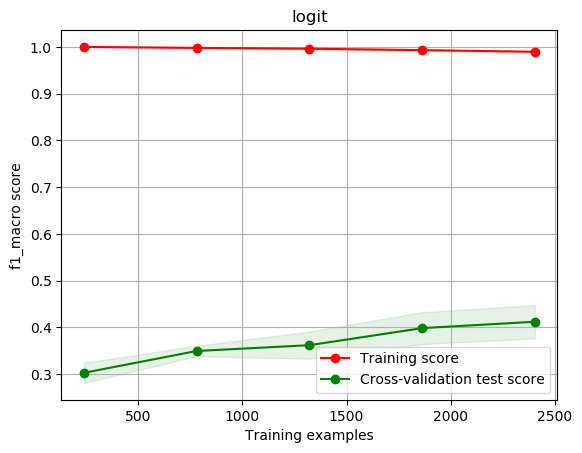}
    \end{minipage}%
    \begin{minipage}{0.5\textwidth}
        \centering
        \includegraphics[width=\textwidth]{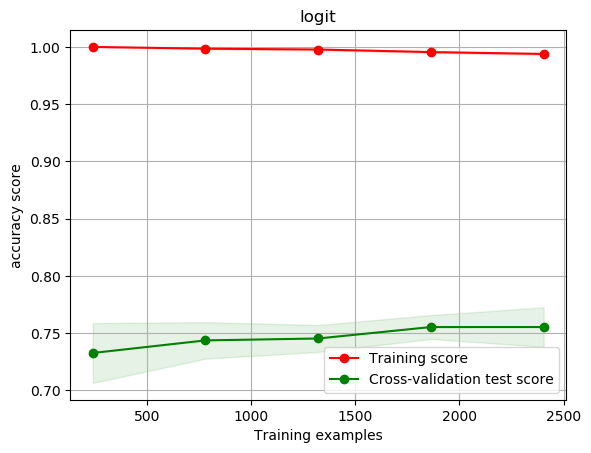}
    \end{minipage}
    \caption{\textit{logit} learning curves for macro $F_1$ (left) and accuracy (right)}
    \label{fig:logit_learningcurves}
\end{figure}

\begin{figure}[h]
    \centering
    \begin{minipage}{.5\textwidth}
        \centering
        \includegraphics[width=\textwidth]{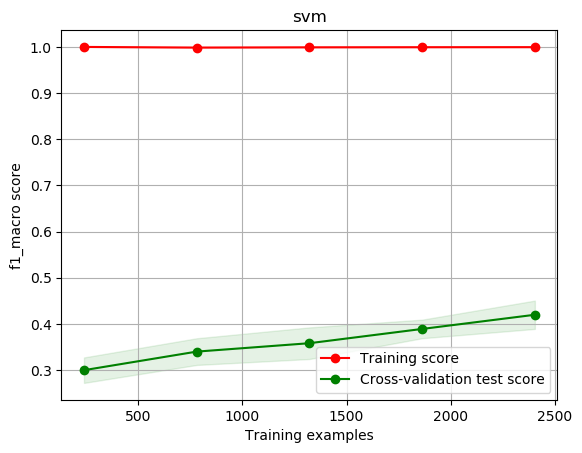}
    \end{minipage}%
    \begin{minipage}{0.5\textwidth}
        \centering
        \includegraphics[width=\textwidth]{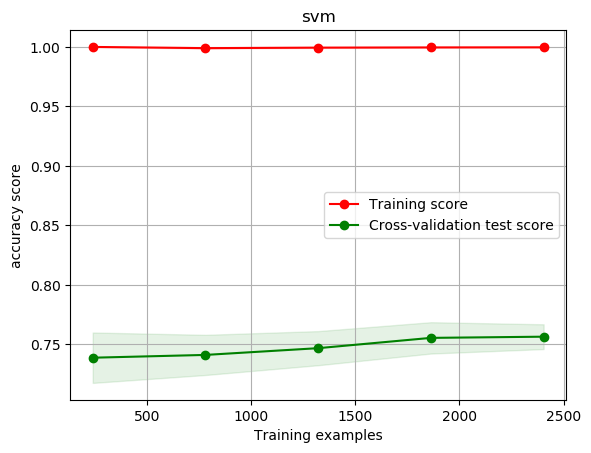}
    \end{minipage}
    \caption{\textit{svm} learning curves for macro $F_1$ (left) and accuracy (right)}
    \label{fig:svm_learningcurves}
\end{figure}

%% file: app_experiments_veracity.tex
\subsection{Veracity experiments}
\label{app_veracity}
This appendix section contains tables for the veracity experiments performed for unverified rumours interpreted as true and for three-way classification.

\subsubsection{Unverified as true results}
\label{app_veracity:unv_true}
\begin{table}[h]
    \centering
    \begin{tabular}{l|c|c|c}
        Structure & Model & Acc. & $F_1$ \\ \hline
        \multirow{3}{*}{SAS}&$\hmm{}$ & 0.74 (+/- 0.21) & 0.49 (+/- 0.13) \\
        &$\mshmm{}$ & \textbf{0.74} (+/- 0.21) & 0.53 (+/- 0.33) \\ 
        & VB & 0.19 (+/- 0.03) & 0.16 (+/- 0.02) \\ \hline
        \multirow{3}{*}{TCAS}&$\hmm{}$ & 0.67 (+/- 0.09) & 0.55 (+/- 0.08) \\
        &$\mshmm{}$ & 0.65 (+/- 0.16) & 0.49 (+/- 0.16) \\ 
        & VB & 0.34 (+/- 0.02) & 0.34 (+/- 0.02) \\ \hline
        \multirow{3}{*}{BAS}&$\hmm{}$ & 0.61 (+/- 0.05) & 0.54 (+/- 0.07) \\
        &$\mshmm{}$ & 0.71 (+/- 0.06) & \textbf{0.62} (+/- 0.05) \\
        & VB & 0.59 (+/- 0.10) & 0.54 (+/- 0.03) \\ \hline
    \end{tabular}
    \caption{Danish veracity results on 3-fold cross validation for unverified being true}
    \label{tab:danish_veracity_unv_true}
\end{table}

\begin{table}[h]
    \centering
    \begin{tabular}{l|c|c|c}
        Structure & Model & Acc. & $F_1$ \\ \hline
        \multirow{3}{*}{SAS} & $\hmm{}$ & 0.75 & \textbf{0.59} \\
        & $\mshmm{}$ & \textbf{0.81} & 0.45 \\ 
        & VB & 0.69 & 0.54 \\ \hline
        \multirow{3}{*}{TCAS} & $\hmm{}$ & 0.72 & 0.54 \\
        & $\mshmm{}$ & 0.76 & 0.52 \\ 
        & VB & 0.70 & 0.56 \\ \hline
        \multirow{3}{*}{BAS} & $\hmm{}$ & 0.62 & 0.56 \\
        & $\mshmm{}$ & 0.60 & 0.51 \\ 
        & VB & 0.61 & 0.58 \\ \hline
    \end{tabular}
    \caption{Training on PHEME and testing on \dataset{} where unverified is set to true}
    \label{tab:pheme_danish_veracity_unv_true}
\end{table}

\begin{table}[h]
    \centering
    \begin{tabular}{l|c|c|c}
        Structure & Model & Acc. & $F_1$ \\ \hline
        \multirow{3}{*}{SAS}&$\hmm{}$ & \textbf{0.84} (+/- 0.02) & 0.52 (+/- 0.13) \\
        &$\mshmm{}$ & 0.78 (+/- 0.09) & 0.51 (+/- 0.05) \\ 
        & VB & 0.43 (+/- 0.33) & 0.40 (+/- 0.24) \\ \hline
        \multirow{3}{*}{TCAS}&$\hmm{}$ & 0.76 (+/- 0.11) & 0.50 (+/- 0.09) \\
        &$\mshmm{}$ & 0.78 (+/- 0.00) & 0.52 (+/- 0.03) \\ 
        & VB & 0.39 (+/- 0.13) & 0.38 (+/- 0.09) \\ \hline
        \multirow{3}{*}{BAS}&$\hmm{}$ & 0.71 (+/- 0.02) & 0.50 (+/- 0.04) \\
        &$\mshmm{}$ & 0.69 (+/- 0.04) & 0.57 (+/- 0.06) \\ 
        & VB & 0.67 (+/- 0.05) & \textbf{0.62} (+/- 0.05) \\ \hline \hline
        \multirow{3}{*}{None}&$\hmm{}$ & \textbf{0.84} (+/- 0.03) & 0.54 (+/- 0.14) \\
        &$\mshmm{}$ & 0.82 (+/- 0.02) & 0.50 (+/- 0.07) \\ 
        & VB & 0.64 (+/- 0.35) &  0.46 (+/- 0.16) \\ \hline
    \end{tabular}
    \caption{Training and testing on mix of PHEME and different \dataset{} structures for unverified true}
    \label{tab:mix_veracity_unv}
\end{table}

\clearpage

\subsubsection{Three-way classification results}
\label{app_veracity:3way}

\begin{table}[h]
    \centering
    \begin{tabular}{l|c|c|c}
        Structure & Model & Acc. & $F_1$ \\ \hline
        \multirow{3}{*}{SAS}&$\hmm{}$ & 0.56 (+/- 0.23) & 0.33 (+/- 0.20) \\
        &$\mshmm{}$ & \textbf{0.61} (+/- 0.35) & 0.37 (+/- 0.37) \\ 
        & VB & 0.31 (+/- 0.17) & 0.35 (+/- 0.30) \\ \hline
        \multirow{3}{*}{TCAS}&$\hmm{}$ & 0.49 (+/- 0.08) & 0.44 (+/- 0.07) \\
        &$\mshmm{}$ & 0.48 (+/- 0.14) & 0.39 (+/- 0.03) \\ 
        & VB & 0.26 (+/- 0.10) & 0.24 (+/- 0.10) \\ \hline
        \multirow{3}{*}{BAS}&$\hmm{}$ & 0.44 (+/- 0.04) & 0.44 (+/- 0.02) \\
        &$\mshmm{}$ & \textbf{0.57} (+/- 0.03) & \textbf{0.53} (+/- 0.04) \\
        & VB & 0.26 (+/- 0.02) & 0.25 (+/- 0.03) \\ \hline
    \end{tabular}
    \caption{Danish veracity results on 3-fold cross validation for three-way rumour classification}
    \label{tab:danish_veracity_3way}
\end{table}

\begin{table}[h]
    \centering
    \begin{tabular}{l|c|c|c}
        Structure & Model & Acc. & $F_1$ \\ \hline
        \multirow{3}{*}{SAS} & $\hmm{}$ & 0.56 & 0.37 \\
        & $\mshmm{}$ & 0.56 & \textbf{0.41} \\ 
        & VB & \textbf{0.62} & 0.38 \\ \hline
        \multirow{3}{*}{TCAS} & $\hmm{}$ & 0.42 & 0.36 \\
        & $\mshmm{}$ & 0.53 & 0.40 \\ 
        & VB & 0.52 & \textbf{0.41} \\ \hline
        \multirow{3}{*}{BAS} & $\hmm{}$ & 0.33 & 0.32 \\
        & $\mshmm{}$ & 0.47 & 0.35 \\ 
        & VB & 0.45 & 0.40 \\ \hline
    \end{tabular}
    \caption{Training on PHEME and testing on \dataset{} for three-way classification}
    \label{tab:pheme_danish_veracity_3way}
\end{table}

\begin{table}[h]
    \centering
    \begin{tabular}{|l|l|r|r|r|}
    \hline
    Structure & \diagbox{\textit{Predicted}}{\textit{Actual}} & False & True & Unverified \\
    \hline
    \hline
    \multirow{2}{*}{BAS} & False & 73 & 10 & 103 \\ 
    \cline{2-5}
    & True & 9 & 46 & 44 \\
    \cline{2-5}
    & Unverified & 57 & 41 & 213 \\
    \hline
    \end{tabular}
    \caption{Truth value distribution with PHEME training and BAS structure testing}
    \label{tab:pheme_danish_cm_3way}
\end{table}

\begin{table}[h]
    \centering
    \begin{tabular}{l|c|c|c}
        Structure & Model & Acc. & $F_1$ \\ \hline
        \multirow{3}{*}{SAS}&$\hmm{}$ & 0.49 (+/- 0.08) & 0.37 (+/- 0.08) \\
        &$\mshmm{}$ & 0.44 (+/- 0.09) & 0.38 (+/- 0.07) \\ 
        & VB & 0.36 (+/- 0.11) & 0.30 (+/- 0.10) \\ \hline
        \multirow{3}{*}{TCAS}&$\hmm{}$ & \textbf{0.53} (+/- 0.09) & 0.40 (+/- 0.03) \\
        &$\mshmm{}$ & 0.52 (+/- 0.05) & \textbf{0.42} (+/- 0.05) \\ 
        & VB & 0.31 (+/- 0.03) & 0.29 (+/- 0.02) \\ \hline
        \multirow{3}{*}{BAS}&$\hmm{}$ & 0.42 (+/- 0.03) & 0.38 (+/- 0.06) \\
        &$\mshmm{}$ & 0.43 (+/- 0.04) & \textbf{0.42} (+/- 0.06) \\ 
        & VB & 0.41 (+/- 0.06) & 0.38 (+/- 0.10) \\ \hline \hline
        \multirow{3}{*}{None}&$\hmm{}$ & 0.50 (+/- 0.08) & 0.32 (+/- 0.08) \\
        &$\mshmm{}$ & 0.50 (+/- 0.07) & 0.35 (+/- 0.04) \\ 
        & VB & 0.43 (+/- 0.03) & 0.33 (+/- 0.08) \\ \hline
    \end{tabular}
    \caption{Training and testing on mix of PHEME and different \dataset{} structures for three-way classification}
    \label{tab:mix_veracity_3way}
\end{table}

%% file: main.bbl
\begin{thebibliography}{}

\bibitem[Aker et~al., 2017]{aker17}
Aker, A., Derczynski, L., and Bontcheva, K. (2017).
\newblock {Simple Open Stance Classification for Rumour Analysis}.
\newblock In {\em Proceedings of Recent Advances in Natural Language
  Processing}, pages 31--39, Varna, Bulgaria.

\bibitem[Al-Rfou et~al., 2013]{polyglot:2013:ACL-CoNLL}
Al-Rfou, R., Perozzi, B., and Skiena, S. (2013).
\newblock Polyglot: Distributed word representations for multilingual nlp.
\newblock In {\em Proceedings of the Seventeenth Conference on Computational
  Natural Language Learning}, pages 183--192, Sofia, Bulgaria. Association for
  Computational Linguistics.

\bibitem[Augenstein et~al., 2016]{augenstein16}
Augenstein, I., Rocktäschel, T., Vlachos, A., and Bontcheva, K. (2016).
\newblock {Stance Detection with Bidirectional Conditional Encoding}.
\newblock In {\em Proceedings of the 2016 Conference on Empirical Methods in
  Natural Language Processing}, pages 876--885, Austin, Texas.

\bibitem[Babakar and Moy, 2016]{babakar17fullfact}
Babakar, M. and Moy, W. (2016).
\newblock The state of automated factchecking.

\bibitem[Baris et~al., 2019]{clearrumor:semeval2019}
Baris, I., Schmelzeisen, L., and Staab, S. (2019).
\newblock Clearumor at semeval-2019 task 7: Convolving elmo against rumors.

\bibitem[Bird et~al., 2009]{nltk_book}
Bird, S., Loper, E., and Klein, E. (2009).
\newblock Natural language processing with python.

\bibitem[Castillo et~al., 2011]{castillo11}
Castillo, C., Medoza, M., and Poblete, B. (2011).
\newblock {Information Credibility on Twitter}.
\newblock In {\em WWW 2011 – Session: Information Credibility}, pages
  675--684, Hyderabad, India.

\bibitem[Derczynski and Bontcheva, 2014]{pheme14}
Derczynski, L. and Bontcheva, K. (2014).
\newblock {\texttt{PHEME:} Veracity in Digital Social Networks}.

\bibitem[Derczynski et~al., 2017]{derczynski17}
Derczynski, L., Bontcheva, K., Liakata, M., Procter, R., Hoi, G. W.~S., and
  Zubiaga, A. (2017).
\newblock {SemEval-2017 Task 8: RumourEval: Determining rumour veracity and
  support for rumours}.
\newblock In {\em Proceedings of the 11th International Workshop on Semantic
  Evaluations (SemEval-2017)}, pages 69--76, Vancouver, Canada.

\bibitem[Dungs et~al., 2018]{dungs18}
Dungs, S., Aker, A., Fuhr, N., and Bontcheva, K. (2018).
\newblock {Can Rumour Stance Alone Predict Veracity?}
\newblock In {\em Proceedings of the 27th International Conference on
  Computational Linguistics}, page 3360–3370, Santa Fe, New Mexico, USA.

\bibitem[Enayet and El-Beltagy, 2017]{enayet17}
Enayet, O. and El-Beltagy, S.~R. (2017).
\newblock {NileTMRG at SemEval-2017 Task 8: Determining Rumour and Veracity
  Support for Rumours on Twitter}.
\newblock In {\em Proceedings of the 11th International Workshop on Semantic
  Evaluations (SemEval-2017)}, pages 470--474, Vancouver, Canada.

\bibitem[Fajcik et~al., 2019]{butfit:semeval2019}
Fajcik, M., Burget, L., and Smrz, P. (2019).
\newblock But-fit at semeval-2019 task 7: Determining the rumour stance with
  pre-trained deep bidirectional transformers.

\bibitem[Giménez et~al., 2017]{gimenez17}
Giménez, M., Baviera, T., Llorca, G., Gámir, J., Calvo, D., Rosso, P., , and
  Rangel, F. (2017).
\newblock {Overview of the 1st Classification of Spanish Election Tweets Task
  at IberEval 2017}.
\newblock In {\em {Second Workshop on Evaluation of Human Language Technologies
  for Iberian Languages (IberEval 2017)}}.

\bibitem[Goldberg, 2016]{neural_primer}
Goldberg, Y. (2016).
\newblock A primer on neural network models for natural language processing.
\newblock {\em Journal of Artificial Intelligence Research 57}, pages 345--420.
\newblock Chapters 10-11.

\bibitem[Gorell et~al., 2018]{semeval_2019}
Gorell, G., Bontcheva, K., Derczynski, L., Kochkina, E., Liakata, M., and
  Zubiaga, A. (2018).
\newblock Rumoureval 2019: Determining rumour veracity and support for rumours.

\bibitem[Grave et~al., 2018]{grave2018learning}
Grave, E., Bojanowski, P., Gupta, P., Joulin, A., and Mikolov, T. (2018).
\newblock Learning word vectors for 157 languages.
\newblock In {\em Proceedings of the International Conference on Language
  Resources and Evaluation (LREC 2018)}.

\bibitem[Han et~al., 2011]{data_mining_book}
Han, J., Kamber, M., and Pei, J. (2011).
\newblock {\em Data Mining, Concepts and Techniques}.
\newblock Morgan Kaufmann Publishers Inc., 3 edition.

\bibitem[Hanselowski et~al., 2018]{hanselowski18}
Hanselowski, A., PVS, A., Schiller, B., Caspelherr, F., Chaudhuri, D., Meyer,
  C.~M., and Gurevych, I. (2018).
\newblock {A Retrospective Analysis of the Fake News Challenge Stance Detection
  Task}.
\newblock In {\em Proceedings of the 27th International Conference on
  Computational Linguistics}, pages 1859--1874, Santa Fe, New Mexico, USA.

\bibitem[Huang et~al., 2015]{huang15}
Huang, Y.~L., Starbird, K., Orand, M., A.~Stanek, S., and T.~Pedersen, H.
  (2015).
\newblock Connected through crisis: Emotional proximity and the spread of
  misinformation online.
\newblock In {\em CSCW '15 Proceedings of the 18th ACM Conference on Computer
  Supported Cooperative Work \& Social Computing}, pages 969--980, Vancouver,
  BC, Canada. Association for Computing Machinery.

\bibitem[Joulin et~al., 2016]{joulin2016bag}
Joulin, A., Grave, E., Bojanowski, P., and Mikolov, T. (2016).
\newblock Bag of tricks for efficient text classification.
\newblock {\em arXiv preprint arXiv:1607.01759}.

\bibitem[Küçük, 2017]{kucuk17}
Küçük, D. (2017).
\newblock Stance detection in turkish tweets.
\newblock In {\em Proceedings of Workshop on Social Media World Sensors
  (SIDEWAYS)}, Prague, Czech Republic.

\bibitem[Kochkina et~al., 2017]{kochkina17}
Kochkina, E., Liakata, M., and Augenstein, I. (2017).
\newblock {Turing at SemEval-2017 Task 8: Sequential Approach to Rumour Stance
  Classification with Branch-LSTM}.
\newblock In {\em Proceedings of the 11th International Workshop on Semantic
  Evaluations (SemEval-2017)}, pages 475--480, Vancouver, Canada.

\bibitem[Lillie and Middelboe, 2018]{thesis-prep}
Lillie, A.~E. and Middelboe, E.~R. (2018).
\newblock {Fake News Detection using Stance Classification: A Survey}.
\newblock {\em arXiv:1907.00181}.

\bibitem[Lillie and Middelboe, 2019]{lillie_middelboe_2019}
Lillie, A.~E. and Middelboe, E.~R. (2019).
\newblock Danish stance-annotated reddit dataset.
\newblock doi: 10.6084/m9.figshare.8217137.v1.

\bibitem[Lou, 1995]{viterbi_paper}
Lou, H.-L. (1995).
\newblock Implementing the viterbi algorithm.
\newblock Chapter: Viterbi Algorithm Applied to HMMs.

\bibitem[Mikolov et~al., 2013]{mikolov13word2vec}
Mikolov, T., Corrado, G., Chen, K., and Dean, J. (2013).
\newblock Efficient estimation of word representations in vector space.
\newblock In {\em Proceedings of the International Conference on Learning
  Representations (ICLR 2013)}, pages 1--12. arXiv preprint arXiv:1301.3781.

\bibitem[Mohammad et~al., 2016]{mohammad16}
Mohammad, S.~M., Kiritchenko, S., Sobhani, P., Zhu, X., and Cherry, C. (2016).
\newblock {SemEval-2016 Task 6: Detecting Stance in Tweets}.
\newblock In {\em Proceedings of SemEval-2016}, pages 31--41, San Diego,
  California.

\bibitem[Pedregosa et~al., 2011]{scikit-learn}
Pedregosa, F., Varoquaux, G., Gramfort, A., Michel, V., Thirion, B., Grisel,
  O., Blondel, M., Prettenhofer, P., Weiss, R., Dubourg, V., Vanderplas, J.,
  Passos, A., Cournapeau, D., Brucher, M., Perrot, M., and Duchesnay, E.
  (2011).
\newblock Scikit-learn: Machine learning in {P}ython.
\newblock {\em Journal of Machine Learning Research}, 12:2825--2830.

\bibitem[Pennington et~al., 2014]{pennington2014glove}
Pennington, J., Socher, R., and Manning, C.~D. (2014).
\newblock Glove: Global vectors for word representation.
\newblock In {\em Empirical Methods in Natural Language Processing (EMNLP)},
  pages 1532--1543.

\bibitem[Pomerleau and Rao, 2017]{fakenewschallenge17}
Pomerleau, D. and Rao, D. (2017).
\newblock Fake news challenge.
\newblock \url{http://www.fakenewschallenge.org} - Visited 26-11-2018 and
  dataset from \url{https://github.com/FakeNewsChallenge/fnc-1} - Visited
  10-12-2018.

\bibitem[Procter et~al., 2013]{procter13}
Procter, R., Vis, F., and Voss, A. (2013).
\newblock {Reading the riots on Twitter: methodological innovation for the
  analysis of big data}.
\newblock {\em International Journal of Social Research Methodology}, pages
  16:3 197--2014.

\bibitem[Qazvinian et~al., 2011]{qazvinian11}
Qazvinian, V., Rosengren, E., Radev, D.~R., and Mei, Q. (2011).
\newblock {Rumor has it: Identifying Misinformation in Microblogs}.
\newblock In {\em Proceedings of the 2011 Conference on Empirical Methods in
  Natural Language Processing}, pages 1589--1599, Edinburgh, Scotland.

\bibitem[Quattrociocchi et~al., 2016]{echochamber}
Quattrociocchi, W., Scala, A., and Sunstein, C.~R. (2016).
\newblock {Echo Chambers on Facebook}.

\bibitem[{\v R}eh{\r u}{\v r}ek and Sojka, 2010]{rehurek_lrec-gensim}
{\v R}eh{\r u}{\v r}ek, R. and Sojka, P. (2010).
\newblock {Software Framework for Topic Modelling with Large Corpora}.
\newblock In {\em {Proceedings of the LREC 2010 Workshop on New Challenges for
  NLP Frameworks}}, pages 45--50, Valletta, Malta. ELRA.
\newblock \url{http://is.muni.cz/publication/884893/en}.

\bibitem[Årup Nielsen, 2011]{afinn}
Årup Nielsen, F. (2011).
\newblock {A new ANEW: evaluation of a word list for sentiment analysis in
  microblogs}.
\newblock In {\em {Proceedings of the ESWC2011 Workshop on 'Making Sense of
  Microposts': Big things come in small packages. Volume 718 in CEUR Workshop
  Proceedings}}, pages 93--98.

\bibitem[Årup Nielsen, 2019]{nielsen19}
Årup Nielsen, F. (2019).
\newblock {Danish resources}.

\bibitem[Shu et~al., 2017]{shu2017fake}
Shu, K., Sliva, A., Wang, S., Tang, J., and Liu, H. (2017).
\newblock Fake news detection on social media: A data mining perspective.
\newblock {\em ACM SIGKDD Explorations Newsletter}, 19(1):22--36.

\bibitem[Stranisci et~al., 2015]{AnnotateItalian_TW-BS}
Stranisci, M., Bosco, C., Farías, D. I.~H., and Patti, V. (2015).
\newblock {Annotating Sentiment and Irony in the Online Italian Political
  Debate on \#labuonascuola}.
\newblock pages 274--279.

\bibitem[Taulé et~al., 2017]{taule17}
Taulé, M., Martí, M.~A., Rangel, F., Rosso, P., Bosco, C., and Patti, V.
  (2017).
\newblock Overview of the task on stance and gender detection in tweets on
  catalan independence at ibereval 2017.
\newblock In {\em Proceedings of the Second Workshop on Evaluation of Human
  Language Technologies for Iberian Languages (IberEval 2017)}, pages 157--177.

\bibitem[Thorne et~al., 2018]{Thorne18Fact}
Thorne, J., Vlachos, A., Cocarascu, O., Christodoulopoulos, C., and Mittal, A.
  (2018).
\newblock The {Fact Extraction and VERification (FEVER)} shared task.
\newblock In {\em Proceedings of the First Workshop on {Fact Extraction and
  VERification (FEVER)}}.

\bibitem[Tufekci, 2014]{tufekci14}
Tufekci, Z. (2014).
\newblock {Big Questions for Social Media Big Data: Representativeness,
  Validity and Other Methodological Pitfalls}.
\newblock In {\em Proceedings of the Eighth International AAAI Conference on
  Weblogs and Social Media}, page 505–514.
\newblock \url{www.aaai.org}.

\bibitem[Wang et~al., 2017]{two_step_stance}
Wang, F., Lan, M., and Wu, Y. (2017).
\newblock Rumour evaluation using effective features and supervised ensemble
  models.
\newblock {\em Proceedings of the 11th International Workshop on Semantic
  Evaluations (SemEval-2017)}, pages 491--496.

\bibitem[Wu et~al., 1999]{baum-welch_paper}
Wu, Y., Ganapathiraju, A., and Picone, J. (1999).
\newblock Baum-welch re-estimation of hidden markov model.
\newblock Chapter 3: Baum-Welch training approach.

\bibitem[Zubiaga et~al., 2018]{zubiaga18}
Zubiaga, A., Aker, A., Bontcheva, K., Liakata, M., and Procter, R. (2018).
\newblock {Detection and Resolution of Rumours in Social Media: A Survey}.
\newblock {\em ACM Computing Surveys, Vol. 51, No. 2, Article 32}.

\bibitem[Zubiaga et~al., 2016a]{pheme-dataset}
Zubiaga, A., Liakata, Maria, Procter, R., Wong Sak~Hoi, G., and Tolmie, P.
  (2016a).
\newblock Pheme rumour scheme dataset: journalism use case.
\newblock Available from: doi:
  \href{http://dx.doi.org/10.6084/m9.figshare.2068650.v1}{10.6084/m9.figshare.2068650.v1}.

\bibitem[Zubiaga et~al., 2015]{zubiaga15}
Zubiaga, A., Liakata, M., Procter, R., Bontcheva, K., and Tolmie, P. (2015).
\newblock {Crowdsourcing the Annotation of Rumourous Conversations in Social
  Media}.
\newblock In {\em Proceedings of the 24th International Conference on World
  Wide Web}, pages 347--353.

\bibitem[Zubiaga et~al., 2016b]{zubiaga16}
Zubiaga, A., Liakata, M., Procter, R., Hoi, G. W.~S., and Tolmie, P. (2016b).
\newblock {Analysing How People Orient to and Spread Rumours in Social Media by
  Looking at Conversational Threads}.
\newblock {\em {PLoS ONE}}.
\newblock 11(3).

\end{thebibliography}
